\pdfoutput=1
\documentclass[10pt,letterpaper]{article}
\makeatletter
\@ifundefined{iflatexml}{\newif\iflatexml\latexmlfalse}{}
\makeatother

\usepackage[T1]{fontenc}
\usepackage{lmodern}
\usepackage[letterpaper,margin=1in,top=0.85in]{geometry}
\usepackage{graphicx}
\usepackage{amsmath,amssymb}
\usepackage{booktabs}
\usepackage[dvipsnames]{xcolor}
\usepackage[authoryear,round]{natbib}
\usepackage{microtype}
\usepackage{array}
\usepackage{multirow}
\usepackage{placeins}
\usepackage{wrapfig}
\usepackage{xurl}
\usepackage[normalem]{ulem}
\usepackage[font=small,labelfont={bf,sf},labelsep=quad,skip=6pt]{caption}
\usepackage{titlesec}
\usepackage[most]{tcolorbox}
\iflatexml
  \usepackage{authblk}
\fi

\definecolor{reportcream}{HTML}{F8F2EA}  %
\definecolor{reportnavy}{HTML}{183040}   %
\definecolor{reportteal}{HTML}{387088}   %
\definecolor{reportmist}{HTML}{EEF2F7}   %
\providecommand{\reportboxcolor}{reportcream}
\colorlet{reportbox}{\reportboxcolor}

\usepackage[colorlinks,linkcolor=reportteal,citecolor=reportteal,urlcolor=reportteal,
  bookmarks=true]{hyperref}
\hypersetup{
  pdftitle={Elucidating the Design Space of End-to-end Latent Representation Learning in World Action Models},
  pdfauthor={Junming Zhao, Shi Sheng, Jiacheng You, Chenxi Li, Bowen Hu, Chengbo Yuan, Yingdong Hu, Weirui Ye, Yang Gao},
  pdfsubject={Technical report},
  pdfkeywords={representation learning, world models, robot manipulation}
}

\newcommand{\reportsans}{\fontfamily{Inter-TLF}\selectfont}

\titleformat{\section}{\sffamily\bfseries\Large}{\thesection}{0.8em}{}
\titleformat{\subsection}{\sffamily\bfseries\large}{\thesubsection}{0.8em}{}
\titleformat{\subsubsection}{\sffamily\bfseries\normalsize}{\thesubsubsection}{0.8em}{}
\titlespacing*{\section}{0pt}{16pt plus 4pt minus 2pt}{8pt plus 2pt minus 2pt}
\titlespacing*{\subsection}{0pt}{12pt plus 3pt minus 2pt}{6pt plus 2pt minus 1pt}
\titlespacing*{\subsubsection}{0pt}{10pt plus 3pt minus 2pt}{4pt plus 2pt minus 1pt}
\newif\ifcolumnfloats
\columnfloatstrue
\makeatletter
\let\report@figure\figure
\let\report@endfigure\endfigure
\let\report@table\table
\let\report@endtable\endtable
\renewenvironment{figure}[1][tbp]{%
  \report@figure[#1]\ifcolumnfloats\centering
  \begin{minipage}{0.6\textwidth}\setlength{\columnwidth}{\linewidth}\fi}{%
  \ifcolumnfloats\end{minipage}\fi\report@endfigure}
\renewenvironment{table}[1][tbp]{%
  \report@table[#1]\ifcolumnfloats\centering
  \begin{minipage}{0.7\textwidth}\setlength{\columnwidth}{\linewidth}\fi}{%
  \ifcolumnfloats\end{minipage}\fi\report@endtable}
\makeatother

\iflatexml
  \newcommand{\reportlayout}[2]{#2}
\else
  \newcommand{\reportlayout}[2]{#1}
\fi

\graphicspath{{figures/}}
\newcommand{\researchquestion}[1]{%
  \par\smallskip
  \noindent\textbf{Research question.}\enspace\emph{#1}\par
  \smallskip}
\newcommand{\studyitem}[2]{(#1)~\textit{\uline{#2}}}

\begin{document}

\iflatexml
  \title{Elucidating the Design Space of End-to-end Latent Representation Learning in World Action Models}
  \author[1,2,3]{Junming Zhao\textsuperscript{*}}
  \author[4]{Shi Sheng\textsuperscript{*}}
  \author[1,2]{Jiacheng You}
  \author[1]{Chenxi Li}
  \author[5]{Bowen Hu}
  \author[1,2,3]{Chengbo Yuan}
  \author[3]{Yingdong Hu}
  \author[6]{Weirui Ye}
  \author[1,2,3]{Yang Gao\textsuperscript{\textdagger}}
  \affil[1]{Tsinghua University}
  \affil[2]{Shanghai Qi Zhi Institute}
  \affil[3]{Spirit.AI}
  \affil[4]{Peking University}
  \affil[5]{Zhejiang University}
  \affil[6]{Massachusetts Institute of Technology}
  \date{\textsuperscript{*}Equal contribution.\quad \textsuperscript{\textdagger}Corresponding author.\\ November 2026}
  \maketitle
  \begin{abstract}
World-Action Models (WAMs) rely on visual latent spaces for both policy perception and future prediction. Yet most prevailing WAMs inherit frozen latents from pixel VAEs or general vision encoders, which are optimized independently of action. In this paper, we explore a direct alternative: \textit{jointly learning a WAM and its latent space representation end-to-end.} We elucidate this design space from a manipulation-oriented perspective, exposing differences between embodied control and previous general visual representation learning. Through controlled studies, we derive empirical insights into shaping manipulation-relevant representations in WAM. Notably, we find that existing anti-collapse techniques are insufficient: a slow-feature shortcut in end-to-end learning impairs downstream manipulation by eroding temporal dynamics, and we introduce targeted remedies to mitigate this shortcut. We distill these findings into \textbf{E2EWAM} and scale it to 1.6B parameters using approximately 6,400 hours of robot and egocentric human data. E2EWAM outperforms frozen-latent WAMs under matched pretraining conditions. It achieves 84.46\% success on LIBERO-Plus and 56.78\% on RoboTwin 2.0 (clean-to-randomized), alongside strong real-world manipulation performance and favorable data-scaling behavior.

  \end{abstract}
  \noindent\textbf{Project page:} \url{http://e2ewam.github.io}
\else
\begin{tcolorbox}[enhanced,colback=reportbox,colframe=reportbox,boxrule=0pt,arc=9pt,
  left=16pt,right=16pt,top=16pt,bottom=14pt,width=\textwidth]
  {\reportsans\bfseries\fontsize{16.5}{20.5}\selectfont\color{reportnavy}
    \raggedright\hyphenpenalty=10000\exhyphenpenalty=10000
    Elucidating the Design Space of End-to-end Latent\\
    Representation Learning in World Action Models\par}
  \vspace{9pt}
  {\reportsans\bfseries\small\raggedright
    \mbox{Junming Zhao\textsuperscript{1,2,3,*}}\quad
    \mbox{Shi Sheng\textsuperscript{4,*}}\quad
    \mbox{Jiacheng You\textsuperscript{1,2}}\quad
    \mbox{Chenxi Li\textsuperscript{1}}\quad
    \mbox{Bowen Hu\textsuperscript{5}}\quad
    \mbox{Chengbo Yuan\textsuperscript{1,2,3}}\\[3pt]
    \mbox{Yingdong Hu\textsuperscript{3}}\quad
    \mbox{Weirui Ye\textsuperscript{6}}\quad
    \mbox{Yang Gao\textsuperscript{1,2,3,\textdagger}}\par}
  \vspace{4pt}
  {\reportsans\footnotesize\raggedright
    \textsuperscript{1}Tsinghua University\quad
    \textsuperscript{2}Shanghai Qi Zhi Institute\quad
    \textsuperscript{3}Spirit.AI\\
    \textsuperscript{4}Peking University\quad
    \textsuperscript{5}Zhejiang University\quad
    \textsuperscript{6}Massachusetts Institute of Technology\par}
  \vspace{3pt}
  {\footnotesize
    \textsuperscript{*}Equal contribution.\quad
    \textsuperscript{\textdagger}Corresponding author.\par}
  \vspace{12pt}
World-Action Models (WAMs) rely on visual latent spaces for both policy perception and future prediction. Yet most prevailing WAMs inherit frozen latents from pixel VAEs or general vision encoders, which are optimized independently of action. In this paper, we explore a direct alternative: \textit{jointly learning a WAM and its latent space representation end-to-end.} We elucidate this design space from a manipulation-oriented perspective, exposing differences between embodied control and previous general visual representation learning. Through controlled studies, we derive empirical insights into shaping manipulation-relevant representations in WAM. Notably, we find that existing anti-collapse techniques are insufficient: a slow-feature shortcut in end-to-end learning impairs downstream manipulation by eroding temporal dynamics, and we introduce targeted remedies to mitigate this shortcut. We distill these findings into \textbf{E2EWAM} and scale it to 1.6B parameters using approximately 6,400 hours of robot and egocentric human data. E2EWAM outperforms frozen-latent WAMs under matched pretraining conditions. It achieves 84.46\% success on LIBERO-Plus and 56.78\% on RoboTwin 2.0 (clean-to-randomized), alongside strong real-world manipulation performance and favorable data-scaling behavior.

  \par\vspace{12pt}
  {\small
    {\reportsans\bfseries\color{reportnavy}Date:} November 2026\par
    {\reportsans\bfseries\color{reportnavy}Project page:} \url{http://e2ewam.github.io}\par}
\end{tcolorbox}
\fi

\suppressfloats[t]

\addtocontents{toc}{\protect\setcounter{tocdepth}{-1}}

\begin{figure*}[t]
  \centering
  \includegraphics[width=\textwidth]{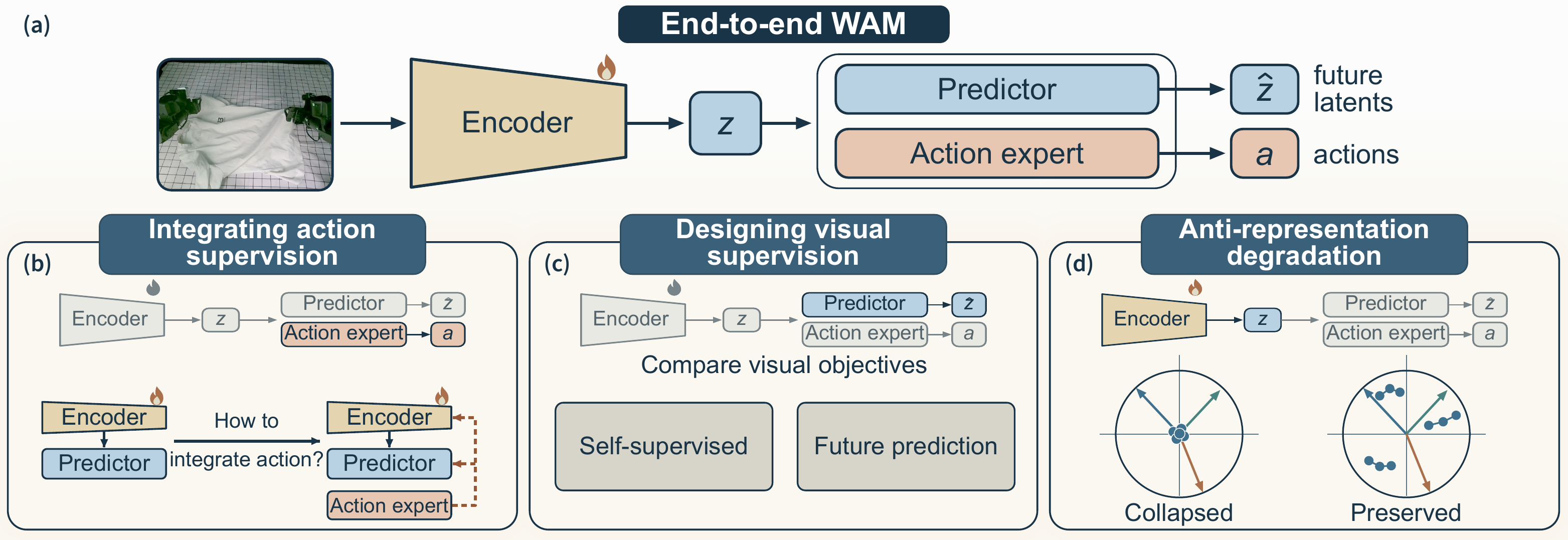}
  \caption{\textbf{End-to-end latent learning in WAMs.}
\textbf{(a)}~Overview of our progressive design framework.
\textbf{(b)}~Starting from a latent world model, how should action be integrated to shape the learned representations?
\textbf{(c)}~Given action supervision, which visual supervision more effectively encourages the learning of manipulation-relevant representations?
\textbf{(d)}~With both supervision objectives established, how can we regularize the learning process to improve representation quality?}
  \label{fig:teaser}
\end{figure*}

\section{Introduction}
\label{sec:introduction}

World-Action Models (WAMs) have recently emerged as a promising paradigm
for generalist robotic manipulation~\citep{dreamzero2026,cosmospolicy2026,
openwam2026,fastwam2026}.
Despite architectural and inference variations, prevailing WAMs share
a core training paradigm when learning from manipulation demonstrations:
beyond \emph{action prediction}, they incorporate a \emph{visual prediction}
objective for co-training.
By supervising the model to predict future frames,
this visual objective encourages the capture of world dynamics
to enhance manipulation performance.

In practice, WAMs perform visual prediction not in raw pixel space,
but within a learned latent space.
Specifically, a visual encoder maps observations into compact latent
representations that serve as both policy inputs and future prediction targets.
Consequently, the information encoded within this latent space
and its representation quality are critical to WAMs:
they directly govern how effectively the model captures physical dynamics
and executes downstream manipulation.

Despite its central role, the latent space in most WAMs is
typically determined before WAM policy training.
Most WAMs build on large-scale pretrained video generators
and predict in their VAE latent spaces~\citep{dreamzero2026,cosmospolicy2026,
genie2025,lingbotva2026,fastwam2026}.
Optimized for pixel reconstruction, these latents preserve
fine-grained visual details regardless of their relevance to manipulation.
Other WAMs instead adopt features from pretrained vision encoders
such as DINOv3 or V-JEPA~2/2.1~\citep{lyu2026lda1b,lawam2026,sun2026vlajepa,jepawam2026}.
Across these approaches, what the latent space preserves is
determined by objectives separate from WAM training.
Consequently, representation learning and WAM training
remain largely separate stages, leading to a mismatch between latent representations and robot manipulation.

Rather than passively inheriting a frozen latent space
from a separate pretraining stage,
a natural direction is to learn the visual latent representation
\textbf{end-to-end} jointly with the WAM (Fig.~\ref{fig:teaser}).
Unlike the external pretext objectives used to pretrain standalone encoders,
WAM training itself provides rich supervision directly
from manipulation demonstrations,
offering a direct signal to shape manipulation-relevant representations.
This makes end-to-end latent learning a promising alternative
and directly motivates this paper, where we aim to find a principled recipe
for developing such models.

To establish the recipe, we anchor our investigation in a single guiding principle:
\textbf{manipulation-oriented design for representation learning}.
This perspective motivates us
to rethink prevailing latent shaping methods, exposing key
differences between embodied control and general vision:
Existing latent shaping paradigms focus on pixel reconstruction fidelity
or general visual semantics.
Yet manipulation requires more: folding clothes, for instance,
demands capturing the fabric's deformable physical properties
and temporal dynamics during interaction,
rather than merely recognizing the garment or reconstructing its texture.
\textit{Most importantly}, end-to-end learning introduces representation degradation risks.
We find that existing anti-collapse techniques in general visual representation learning are insufficient: even when representation diversity is preserved,
end-to-end learning exhibits a slow-feature shortcut
that progressively erodes temporal dynamics critical for manipulation.
To counter this shrinkage, we introduce targeted regularizations
that explicitly preserve dynamic information.

Together, we distill these findings into a unified recipe: \textbf{E2EWAM}
(\textbf{E}nd-\textbf{to}-\textbf{E}nd latent learning in
\textbf{WAM}).
We further scale E2EWAM to a 1.6B-parameter WAM and
evaluate its effectiveness, using approximately 6{,}400 hours of
diverse robot and egocentric human pretraining data.
Under matched pretraining conditions,
E2EWAM outperforms WAMs using frozen VAE latents
or foundation vision features.
It achieves success rates of 56.78\% on RoboTwin 2.0
(clean-to-randomized)~\citep{robotwin2_2025}
and 84.46\% on LIBERO-Plus~\citep{liberoplus2025},
surpassing larger WAMs trained on comparable amounts of data.
As shown in Figs.~\ref{fig:pareto_libero_plus} and~\ref{fig:pareto_robotwin_cr},
E2EWAM lies on the Pareto frontier for both LIBERO-Plus and
RoboTwin~2.0 (clean-to-randomized) in terms of success rate,
model size, and pretraining data scale.
Across five real-world tasks probing distinct manipulation capabilities,
E2EWAM outperforms leading VLA and WAM baselines.
Finally, E2EWAM exhibits a log-linear data scaling trend
in validation action MSE and downstream OOD success rate.
Together, these findings suggest that scaling visual representations \emph{in-task} may be more effective for robotics than relying solely on passive self-supervision. Looking ahead, we hope our works could offer a promising pathway for learning and scaling representations in embodied intelligence.

\section{Design Space and Experimental Protocol}
\label{sec:design-space}
\begin{figure*}[!t]
  \centering
  \iflatexml\includegraphics[width=\textwidth,keepaspectratio]{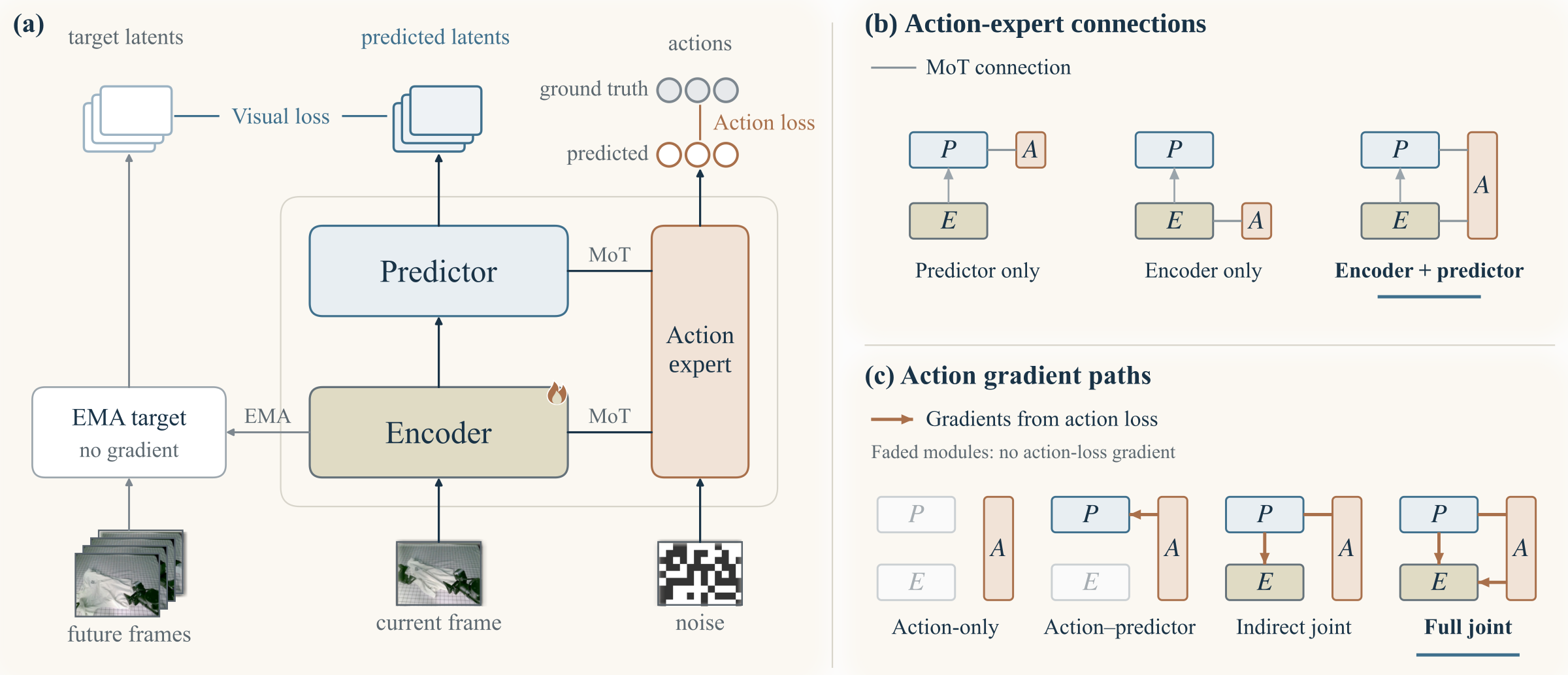}\else\includegraphics[width=\textwidth,keepaspectratio]{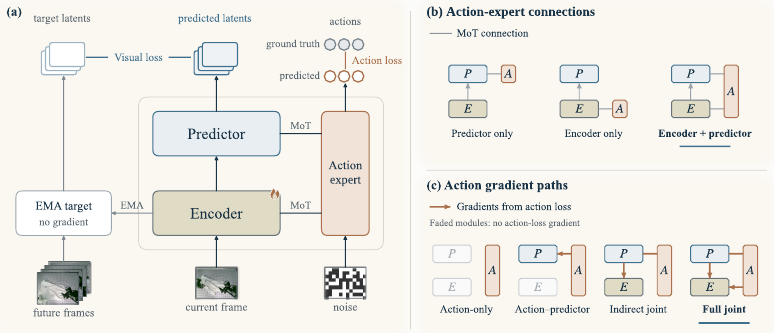}
\fi
  \caption{\textbf{Model architecture overview and action integration study.}
(a) Overview of our architecture.
(b) Action-expert connection topology.
(c) Action gradient routing. Using stop-gradient, we progressively extend action supervision from the action expert alone (\textit{Action-only}), to additionally supervising the predictor (\textit{Action--predictor}), then the encoder through the predictor (\textit{Indirect joint}), and finally the encoder directly via MoT as well (\textit{Full joint}).
Faded modules receive no action-loss gradients. Underlines indicate our final design choices.}
  \label{fig:supervision_framework}
\end{figure*}

Before presenting our investigation, we first establish our starting point
and disentangle the design space.
To learn visual representations end-to-end,
Joint-Embedding Predictive Architectures (JEPAs)~\citep{Assran_2023_CVPR}
offer a compelling paradigm.
Operating entirely in latent space, JEPAs jointly train the encoder during target prediction 
so that the learning objective guides what the representations capture.
This property makes a JEPA-style end-to-end world model~\citep{pldm2025,
lewm2026} a promising starting point.
As shown in Fig.~\ref{fig:supervision_framework}(a),
our starting world model trains an online encoder and a world predictor
\textbf{end-to-end}: given latent inputs from the online encoder
and target latents from an EMA target encoder,
the predictor initially minimizes a next-frame MSE loss
following prior works~\citep{dinowm2025,lewm2026}.

\noindent\textbf{Integrating Action Supervision.}
Our first step focuses on turning the starting JEPA world model into an end-to-end WAM. 
Doing so introduces an action expert for manipulation. We investigate how world model's representations can effectively used for action decision, and how action supervision can be incorporated into representation learning.

\noindent\textbf{Designing the Visual Supervision Objective.}
Once action supervision is integrated, the next question is
what the visual supervision objective itself should target.
Existing paradigms in visual representation learning and world modeling
offer rich precedents, either by shaping general visual semantic features
through self-supervised objectives~\citep{byol2020,dino2021,mae2022,Assran_2023_CVPR,vjepa2024,vjepa2_2025}
or by structuring latent spaces via future state
prediction~\citep{genie2025,lewm2026}.
However, in end-to-end WAMs, the visual objective operates alongside action supervision and targets embodied control rather than general vision. With action supervision’s coapperance, it remains an open question how to formulate this visual objective so that it guides
the latent representation to encode manipulation-relevant information.
\textbf{Regularizing Against Representation Degradation.}
Having established the supervision objectives, we now turn to another core aspect of end-to-end representation learning: regularizing the learning process itself.
Unlike WAMs built on frozen latents, end-to-end latent learning allows the representation serving as the visual prediction target to evolve during training.
This introduces a fundamental risk: the model may reduce prediction difficulty by discarding information from the latent space rather than learning to predict it.
In the extreme case, distinct observations are mapped to nearly identical features---a longstanding challenge known as \emph{representation collapse}~\citep{Assran_2023_CVPR,vjepa2024,lewm2026}.
To address this, our third design axis \textbf{(i)} examines the role and effectiveness of existing anti-collapse techniques in end-to-end latent WAM training. 
However, we further find that existing anti-collapse techniques alone are insufficient.
Even when the regularization successfully maintains representation diversity across various scenes,  it can still progressively lose temporal dynamic information as the latent space evolves during training.
This reveals a more subtle failure mode beyond conventional representation collapse.
We therefore \textbf{(ii)} investigate this temporal dynamics shrinkage and develop a corresponding regularization strategy to mitigate it.

\noindent\textbf{Controlled Evaluation Protocol.} For the studies in Section~\ref{sec:roadmap}, we adopt a unified pretraining-and-finetuning pipeline with a small-scale end-to-end WAM initialized from scratch. 
We construct a $245$-hour pretraining corpus by performing task-stratified sampling across six open-source robot datasets. 
Each variant is pretrained on this shared corpus and fine-tuned on LIBERO~\citep{liu2023libero}. We evaluate the resulting policies on standard LIBERO suites for in-domain manipulation capabilities, and on LIBERO-Plus~\citep{liberoplus2025} for zero-shot OOD generalization across seven shift dimensions (background, camera, language, lighting, object layout, robot initialization, and sensor noise). 
Across all experiments, we fix the backbone architecture and training budget, varying only one design factor per step. For each variant, we sweep its critical hyperparameters to report its best-performing configuration, deferring detailed sweep analyses and ablation findings to Appendix~\ref{app:sweeps}.
\providecommand{\reportlayout}[2]{#2}

\section{Building End-to-End Latent World-Action Model}
\label{sec:roadmap}
In this section, we progressively build an end-to-end latent WAM
through four steps organized around the design axes introduced above.

\subsection{Integrating Action Supervision.}
\label{sec:roadmap-supervision}

Following the design axes in Section~\ref{sec:design-space},
we start from a JEPA end-to-end latent world model.
Our first step focuses on integrating action supervision
into this framework (Fig.~\ref{fig:supervision_framework}(a)),
which raises the following research question:

\researchquestion{How can we effectively integrate action prediction
into end-to-end WAM training?
How does this supervision shape the learned representations?}

This integration introduces an action expert in representation training:
In the forward pass, the action expert reads out the representations
learned by the world model.
We first investigate which representations the action expert should condition on.
Adopting the Mixture-of-Transformers (MoT) architecture commonly used in WAMs,
we compare three connection topologies: predictor only, encoder only, and both,
while keeping model size and depth fixed
(Fig.~\ref{fig:supervision_framework}(b)).
In the backward pass, action loss gradients can propagate into the world model
and shape its representations.
We investigate which components of the world model
should receive action supervision via gradient routing within a unified topology
(Fig.~\ref{fig:supervision_framework}(c)).

\begin{figure*}[!tbp]
  \centering
  \includegraphics[width=\textwidth]{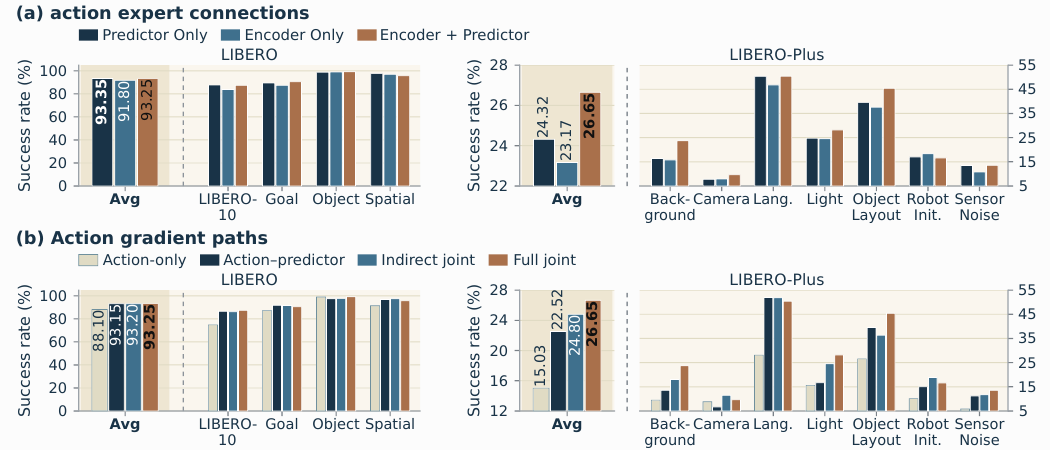}
  \caption{\textbf{Integrating action supervision.}
  \textbf{(a) Action-expert connections.} The action expert conditions on
  features from the predictor alone, the encoder alone, or both.
  \textbf{(b) Action gradient paths.} With both connections in place, we compare
  the four routing variants shown in Fig.~\ref{fig:supervision_framework}(c).}
  \label{fig:supervision-allocation}
\end{figure*}

\textbf{Action expert should condition on both encoder and predictor features.}
We evaluate the three condition topologies
in Fig.~\ref{fig:supervision_framework}(b).
As shown in Fig.~\ref{fig:supervision-allocation}(a),
connecting the action expert to both the encoder and predictor
achieves the highest OOD average success rate on LIBERO-Plus.
We therefore adopt this joint conditioning topology
in all subsequent experiments.

\textbf{Broader action supervision improves OOD robustness.}
With the action-conditioning topology fixed,
we further investigate which model components should receive action supervision.
Specifically, via stop-gradient
(Fig.~\ref{fig:supervision_framework}(c)),
we progressively route action gradients into more model components:
(i) blocking them from the latent world model (\textit{Action-only}),
(ii) supervising only the predictor (\textit{Action-predictor}),
(iii) further backpropagating through the predictor into the encoder
(\textit{Indirect joint}),
and (iv) adding direct encoder supervision via MoT (\textit{Full joint}).
As shown in Fig.~\ref{fig:supervision-allocation}(b),
progressively broader and more direct action supervision
monotonically improves OOD robustness on LIBERO-Plus.
This is likely because action supervision encourages the model
to focus on action-relevant representations,
mitigating the impact of task-irrelevant visual distractors.

\FloatBarrier

\subsection{Designing the Visual Supervision Objective}
\label{sec:roadmap-prediction}
Building on the architecture and supervision allocation established
in Section~\ref{sec:roadmap-supervision},
we next investigate the choice of visual supervision objective.
Prevailing WAMs commonly inherit future frame prediction objectives
directly from video generation paradigms~\citep{dreamzero2026,
cosmospolicy2026, lingbotva2026, fastwam2026}.
Meanwhile, the broader latent representation learning literature offers
diverse self-supervised objectives~\citep{dino2021, mae2022, Assran_2023_CVPR,
vjepa2024}.
However, these objectives are neither originally designed nor optimized for
robotic manipulation, and their effectiveness under combined action and visual
supervision remains unexplored.
This raises the open problem in this section:

\researchquestion{Which visual supervision more effectively encourages
the learning of manipulation-relevant representations?}

To ensure a controlled and fair comparison, we keep the
action prediction inputs and outputs, as well as the training
budget, identical across variants, varying only the visual
supervision objective.
Moreover, we sweep key hyperparameters for each objective variant
and select the best configuration.
Detailed formulations, implementation choices, and tuning findings
are provided in Appendix~\ref{app:visual_objectives}.

\begin{figure*}[!t]
  \centering
  \iflatexml\includegraphics[width=\textwidth,keepaspectratio]{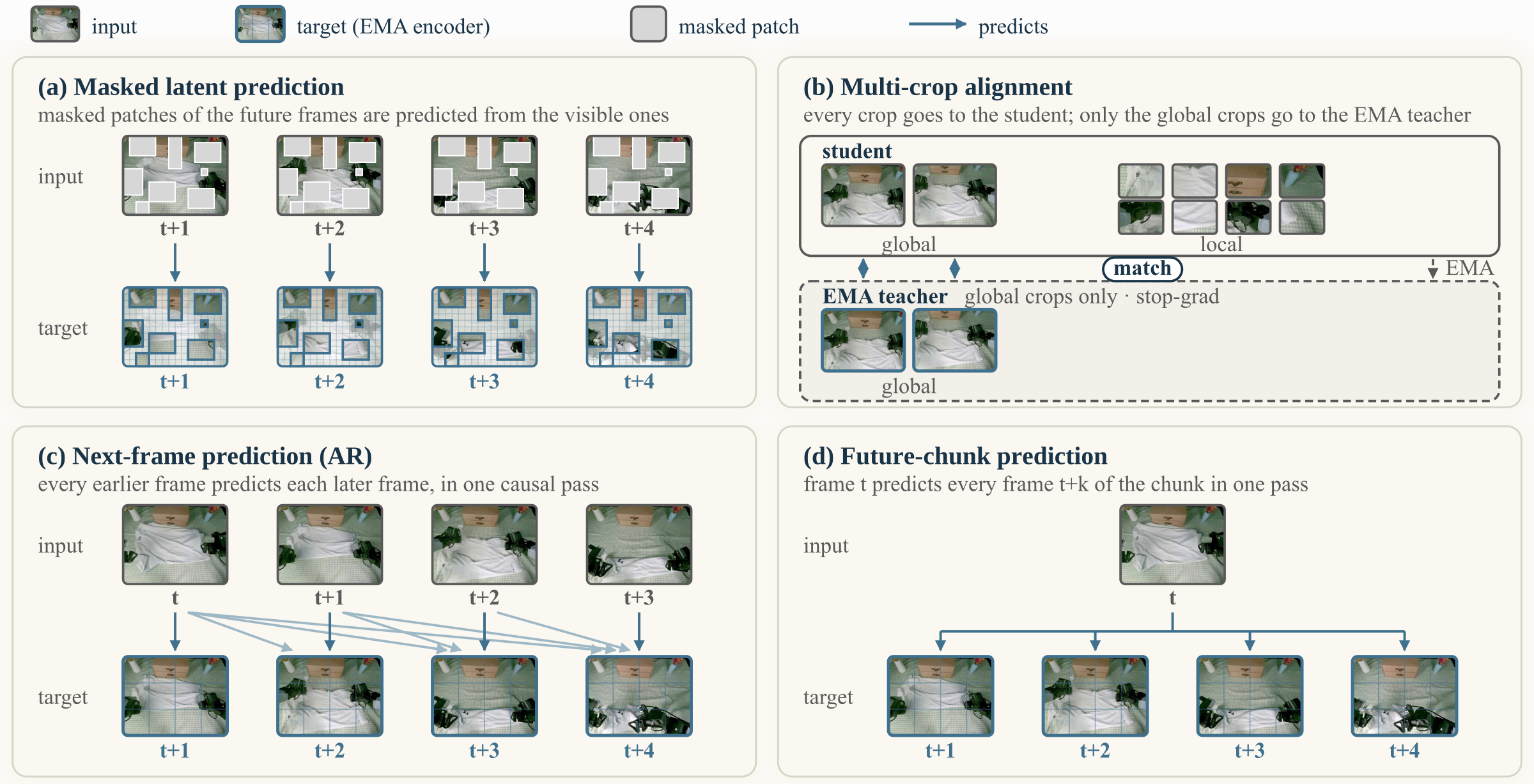}\else\includegraphics[width=\textwidth,keepaspectratio]{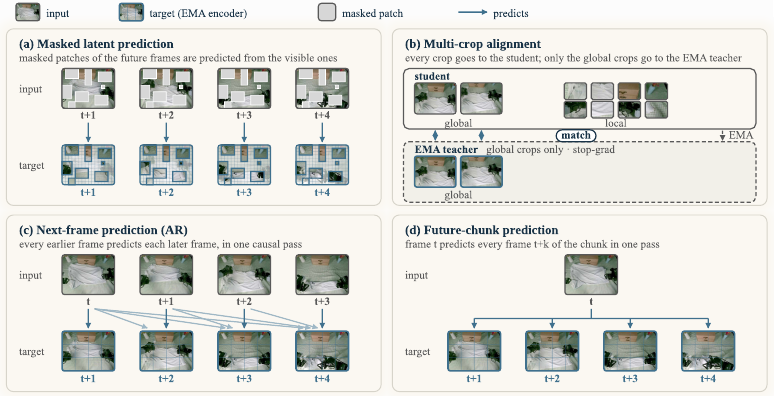}
\fi
  \caption{\textbf{Visual supervision objectives in
  Section~\ref{sec:roadmap-prediction}.}
  \textbf{(a)} Masked latent prediction predicts masked latent patches.
  \textbf{(b)} Multi-crop alignment matches student crops to an EMA teacher.
  \textbf{(c)} Next-frame prediction (AR) uses observed latent prefixes.
  \textbf{(d)} Future-chunk prediction jointly predicts four future frames
  from the same current visual latent.}
  \label{fig:visual_objective}
\end{figure*}

\studyitem{i}{Masked latent prediction.}
Following V-JEPA~\citep{vjepa2024},
this variant predicts masked spatiotemporal features of a future
clip from its visible context (Fig.~\ref{fig:visual_objective}(a)).
An EMA encoder provides the target features at the masked positions.

\studyitem{ii}{Multi-crop alignment.}
Multi-crop alignment encourages semantic consistency
across differently augmented views of the same scene,
a paradigm that has demonstrated effectiveness
in the DINO families~\citep{dino2021,oquab2023dinov2,simeoni2025dinov3}
and LeVJEPA~\citep{kuhn2026levjepa}.
For each future clip, we align student outputs from global and local
crops with EMA-teacher outputs from global crops
(Fig.~\ref{fig:visual_objective}(b)).
The objective and implementation details are provided
in Appendix~\ref{app:visual_objectives}.

\studyitem{iii}{Next-frame prediction (AR).}
Following LeWorldModel~\citep{lewm2026} and DINO-WM~\citep{dinowm2025},
this variant learns one-step latent transitions.
Under teacher forcing, each future frame is predicted from
the latent representations of preceding observed frames
and the action segments connecting them to the target time.
The observed visual prefix therefore grows with the prediction horizon.
We supervise $K=4$ next-frame predictions in a single causal
forward pass (Fig.~\ref{fig:visual_objective}(c)),
matching the per-batch frame budget of the other variants.

\studyitem{iv}{Future-chunk prediction.}
Inspired by video-generation WAMs~\citep{dreamzero2026,cosmospolicy2026,
genie2025,fastwam2026},
this variant jointly predicts $K=4$ future frames from
the same current-frame latent.
Each prediction is conditioned on the action segments up to
its target time, without using intermediate future observations
as visual inputs (Fig.~\ref{fig:visual_objective}(d)).
Both next frame prediction and future chunk prediction use the same MSE
loss against EMA-encoded targets
and supervise the same future frames in a single causal forward pass,
isolating the effect of visual conditioning.
Detailed formulations are provided in
Appendix~\ref{app:visual_objectives}.

\begin{figure*}[!t]
  \centering
  \includegraphics[width=\textwidth]{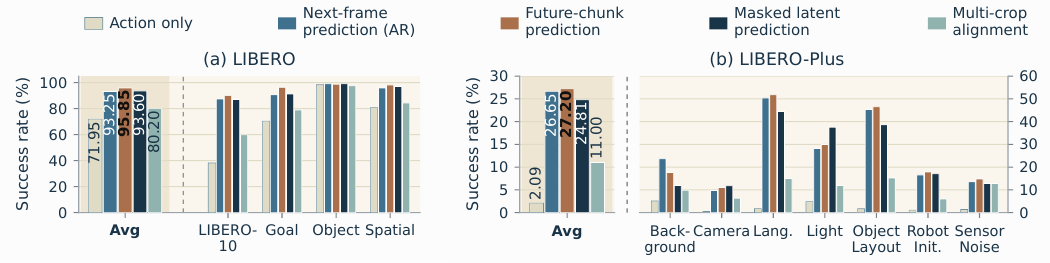}
  \caption{\textbf{Designing the visual supervision objective.}
  We compare the four variants in Section~\ref{sec:roadmap-prediction}
  with a baseline trained solely on the action loss, without a visual prediction term.}
  \label{fig:prediction-target}
\end{figure*}

\textbf{Future-chunk prediction is well suited to robotic manipulation.}
As shown in Fig.~\ref{fig:prediction-target}, incorporating visual supervision
generally improves manipulation performance over the action-only baseline.
Overall, the two action-conditioned future prediction variants compare
favorably with masked latent prediction and multi-crop alignment.
Future-chunk prediction achieves the highest average success rates, improving
from 71.95\% to 95.85\% on LIBERO and from 2.09\% to 27.20\% on LIBERO-Plus
over the action-only baseline.
Physical-state probing
(Appendix~\ref{app:probing-visual},
Table~\ref{tab:prediction-target-probing})
further supports this finding, showing that future prediction
yields representations that better preserve
manipulation-relevant information.
These results suggest that modeling action-conditioned temporal dynamics is
more relevant to robotic manipulation than inferring masked features or
capturing view-invariant semantics.

\begin{figure*}[!t]
  \centering
  \includegraphics[width=\textwidth]{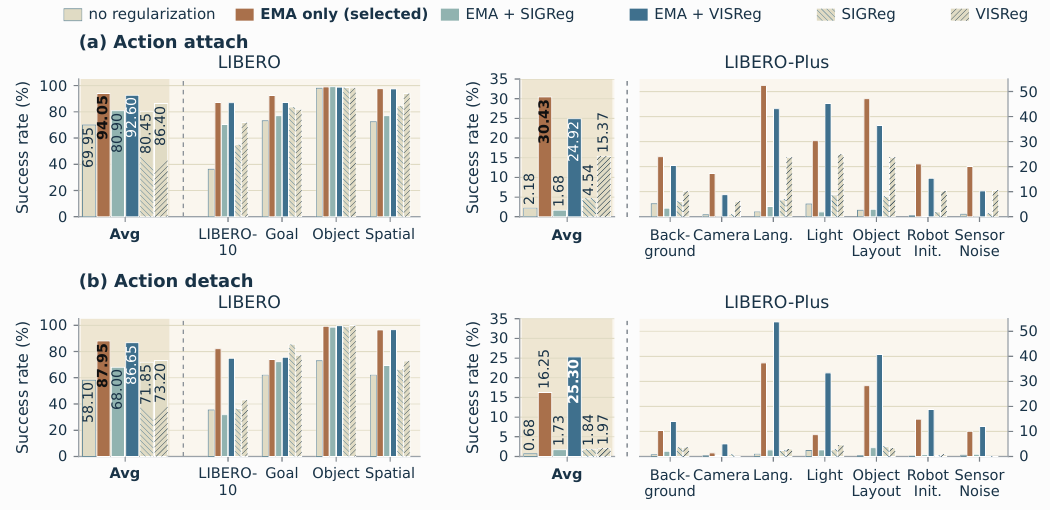}
  \caption{\textbf{Regularization against representation collapse.}
  Comparison of the anti-collapse variants (EMA, SIGReg, VISReg, and their
  combinations) against a baseline without any anti-collapse technique
  (\textit{no regularization}), evaluated under two gradient settings:
  \textbf{(a) Action attach}, which routes action supervision into the world
  model following the design from preceding steps; and \textbf{(b) Action detach},
  which blocks action-loss gradients from the world model.}
  \label{fig:collapse-regularization}
\end{figure*}

\textbf{Future-Chunk Prediction Formulation.}
Building on the effectiveness of future-chunk prediction,
we further investigate how to formulate its prediction objective.
We compare direct MSE regression with two flow-matching variants
that predict either velocity ($v$) or clean latents ($x_0$)
from noisy future latents.
We find that velocity prediction at a noise shift of $s=24$
achieves the best performance.
Details are provided in Appendix~\ref{app:visual_objectives}.

\subsection{Regularization Against Representation Collapse}
\label{sec:roadmap-collapse}

Having established the representation supervision objectives,
we next turn to regularizing the training process
to stabilize representation quality.
While prior self-supervised representation learning has established
a rich suite of anti-collapse techniques~\citep{byol2020,
bardes2022vicreg, Assran_2023_CVPR, lewm2026},
these regularizers were originally designed
for general self-supervised visual learning.
How they perform in end-to-end WAM training remains largely unexplored.
This motivates our central question:
\researchquestion{How effective are existing anti-collapse regularizations
in end-to-end latent WAM training?}

To answer this question, we investigate
two primary families of anti-collapse techniques
under a controlled training setup:

\studyitem{i}{Exponential Moving Average (EMA)}~\citep{moco2020,byol2020,vjepa2024}
has long been used in self-supervised learning
to maintain a slowly updated target encoder.
By slowing the evolution of prediction targets,
EMA stabilizes representation learning and mitigates the risk of collapse.

\studyitem{ii}{Isotropic Gaussian Regularization}~\citep{wu2026visreg,
lewm2026, kuhn2026levjepa}.
This class of methods explicitly regularizes
the distribution of encoded latents toward a standard isotropic Gaussian,
$\mathcal{N}(\mathbf{0},\mathbf{I})$, to prevent representation collapse.
We investigate two representative methods,
SIGReg~\citep{balestriero2025lejepa} and VISReg~\citep{wu2026visreg},
which differ in how they enforce this distributional constraint.
Implementation details
and hyperparameter optimization findings are provided
in Appendix~\ref{app:collapse_details}.

\reportlayout{\newpage}{}
\providecommand{\reportlayout}[2]{#2}
\reportlayout{\begin{wraptable}{r}{0.43\textwidth}}{\begin{table}[t]}
  \centering
  \caption{\textbf{Encoder effective rank under anti-collapse strategies.}
  \textit{Action attached} allows action-loss gradients to reach the world
  model, and \textit{Action detached} blocks that gradient path.
  Effective rank measures how evenly singular values are distributed;
  higher values do not necessarily imply more informative representations,
  whereas very low values indicate collapse.}
  \label{tab:regularization-rankme}
  \begingroup
  \setlength{\tabcolsep}{2.2pt}
  \renewcommand{\arraystretch}{1.1}
  \setlength{\aboverulesep}{1.4pt}
  \setlength{\belowrulesep}{1.4pt}
  \fontsize{8}{9.5}\selectfont
  \begin{tabular*}{\linewidth}{@{}l@{\extracolsep{\fill}}rr@{}}
    \toprule
    Method & Action attached & Action detached \\
    \midrule
    no regularization & 1.034 & 1.006 \\
    EMA only & 76.784 & 69.579 \\
    EMA + SIGReg & 44.266 & 19.923 \\
    EMA + VISReg & 46.486 & 63.878 \\
    SIGReg & 76.821 & 48.085 \\
    VISReg & 10.986 & 65.157 \\
    \bottomrule
  \end{tabular*}
  \endgroup
\reportlayout{\end{wraptable}}{\end{table}}

\textbf{EMA stabilizes end-to-end WAM training
and outperforms isotropic Gaussian regularization.}
Table~\ref{tab:regularization-rankme} reports
the RankMe effective rank~\citep{garrido2023rankme}
of encoder representations under different anti-collapse strategies.
Without regularization, the effective rank drops to nearly one,
indicating severe representation collapse.
In contrast, all evaluated anti-collapse techniques increase
the effective rank, successfully mitigating this failure mode.
However, these methods differ in their impact
on downstream manipulation performance.
As shown in Fig.~\ref{fig:collapse-regularization}(a),
EMA alone consistently outperforms isotropic Gaussian regularization
(SIGReg and VISReg).
Compared with no regularization, EMA improves average success rates
from $69.95\%$ to $94.05\%$ on LIBERO
and from $2.18\%$ to $30.43\%$ on LIBERO-Plus.

\textbf{Adding VISReg is more effective without action supervision.}
Isotropic Gaussian regularization has proven effective
in self-supervised representation learning~\citep{balestriero2025lejepa,
wu2026visreg,kuhn2026levjepa}.
Whereas in end-to-end WAM training, action supervision directly shapes the latent.
We therefore examine whether action supervision explains
why this regularization underperforms EMA in our setting.
Specifically, we detach the action supervision gradient,
leaving the latent to be shaped solely by self-supervised visual prediction.
As shown in Fig.~\ref{fig:collapse-regularization}, adding VISReg to EMA
improves LIBERO-Plus success from 16.25\% to 25.30\% when action gradients are
detached, but reduces it from 30.43\% to 24.92\% when they are attached.
This contrast reveals a possible explanation: without an explicit downstream
task signal, matching a predefined Gaussian distribution helps improve
representation quality; once task-specific supervision such as the action loss
is introduced, however, the Gaussian prior may conflict with task-driven
supervision.

\subsection{Regularization Against Slow Feature Shortcut}

\reportlayout{\begin{wrapfigure}{r}{0.42\textwidth}}{\begin{figure}[!htbp]}
  \centering
  \includegraphics[width=\reportlayout{\linewidth}{0.7\linewidth}]{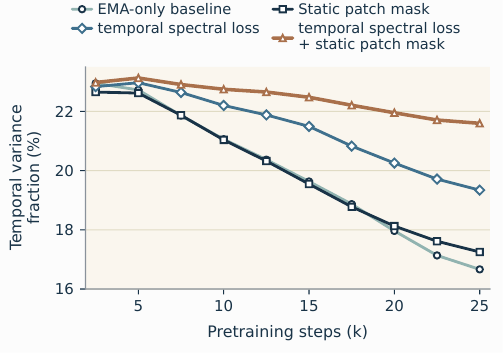}
  \caption{Temporal variance fraction of encoder patch features
  during pretraining,
  evaluated on validation clips containing \mbox{$T=16$} frames each.}
  \label{fig:temporal-variance-fraction}
\reportlayout{\end{wrapfigure}}{\end{figure}}

As established in Section~\ref{sec:roadmap-collapse},
existing anti-collapse regularizations maintain representation diversity
across diverse scenes.
We ask a further question: does preserving overall representation diversity
genuinely guarantee that manipulation-relevant information is retained?
We conduct a diagnostic experiment on a held-out validation dataset
to examine how latent diversity is distributed between
static appearance and temporal dynamics.
Specifically, we quantify the fraction of
total latent variance attributable to temporal variation:
\begin{equation}
  R_{\mathrm{temporal}}
  = \frac{\mathbb{E}_{N}\!\left[\operatorname{Var}_{T}(z)\right]}
         {\operatorname{Var}_{N,T}(z)}.
  \label{eq:temporal_variance_ratio}
\end{equation}
where $N$ indexes patch trajectories and $T$ indexes frames within each trajectory. Here, $\mathrm{Var}_{N,T}(z) = \mathbb{E}_N[\mathrm{Var}_T(z)] + \mathrm{Var}_N(\mathbb{E}_T[z])$ decomposes total variance into within-trajectory temporal variance and cross-trajectory static variance. 
For the EMA-only baseline (Fig.~\ref{fig:temporal-variance-fraction}),
$R_{\mathrm{temporal}}$ quickly shrinks as pretraining progresses.
This reveals a failure mode beyond conventional representation collapse:
latent diversity increasingly concentrates on time-invariant static features
while progressively losing sensitivity to temporal dynamics—a phenomenon
called slow feature observed in prior studies~\citep{sobal2022slow,vanassel2026spectral}.
Yet, existing regularization methods have largely focused
on preventing representation collapse.
Effective remedies for this temporal degradation
and its impact on robotic manipulation still remain underexplored.
This motivates the open problem in this section:

\researchquestion{How can we preserve temporal dynamics information
in end-to-end latent learning, and to what extent does it improve
downstream manipulation?}

\begin{figure*}[!tbp]
  \centering
  \iflatexml\includegraphics[width=\textwidth]{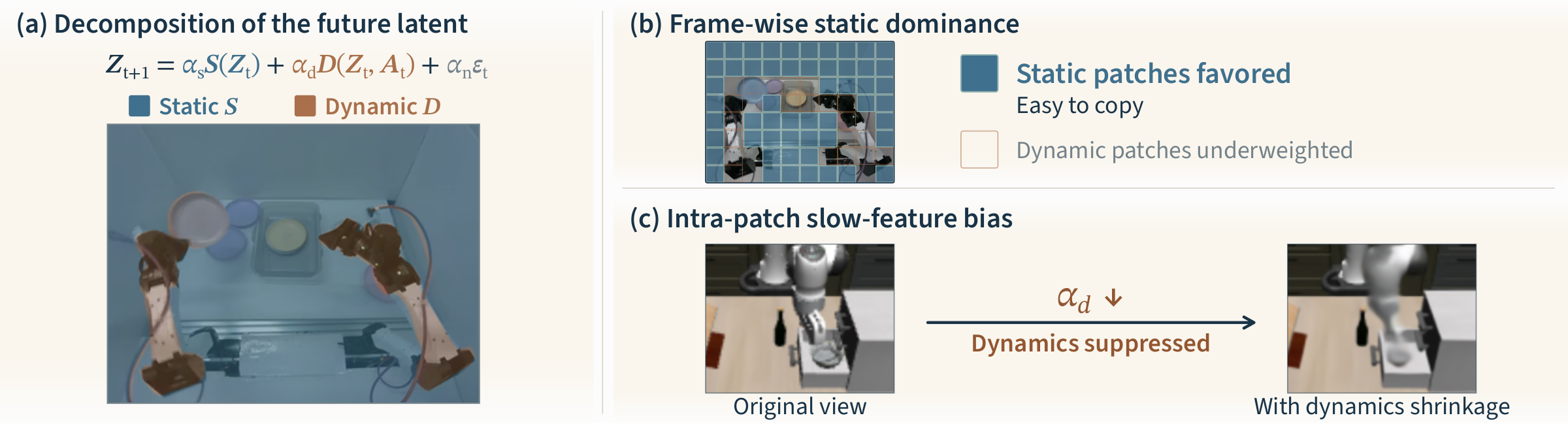}\else\includegraphics[width=\textwidth]{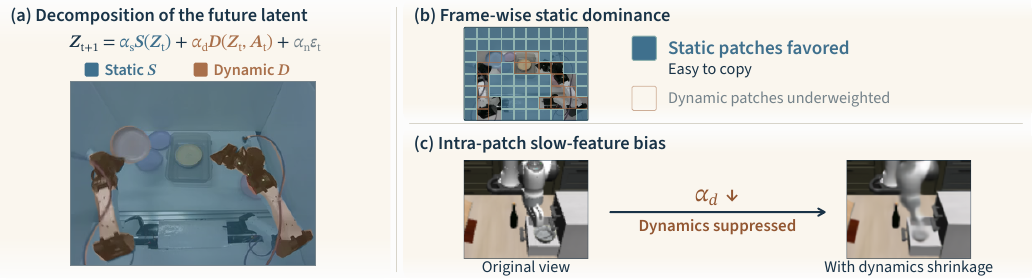}
\fi
  \caption{\textbf{Understanding the slow-feature shortcut.}
    \textbf{(a) Decomposition of the future latent} $Z_{t+1}$ into static
    features $S$, action-dependent dynamics $D$, and unpredictable noise
    $\epsilon_t$, exposing the shortcut at two levels:
    \textbf{(b) frame-wise static dominance}, where abundant static patches
    dominate the frame and dilute dynamic supervision; and
    \textbf{(c) intra-patch slow-feature bias}, within dynamic patches,
    encoder training shrinks $\alpha_d$ and erases dynamic information.}
  \label{fig:recipe4_shortcuts}
\end{figure*}

\subsubsection{Understanding Slow Feature Shortcut: A Simplified View}
\label{sec:slow-feature-analysis}

To understand the cause of this phenomenon, we consider a conceptual
decomposition of the target future latent
$Z_{t+1}$ into three components:
\begin{equation}
  Z_{t+1} = \alpha_s S(Z_t) + \alpha_d D(Z_t, A_t) + \alpha_n \epsilon_t.
  \label{eq:latent_decomp}
\end{equation}
Here, (i) $S(Z_t)$ captures static or slowly varying features, such as
persistent backgrounds and slowly changing textures; (ii) $D(Z_t, A_t)$
represents the dynamic component governed by the action $A_t$; and
(iii) $\epsilon_t$ denotes unpredictable noise.
We normalize the components in mean square:
$\mathbb{E}\|S(Z_t)\|^2 = \mathbb{E}\|D(Z_t,A_t)\|^2
= \mathbb{E}\|\epsilon_t\|^2 = 1$.
With this normalization, the nonnegative scalars $\alpha_s$, $\alpha_d$,
and $\alpha_n$ specify the scale of their respective contributions to the
latent representation.
For simplicity, we assume $\mathbb{E}[\epsilon_t \mid Z_t,A_t]=0$.
The future latent prediction objective then takes the form:
\begin{equation}
  \mathcal{L}_{\mathrm{pred}}
  = \mathbb{E}\|Z_{t+1} - f_\phi(Z_t, A_t)\|^2
  = \mathbb{E}\|\alpha_s S(Z_t) + \alpha_d D(Z_t, A_t)
      - f_\phi(Z_t, A_t)\|^2 + \alpha_n^2 \mathbb{E}\|\epsilon_t\|^2.
  \label{eq:pred_loss}
\end{equation}

Because future prediction is performed directly in the learned latent space,
$\mathcal{L}_{\mathrm{pred}}$ scales directly with the difficulty of
predicting the encoded information. Crucially, end-to-end learning allows the
target representation to evolve throughout training. This opens a shortcut:
instead of learning a better predictor $f_\phi$, the model can attenuate
the loss by
discarding the infromation that is hard to predict.
We therefore identify two complementary levels of this shortcut:

\noindent\studyitem{i}{Frame-wise static dominance.}
We partition the spatial patches $\mathcal{P}$ into nearly static patches
$\mathcal{P}_{\mathrm{stat}}$, where $D_p(Z_t,A_t)\approx 0$ (e.g., persistent
background patches), and dynamic patches $\mathcal{P}_{\mathrm{dyn}}$
capturing action-dependent changes.
In Eq.~\eqref{eq:pred_loss}, the static component $S(Z_t)$ can be trivially
predicted by copying from $Z_t$ or applying a simple transformation, whereas
predicting the dynamic component $D(Z_t,A_t)$ in $\mathcal{P}_{\mathrm{dyn}}$
requires modeling action-dependent state transitions and physical interactions.
In typical manipulation scenes (Fig.~\ref{fig:recipe4_shortcuts}(b)),
nearly static patches occupy the frame.
Since the prediction loss is averaged over all patches, these abundant
easy-to-predict static patches dilute the contribution of dynamic patches,
allowing the model to achieve a low prediction loss without adequately
modeling temporal dynamics.

\noindent\studyitem{ii}{Intra-patch slow-feature bias.}
Even within dynamic patches $\mathcal{P}_{\mathrm{dyn}}$, where
action-dependent motion is present, Eq.~\eqref{eq:pred_loss} can still be
rapidly reduced by shrinking $\alpha_d$ and $\alpha_n$.
While suppressing $\alpha_n$ may beneficially filter out unpredictable noise,
attenuating $\alpha_d$ directly erases dynamic information from the latent
space (Fig.~\ref{fig:recipe4_shortcuts}(c)).
For instance, encoding only the slowly changing color and coarse contour
of a moving robot arm while ignoring fast-changing pose information.

\noindent\textbf{Exsiting anti-collapse regularization fail to block this
shortcut.}
Consider a simplified regime with $\alpha_n = 0$,
where the static component $S$ is trajectory-invariant
and independent of the dynamic component $D_t$.
Under a degenerate copy predictor
$f_{\mathrm{copy}}(Z_t, A_t) = Z_t$,
the prediction loss simplifies to:
\begin{equation}
  \mathcal{L}_{\mathrm{copy}}(\alpha_d)
  = \mathbb{E}\|Z_{t+1} - Z_t\|^2
  = \alpha_d^2\,\mathbb{E}\|D_{t+1} - D_t\|^2
  \label{eq:copy_loss}
\end{equation}

Consequently, attenuating the dynamic weight $\boldsymbol{\alpha_d}$
directly reduces the copy prediction error, favoring the degenerate solution
$\alpha_d = 0$. Crucially, this solution still satisfies existing
anti-collapse constraints. Under \textit{isotropic Gaussian regularization},
a static-only representation $Z_t = \alpha_s S$ with marginal distribution
$\mathcal{N}(0, I)$ across scenes satisfies the distributional constraint
while attaining zero copy-prediction error. Likewise, \textit{EMA} imposes
no constraint against the copy predictor: because the static component $S$
consistently yields lower prediction residuals, the slowly evolving EMA
target still drifts along this shortcut, progressively shrinking $\alpha_d$
and filtering out temporal dynamics information.

\subsubsection{Mitigating Slow Feature Shortcut}

Guided by the analysis in Section~\ref{sec:slow-feature-analysis},
we design two complementary regularization strategies
to mitigate the static shortcut.

\textbf{Making temporal variation matter in future prediction.}
To counteract the intra-patch preference for slowly varying features,
we reformulate the future prediction objective
in the temporal frequency domain.
Specifically, given a noisy future latent chunk $\widetilde{Z}^{\,\tau}$
and the predicted velocity $\hat{V}$,
we first recover the implied clean latent $\hat{Z}$
and project the prediction residual across the $K$ future frames
onto an orthonormal Discrete Cosine Transform (DCT-II) basis
$\mathbf{U} \in \mathbb{R}^{K \times K}$:
\begin{equation}
  \reportlayout{}{\begin{aligned}}
    \hat{\mathbf{Z}}
    \reportlayout{=}{&=} \widetilde{\mathbf{Z}}^{\,\tau} - \tau\hat{\mathbf{V}},
    \reportlayout{\qquad}{\\[4pt]}
    \widetilde{\mathbf{E}}_k
    \reportlayout{=}{&=} \sum_{j=1}^{K} U_{kj}
      \left(\hat{\mathbf{Z}}_j - \bar{\mathbf{Z}}_j\right),
    \reportlayout{\qquad}{\\[4pt]}
    \mathcal{L}_{\mathrm{spec}}
    \reportlayout{=}{&=} \mathbb{E}\!\left[
      \sum_{k=0}^{K-1}\omega_k\|\widetilde{\mathbf{E}}_k\|_2^2
    \right],
  \reportlayout{}{\end{aligned}}
  \label{eq:temporal_spectral_loss}
\end{equation}
For $j=1,\ldots,K$ and $k=0,\ldots,K-1$,
the basis entries $U_{kj}$ are defined as
\begingroup
\setlength{\abovedisplayskip}{5pt}
\setlength{\belowdisplayskip}{5pt}
\begin{equation}
  U_{kj} =
  \begin{cases}
    \dfrac{1}{\sqrt{K}}, & k=0, \\[2pt]
    \sqrt{\dfrac{2}{K}}\,
      \cos\!\left[\dfrac{\pi k}{K}\left(j-\dfrac{1}{2}\right)\right],
      & k>0.
  \end{cases}
\end{equation}
\endgroup
Here, $k=0$ corresponds to the time-invariant (DC) mean component
across the prediction horizon,
whereas $k \ge 1$ captures time-varying (AC) temporal fluctuations.
We set the AC-to-DC weighting ratio to
$\omega_{k \ge 1} : \omega_0 = 2 : 1$,
which explicitly prioritizes learning temporal transitions
over static appearance from the onset of training.

As shown in Fig.~\ref{fig:temporal-variance-fraction},
replacing the standard flow-matching objective
with $\mathcal{L}_{\mathrm{spec}}$ slows the decay
of the $R_{\mathrm{temporal}}$ during pretraining.
To verify whether this preserved temporal variance
encodes physically meaningful motion,
we linearly probe the learned representations
for dynamic physical quantities.
Table~\ref{tab:temporal-spectral-dc-probing} shows that
the temporal spectral loss consistently improves dynamic readout
across both encoder latents and predicted future latents.

\begin{figure*}[!tbp]
  \centering
  \includegraphics[width=\textwidth]{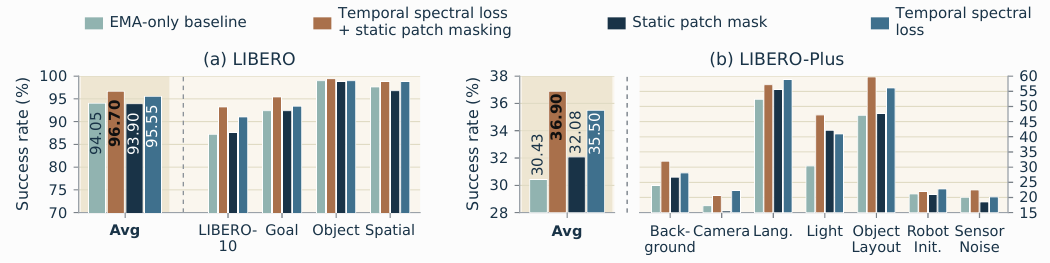}
  \caption{\textbf{Regularization against temporal dynamics shrinkage.}
    Comparison of the EMA-only baseline and variants using temporal spectral
    loss, static patch masking, or both.}
  \label{fig:temporal-spectral-dc}
\end{figure*}

In downstream manipulation (Fig.~\ref{fig:temporal-spectral-dc}),
the temporal spectral loss $\mathcal{L}_{\mathrm{spec}}$ improves
overall LIBERO-Plus success from \(30.43\%\) to \(35.50\%\).
Notably, the OOD gains are most pronounced under static visual
perturbations (background shifts, lighting changes,
and object layout distractors).
This suggests that encouraging the representation to focus
on manipulation-relevant dynamics may reduce the policy's sensitivity
to task-irrelevant static visual shifts.

\textbf{Reducing shortcuts on nearly static patches.}
To mitigate the frame-wise static dominance,
we introduce a static patch masking regularization:
let $\bar{z}_{t,p}$ denote the clean EMA target feature
of patch $p$ at step $t$, and let $\mathcal{T}_p$ index the valid steps
in its history and future latent sequence.
Its temporal mean $\mu_p$ and dynamic energy fraction $r_p$ are given by:
\begin{equation}
  \reportlayout{}{\begin{aligned}}
    \mu_p
    \reportlayout{=}{&=} \frac{1}{|\mathcal{T}_p|}\sum_{t\in\mathcal{T}_p}\bar{z}_{t,p},
    \reportlayout{\qquad}{\\}
    r_p
    \reportlayout{=}{&=} \frac{\sum_{t\in\mathcal{T}_p}\|\bar{z}_{t,p}-\mu_p\|_2^2}
                            {\sum_{t\in\mathcal{T}_p}\|\bar{z}_{t,p}\|_2^2}.
  \reportlayout{}{\end{aligned}}
\end{equation}
A smaller $r_p$ indicates that the patch feature remains closer
to a temporal constant.
By masking out nearly static patches with $r_p \le \eta$,
we define the active supervision set as:
\begin{equation}
  \mathcal{R} = \{p \in \mathcal{P} : r_p > \eta\},
\end{equation}
Combining static patch masking with the temporal spectral loss
in Eq.~\eqref{eq:temporal_spectral_loss} yields the joint objective:
\begin{equation}
  \mathcal{L}_{\mathrm{spec\text{-}mask}}
  = \mathbb{E}_{p\in\mathcal{R}}\left[
      \sum_{k=0}^{K-1}\omega_k\|\widetilde{E}_{k,p}\|_2^2
    \right],
\end{equation}
where $\widetilde{E}_{k,p}$ is the $k$-th DCT frequency residual
at patch $p$, and the expectation averages uniformly
over the retained patches $\mathcal{R}$ within each sample.
The two strategies complement each other:
static patch masking reduces supervision
from nearly static positions,
while temporal spectral loss emphasizes temporal variations
within the retained patches.

As shown in Fig.~\ref{fig:temporal-variance-fraction},
combining static patch masking with $\mathcal{L}_{\mathrm{spec}}$
further mitigates temporal dynamics shrinkage.
In dynamic physical-state probing (Table~\ref{tab:temporal-spectral-dc-probing}),
combining both regularizations achieves the best overall accuracy.
These representation gains also translate into stronger
OOD manipulation performance (Fig.~\ref{fig:temporal-spectral-dc}):
combining both strategies achieves the highest OOD success rate
of 36.90\% on LIBERO-Plus.

\begin{table*}[!htbp]
\centering
\caption{\textbf{Velocity linear probing across temporal regularization variants.}
Linear $R^2\!\uparrow$.}
\label{tab:temporal-spectral-dc-probing}
\begingroup
\setlength{\tabcolsep}{3pt}
\setlength{\aboverulesep}{1.4pt}
\setlength{\belowrulesep}{1.4pt}
\fontsize{8}{9}\selectfont
\newcommand{\probeCell}[2]{#1\,\ensuremath{\pm}\,#2}
\newsavebox{\temporalProbingTable}
\iflatexml\else
\begin{lrbox}{\temporalProbingTable}
\fi
\begin{tabular}{@{}llcccc@{}}
\toprule
 & & \multicolumn{2}{c}{Encoder latents} & \multicolumn{2}{c@{}}{Predicted latents} \\
\cmidrule(lr){3-4}\cmidrule(lr){5-6}
Property & Variant & LIBERO & LIBERO-Plus & LIBERO & LIBERO-Plus \\
\midrule
\multirow[c]{4}{*}{\shortstack[l]{EEF\\velocity}} & EMA-only baseline
  & \probeCell{0.8866}{0.0034} & \probeCell{0.7557}{0.0028}
  & \probeCell{0.7250}{0.0043} & \probeCell{0.5431}{0.0044} \\
 & static patch masking
  & \probeCell{\textbf{0.8989}}{\textbf{0.0011}} & \probeCell{0.7780}{0.0039}
  & \probeCell{0.7481}{0.0036} & \probeCell{0.5647}{0.0058} \\
 & temporal spectral loss
  & \probeCell{0.8850}{0.0031} & \probeCell{0.7636}{0.0060}
  & \probeCell{0.7415}{0.0021} & \probeCell{0.5722}{0.0020} \\
 & joint objective
  & \probeCell{0.8961}{0.0046} & \probeCell{\textbf{0.7821}}{\textbf{0.0046}}
  & \probeCell{\textbf{0.7596}}{\textbf{0.0040}} & \probeCell{\textbf{0.5907}}{\textbf{0.0048}} \\
\midrule
\multirow[c]{4}{*}{\shortstack[l]{Arm joint\\velocity}} & EMA-only baseline
  & \probeCell{0.7184}{0.0101} & \probeCell{0.5868}{0.0056}
  & \probeCell{0.5643}{0.0090} & \probeCell{0.4289}{0.0110} \\
 & static patch masking
  & \probeCell{0.7371}{0.0067} & \probeCell{0.6024}{0.0069}
  & \probeCell{0.5841}{0.0083} & \probeCell{0.4397}{0.0121} \\
 & temporal spectral loss
  & \probeCell{0.7236}{0.0088} & \probeCell{0.5969}{0.0065}
  & \probeCell{0.5817}{0.0039} & \probeCell{0.4399}{0.0088} \\
 & joint objective
  & \probeCell{\textbf{0.7415}}{\textbf{0.0064}} & \probeCell{\textbf{0.6051}}{\textbf{0.0052}}
  & \probeCell{\textbf{0.5884}}{\textbf{0.0046}} & \probeCell{\textbf{0.4525}}{\textbf{0.0119}} \\
\midrule
\multirow[c]{4}{*}{\shortstack[l]{Gripper joint\\velocity}} & EMA-only baseline
  & \probeCell{0.8545}{0.0166} & \probeCell{0.7802}{0.0062}
  & \probeCell{0.6694}{0.0269} & \probeCell{0.4748}{0.0073} \\
 & static patch masking
  & \probeCell{0.8673}{0.0121} & \probeCell{0.8027}{0.0081}
  & \probeCell{0.6642}{0.0308} & \probeCell{0.5069}{0.0046} \\
 & temporal spectral loss
  & \probeCell{0.8663}{0.0095} & \probeCell{0.8079}{0.0080}
  & \probeCell{0.6689}{0.0294} & \probeCell{0.5293}{0.0084} \\
 & joint objective
  & \probeCell{\textbf{0.8695}}{\textbf{0.0103}} & \probeCell{\textbf{0.8156}}{\textbf{0.0041}}
  & \probeCell{\textbf{0.6834}}{\textbf{0.0271}} & \probeCell{\textbf{0.5379}}{\textbf{0.0037}} \\
\midrule
\multirow[c]{4}{*}{\shortstack[l]{Manipulated\\object velocity}} & EMA-only baseline
  & \probeCell{0.8442}{0.0070} & \probeCell{0.7570}{0.0069}
  & \probeCell{0.7840}{0.0065} & \probeCell{0.6892}{0.0117} \\
 & static patch masking
  & \probeCell{0.8521}{0.0103} & \probeCell{0.7671}{0.0066}
  & \probeCell{0.7894}{0.0089} & \probeCell{0.6891}{0.0054} \\
 & temporal spectral loss
  & \probeCell{0.8484}{0.0078} & \probeCell{0.7628}{0.0055}
  & \probeCell{0.7910}{0.0131} & \probeCell{0.7017}{0.0057} \\
 & joint objective
  & \probeCell{\textbf{0.8547}}{\textbf{0.0060}} & \probeCell{\textbf{0.7762}}{\textbf{0.0062}}
  & \probeCell{\textbf{0.7925}}{\textbf{0.0081}} & \probeCell{\textbf{0.7143}}{\textbf{0.0064}} \\
\bottomrule
\end{tabular}
\iflatexml\else
\end{lrbox}
\usebox{\temporalProbingTable}
\fi
\par\vspace{2pt}
\iflatexml\else
\begin{minipage}{\wd\temporalProbingTable}
\fi
\fontsize{7}{8.5}\selectfont\raggedright
The joint objective combines temporal spectral loss with static patch masking.
Each cell reports the mean $\pm$ sample SD of pooled $R^2$ over five episode
splits. Bold indicates the best result per property, latent type, and domain.
\iflatexml\else
\end{minipage}
\fi
\endgroup
\end{table*}

\FloatBarrier

        \newenvironment{exptable}{%
          \begin{table}[!ht]
          \centering\small
        }{%
          \end{table}
        }

        \section{Empirical Evaluation at Scale}
        \label{sec:experiments}
So far, all experiments in Section~\ref{sec:roadmap} have focused on small-scale internal comparisons across design variants. To further validate the effectiveness of our method,
we scale E2EWAM to 1.6B parameters and pretrain it on approximately
6,400 hours of diverse robot and egocentric human data.
Through extensive simulation and real-robot evaluations,
we aim to answer two key questions:
(i) does an end-to-end latent WAM trained with our recipe
outperform WAMs built on frozen pretrained latent spaces?
(ii) How does E2EWAM's performance scale as pretraining data increases?

\subsection{Settings}
\label{sec:experiment-settings}
\noindent\textbf{Pretraining.}
We pretrain a 1.6B-parameter E2EWAM on approximately
6,400 hours of data from six open-source robot datasets
(DROID~\citep{khazatsky2024droid},
BC-Z~\citep{jang2022bcz},
RT-1~\citep{brohan2023rt1},
MolmoAct2-BimanualYAM~\citep{fang2026molmoact2},
Bridge~\citep{walke2023bridgedata},
and ABC-130K~\citep{allshire2026abc})
and two egocentric human datasets
(EgoDex~\citep{hoque2025egodex}
and EgoVerse~\citep{punamiya2026egoverse}).
We initialize the visual encoder from pretrained V-JEPA 2.1
weights and train the entire model end-to-end for 125K steps
with a global batch size of 1,024.

\noindent\textbf{Controlled comparison with frozen latent WAMs.}
Beyond comparing against existing state-of-the-art WAMs,
we conduct a controlled study to rigorously evaluate whether
our end-to-end recipe outperforms WAMs built on frozen latent spaces.
Specifically, we retain the predictor and action expert of E2EWAM,
replacing only the visual encoder with a frozen pretrained foundation model.
All variants are pretrained on identical data under the same training budget,
reducing confounding factors from model architecture and training.
We evaluate three widely used frozen encoders:
Wan2.2 VAE, V-JEPA 2.1, and DINOv3.
For each latent space, we tune critical hyperparameters,
including noise settings and loss weights,
to ensure that no baseline suffers from suboptimal configurations
(Appendix~\ref{app:frozen_details}).

\noindent\textbf{Simulation Evaluation Protocol}.
We evaluate E2EWAM on three widely used robotic manipulation benchmarks:
LIBERO, LIBERO-Plus, and RoboTwin 2.0.
For LIBERO-Plus, we follow the official protocol of fine-tuning on LIBERO
and evaluating directly on LIBERO-Plus to assess out-of-distribution (OOD)
generalization across seven perturbation dimensions.
For RoboTwin 2.0, we follow the official clean-only fine-tuning protocol
and evaluate under both clean (C→C) and randomized (C→R) conditions
to measure in-distribution performance and OOD generalization, respectively.
We compare against results from official benchmark leaderboards
and recent state-of-the-art VLA and WAM papers,
including only results obtained under the same evaluation protocols.

\noindent\textbf{Physical Probing.}
To assess how well the latent space encodes manipulation-relevant information,
we replay trajectories from LIBERO, LIBERO-Plus, and RoboTwin 2.0 (Randomized)
in simulation and collect privileged physical states,
including poses and velocities of the robot arm and gripper,
as well as positions and velocities of manipulated objects.
We apply linear probing to measure how accurately
each latent space reads out these quantities.
For a fair comparison, we evaluate the pretrained E2EWAM encoder directly
without fine-tuning, ensuring that none of the encoders
have seen these benchmark datasets.
We compare E2EWAM against a pixel VAE (Wan2.2 VAE),
foundation visual representations (V-JEPA 2.1 and DINOv3),
and robotic visual encoders, including R3M~\citep{r3m2022},
VC-1~\citep{vc1_2023}, and CAIP~\citep{sharma2026caip}.

\subsection{Simulation Benchmark Results}

\newcommand{\controlledstudyfigure}{%
  \reportlayout{\begin{wrapfigure}{r}{0.50\textwidth}}{\begin{figure}[!htbp]}
    \centering
    \includegraphics[width=\linewidth]{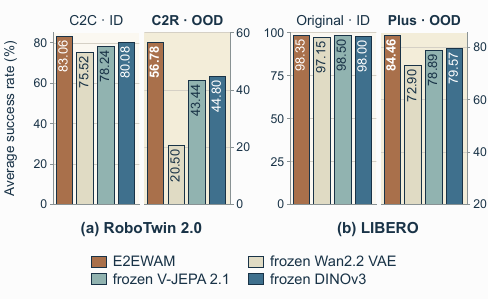}
    \caption{\textbf{Controlled comparison with frozen-latent WAMs.}
    Average success rates (\%) on (a) RoboTwin~2.0 and
    (b) LIBERO and LIBERO-Plus. Separate scales show ID on the left axis and OOD (shaded) on the right.}
    \label{fig:controlled-study}
  \reportlayout{\end{wrapfigure}}{\end{figure}}%
}
\reportlayout{\controlledstudyfigure}{%
  \begingroup
  \clubpenalties 3 10000 10000 0
  \noindent}%
\textbf{End-to-end latent learning yields stronger OOD generalization
than frozen-latent WAMs.}
In our controlled study (Fig.~\ref{fig:controlled-study}),
E2EWAM consistently surpasses frozen-latent WAM baselines across benchmarks,
with particularly pronounced gains under out-of-distribution settings.
Similar to prior findings~\citep{lyu2026lda1b, lawam2026},
frozen vision foundation features (V-JEPA~2.1 and DINOv3)
generalize better to OOD perturbations than the pixel-level Wan2.2~VAE,
likely because a pixel VAE forces the latent space
to retain task-irrelevant appearance details
that heighten sensitivity to visual distractors.
E2EWAM further shapes the latent space toward manipulation-relevant information
through end-to-end training, which may explain its additional gains
over frozen foundation features.
\reportlayout{}{%
  \par
  \endgroup
  \controlledstudyfigure}

Linear probing results in Table~\ref{tab:encoder-probing}
(Appendix~\ref{app:probing-encoder}) corroborate this insight:
across nearly all physical states (both static and dynamic)
on all three benchmarks, E2EWAM achieves the highest readout $R^2$,
suggesting that its learned representations better preserve
information relevant to robotic manipulation.

\begin{figure*}[!t]
  \begin{minipage}[t]{0.485\textwidth}
    \setlength{\columnwidth}{\linewidth}
    \centering
    \includegraphics[width=\columnwidth]{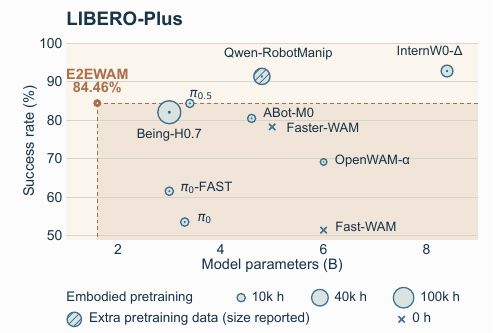}
    \caption{\textbf{Results on LIBERO-Plus.} Each marker's center gives the model's success rate (vertical axis) and parameter count in billions (horizontal axis). Circle area indicates embodied pretraining hours. Shading marks models with more parameters and lower success rates than E2EWAM.}
    \label{fig:pareto_libero_plus}
  \end{minipage}
  \hfill
  \begin{minipage}[t]{0.485\textwidth}
    \setlength{\columnwidth}{\linewidth}
    \centering
    \includegraphics[width=\columnwidth]{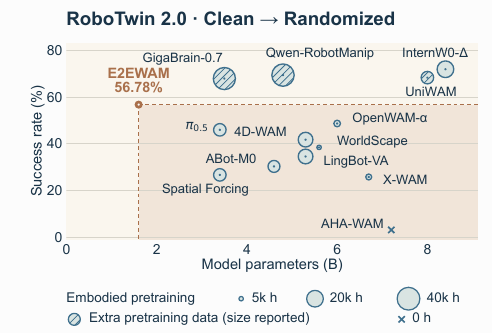}
    \caption{\textbf{Results on RoboTwin 2.0 (clean-to-randomized).} Each marker's center gives the model's success rate (vertical axis) and parameter count in billions (horizontal axis). Circle area indicates embodied pretraining hours. Shading marks models with more parameters and lower success rates than E2EWAM.}
    \label{fig:pareto_robotwin_cr}
  \end{minipage}
\end{figure*}

\noindent\textbf{E2EWAM outperforms larger models at comparable pretraining scales.}
With only 1.6B parameters and approximately 6,400 hours of pretraining,
E2EWAM achieves strong performance against substantially larger models,
particularly in OOD generalization
(Figs.~\ref{fig:pareto_libero_plus} and~\ref{fig:pareto_robotwin_cr}).
On LIBERO-Plus, E2EWAM achieves 84.46\% success,
ranking second among WAMs, behind only InternW0-$\Delta$
(8.4B parameters, over three times the pretraining data).
On RoboTwin 2.0 (Clean-to-Randomized),
E2EWAM leads all methods pretrained on fewer than 10K hours (56.78\%).
Notably, it surpasses the previously leading OpenWAM-$\alpha$ (48.70\%),
a 6B-parameter WAM trained on a comparable amount of data
while leveraging large-scale pretrained video generation weights.

Moreover, E2EWAM also surpasses several larger policies
trained on larger robot pretraining corpora,
including $\pi_{0.5}$, 4D-WAM, LingBot-VA, and Spatial Forcing.
Methods achieving higher success generally rely on substantially larger models
and more extensive pretraining, sometimes supplemented
by vision-language co-training or tactile pretraining
(Appendix~\ref{app:simulation_results}).
As shown in Figs.~\ref{fig:pareto_libero_plus} and~\ref{fig:pareto_robotwin_cr},
E2EWAM lies on the Pareto frontier of task success
against both model size and pretraining data volume,
suggesting substantial room for further improvements
through scaling model capacity and pretraining data.

\subsection{Real-World Evaluation}

\noindent\textbf{Real-robot experimental setup.}
To further validate E2EWAM in real-world manipulation,
we conduct experiments on the I2RT YAM robot,
comparing against the leading VLA
$\pi_{0.5}$~\citep{physicalintelligence2025pi05}
and WAM OpenWAM-$\alpha$~\citep{openwam2026}.
We follow each baseline's official fine-tuning guidelines.
We design five real-world tasks targeting distinct manipulation capabilities.
We report success rate (SR) in Fig.~\ref{fig:real-robot-results}
and progress score (PS) in Appendix~\ref{app:real_progress},
with experimental details in Appendix~\ref{app:real_protocol}.

\begin{figure*}[!t]
  \centering
\textbf{}  \iflatexml\includegraphics[width=\textwidth]{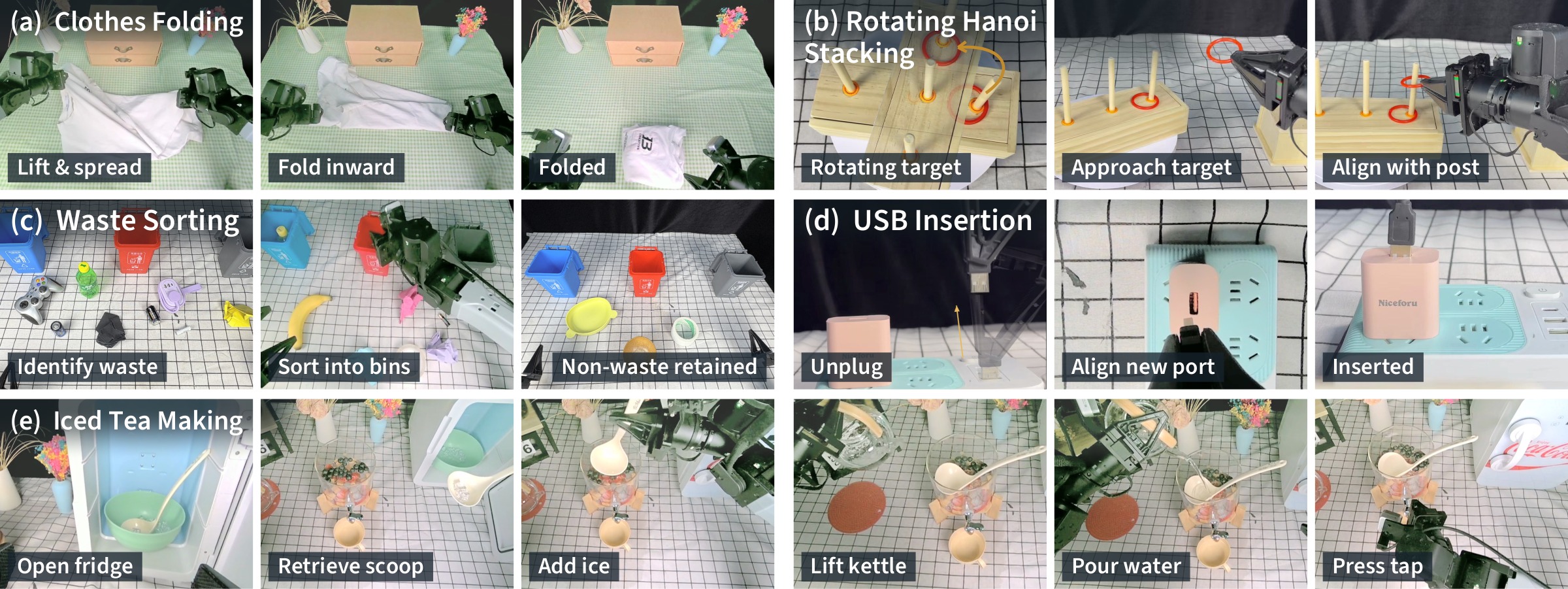}\else\includegraphics[width=\textwidth]{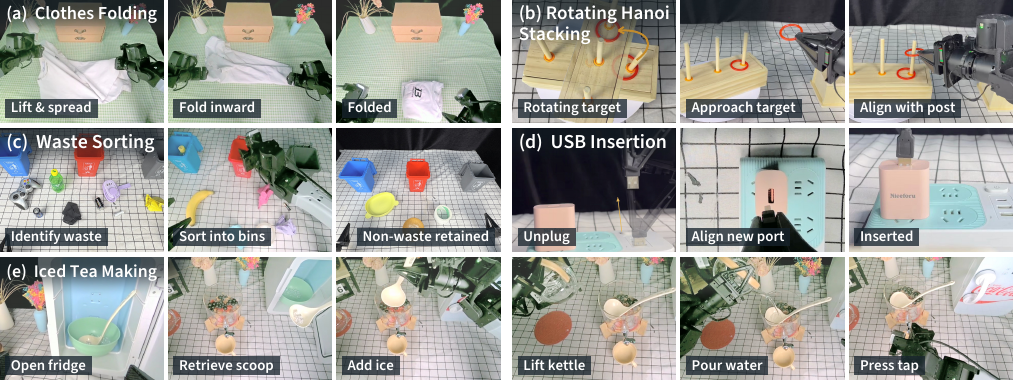}
\fi
  \caption{\textbf{Real-robot task overview.} Representative stages of
  Clothes Folding, Rotating Hanoi Stacking,
  Waste Sorting, USB Insertion, and Iced Tea Making.}
  \label{fig:real_tasks}
\end{figure*}

\begin{itemize}
  \item \textbf{Semantic Understanding: Waste Sorting.}
  In this task, the policy must identify waste items
  on a cluttered tabletop with distractors
  and sort them into the correct bins (Fig.~\ref{fig:real_tasks}).
  The task covers three categories:
  recyclable bottles, batteries, and paper scraps.
  For Object OOD, we test whether the policy can map unseen object appearances
  to their semantic categories amidst clutter.

  \item \textbf{Deformable Manipulation: Clothes Folding.}
  Clothes Folding requires tracking the changing physical state
  of deformable fabric.
  We evaluate three OOD settings:
  object appearance, background, and their combination.

  \item \textbf{Dynamic Understanding: Rotating Hanoi Stacking.}
  In this task, the policy must place a ring onto the post that already holds a ring
  on a Tower-of-Hanoi platform rotating at a constant speed.
  The platform operates at three discrete speed levels:
  demonstrations are collected under the low- and medium-speed settings (ID),
  whereas Velocity OOD evaluates the policy at the unseen high-speed setting.
  We train each policy with two history frames as input,
  and each policy must infer the platform's rotational speed
  and anticipate the target post's position at the moment of release.

  \item \textbf{Contact-Rich Manipulation: USB Insertion.}
  This task requires the policy to unplug a USB connector from one port
  and insert it into another port at a different height
  (Fig.~\ref{fig:real_tasks}).
  For Object OOD, we evaluate insertion into unseen target devices
  with different shapes and appearances.

  \item \textbf{Long-Horizon Manipulation: Iced Tea Making.}
  This multi-stage task requires coordinated bimanual manipulation
  across multiple objects, involving six sequential stages of manipulation.
  To assess robustness against environmental distractions,
  we set Background OOD by varying tabletop textures and distractor layouts.
\end{itemize}

\begin{figure*}[!t]
  \centering
  \includegraphics[width=\textwidth]{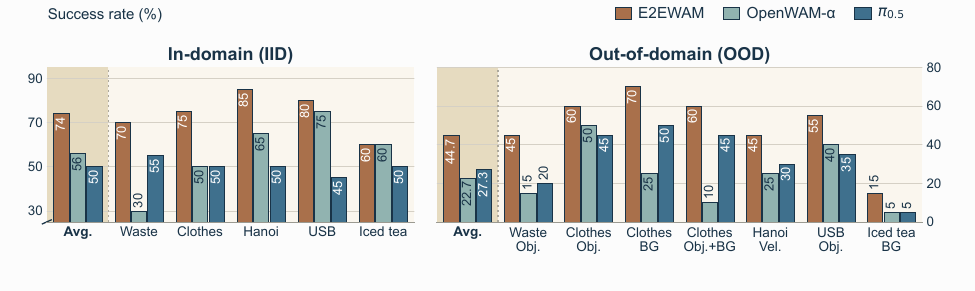}
  \caption{\textbf{Real-robot evaluation.}
  Success rate (SR) across five tasks and
  12 task--condition pairs. Shaded groups report task-balanced averages:
  Avg IID averages the five IID settings; Avg OOD first averages OOD
  conditions within each task, then averages over tasks.
  IID denotes in-distribution evaluation. Obj., BG, and Vel. denote
  object, background, and velocity perturbations, respectively;
  Obj.+BG combines object and background perturbations.}
  \label{fig:real-robot-results}
\end{figure*}

Overall, E2EWAM outperforms both OpenWAM-$\alpha$ and $\pi_{0.5}$ in average success rates under both IID and OOD settings (Fig.~\ref{fig:real-robot-results}). From these experiments, we highlight three observations: First, E2EWAM shows stronger robustness to visual shifts, particularly under substantial changes in backgrounds and distractor layouts. Second, E2EWAM captures task-relevant object semantics: in Waste Sorting, both E2EWAM and $\pi_{0.5}$ generally map unseen object appearances to their semantic categories, 
whereas OpenWAM-$\alpha$ more frequently makes errors in target selection or bin placement. Third, while the dynamic understanding task remains challenging for all models, E2EWAM more accurately captures the unseen velocity and releases the ring at the appropriate timing. Detailed analyses are provided in Appendix~\ref{app:real_task_analysis}.

\subsection{Data Scaling Effect for E2EWAM}
\label{sec:data_scaling}

Scaling effects have long driven the exceptional generalization capabilities
of large language models (LLMs), motivating growing interest in data scaling laws
for robotic manipulation~\citep{lin2025data, zheng2026egoscale,
allshire2026abc, lyu2026lda1b}.
In particular, an effective foundation policy is expected to benefit
from diverse cross-embodiment data, translating larger pretraining scale
into stronger out-of-distribution generalization.
In this section, we conduct an empirical study
to investigate the data scaling effect of E2EWAM.

\newcommand{\scalingfigure}{%
  \reportlayout{\begin{wrapfigure}{R}{0.55\textwidth}}{\begin{figure}[!htbp]}
    \centering
    \includegraphics[width=\linewidth]{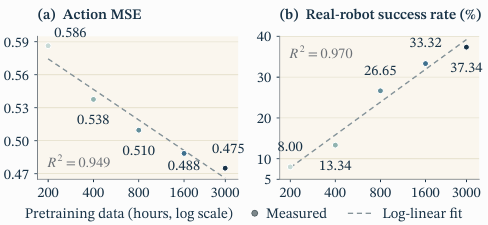}
    \caption{\textbf{Pretraining-data scaling.}
    (a) Action MSE (z-score normalization) and (b) real-robot success rate.
    Points denote measured values; dashed lines show log-linear fits.}
    \label{fig:data_scaling}
  \reportlayout{\end{wrapfigure}}{\end{figure}}%
}
\reportlayout{\scalingfigure}{}

We construct five nested pretraining subsets
of 200, 400, 800, 1,600, and 3,000 hours
from the same six cross-embodiment robot datasets
used in Section~\ref{sec:experiment-settings}.
We perform task-stratified sampling to preserve the task distribution
of the full corpus, ensuring that each smaller subset
is strictly contained in the larger ones.
To isolate the effect of data scale, we pretrain E2EWAM
with an identical architecture and training budget across all scales,
fixing the global batch size to 1,024
and the number of training steps to 32K,
approximately one epoch over the 3,000-hour subset.
This controlled setup allows us to investigate how E2EWAM benefits
from increasing pretraining data under a fixed compute budget.

\noindent\textbf{E2EWAM exhibits log-linear scaling with pretraining data.}
As shown in Fig.~\ref{fig:data_scaling}(a),
held-out validation action MSE ($z$-score normalized)
decreases log-linearly as the pretraining corpus grows ($R^2 = 0.949$).
To further verify whether this trend translates into downstream generalization,
we fine-tune a generalist policy from each pretrained checkpoint
on our five-task real-robot benchmark and evaluate it under OOD settings.
Consistent with the validation loss,
average OOD success rate scales log-linearly with pretraining data volume
($R^2 = 0.970$; Fig.~\ref{fig:data_scaling}(b)), rising from 8.00\% to 37.34\%.
Together, these results demonstrate favorable data scaling behavior of E2EWAM,
suggesting further potential for improvement
with larger pretraining datasets and additional compute.

\reportlayout{}{\scalingfigure}

\FloatBarrier
\section{Related Works}
\label{sec:related_works}
\textbf{World-Action Models.}
WAMs learn world dynamics through future latent prediction
to benefit robotic manipulation.
Most prevailing WAMs inherit large-scale pretrained video generators
and operate within their frozen pixel-VAE latent
spaces~\citep{dreamzero2026, nvidia2026gr00t2, lingbotva2026, fastwam2026,
ma2026fasterwam, miao2026internw0delta, openwam2026, cosmospolicy2026,
guo2026videopredictionpolicy2, motus2025}.
An alternative line of work adopts frozen visual foundation features
as the latent space~\citep{lyu2026lda1b, lawam2026, luo2026beingh07,
jepawam2026, sun2026vlajepa},
encoding general visual semantics that are less susceptible to visual distractors.
Recent approaches further enrich these representations
through semantic alignment \citep{lingbotva2_2026}
or explicit geometric and depth supervision \citep{prismwam}.
Nevertheless, these approaches still decouple latent shaping from WAM training.
More closely related to our work, ReWAM \citep{rewam2026}
shapes a learnable representation bottleneck exclusively through action supervision,
whereas LeWAM \citep{lewam2026} and SLIM-0.5B~\citep{slim2026}
jointly learn to predict future latents and actions.
LiLa-WAM and Copper-Policy instead learn trainable adapters
on top of frozen visual foundation encoders \citep{lilawam,copperpolicy}.
However, these methods adopt diverging, coupled designs
without a clear consensus or systematic comparison,
leaving controlled comparisons against frozen-latent WAMs
and generalist performance under large-scale pretraining underexplored.
Instead, we systematically disentangle the design space through controlled studies,
and validate our recipe at scale against both frozen-latent counterparts
and leading policies.

\textbf{Visual Latent Representation Learning.}
Learning generalizable visual representations
has long been a central focus in computer vision.
A rich body of work learns transferable representations from images and videos
through masked modeling \citep{mae2022,tong2022videomae},
self-distillation \citep{byol2020,dino2021,oquab2023dinov2,simeoni2025dinov3},
joint-embedding objectives \citep{Assran_2023_CVPR,vjepa2024,vjepa2_2025,
murlabadia2026vjepa21,balestriero2025lejepa,kuhn2026levjepa}.
Another line of work focuses on robotics-oriented visual representation learning,
applying SSL or action-related contrastive learning
to egocentric human videos and robot
data~\citep{r3m2022,mvp2022,vc1_2023,sharma2026caip}.
Building on this literature,
we study representation learning within WAMs under explicit task supervision,
systematically revisiting established representation paradigms
from a manipulation-oriented perspective.

\textbf{Latent World Models.}
Latent world models learn predictive dynamics in compact latent spaces.
Early approaches~\citep{ha2018world,hafner2019learning,hafner2020dream,
hafner2021mastering, hafner2025mastering} rely on pixel-reconstruction,
whereas the TD-MPC family~\citep{hansen2022tdmpc, tdmpc2_2024}
learns latent dynamics.
More recently, reconstruction-free world models
have gained increasing attention.
For instance, DINO-WM~\citep{dinowm2025} and V-JEPA~2-AC~\citep{vjepa2_2025}
model transition dynamics within frozen representation spaces.
Going further, JEPA-style end-to-end latent world
models~\citep{pldm2025,lewm2026,zhang2026hjepa}
jointly learn transition dynamics
while shaping latent representations,
with INTACT~\citep{sun2026intact}
and Sensorimotor World Models~\citep{ivashkov2026sensorimotor}
incorporating inverse-dynamics-based supervision to shape representations.
While end-to-end latent WAMs can be viewed as a generalized form under JEPA architecture,
most JEPA world models primarily focus on world-state planning
via model predictive control (MPC).
Our work instead revisits this design space
from the perspective of manipulation policy learning,
systematically diagnosing how longstanding issues
such as the slow-feature shortcut~\citep{sobal2022slow}
undermine manipulation-relevant dynamics
and developing targeted remedies to mitigate them.

\section*{Conclusion and Limitations}
In this work, we systematically investigate end-to-end latent representation learning in world-action models from a manipulation-oriented perspective. Through controlled studies, we establish a progressive design roadmap covering action integration, visual supervision objectives, and representation regularization. Notably, we reveal that conventional anti-collapse regularization fails to prevent a slow-feature shortcut that undermines downstream manipulation, and develop targeted remedies to preserve manipulation-relevant temporal dynamics. We distill these findings into \textbf{E2EWAM}, which demonstrates strong performance against frozen-latent WAMs and leading policies across simulation and real-world benchmarks. Together with its log-linear data scaling behavior, our findings highlight end-to-end latent learning as an effective and scalable paradigm for generalist robotic manipulation.

Despite these results, several limitations remain. Due to computational constraints, E2EWAM is limited to 1.6B parameters and approximately 6,400 pretraining hours, and thus trails some larger models with more extensive embodied pretraining on simulation benchmarks. Nevertheless, its favorable data scaling behavior suggests potential for further improvements with additional compute. Moreover, our training relies solely on RGB observations without depth or explicit 3D cues. Future work could explore richer data sources and geometry-aware latent representations. Finally, as this work focuses on visual latent representation learning and positions E2EWAM as a System 1 policy, it lacks dedicated designs for language planning and memory. Extending these capabilities and integrating E2EWAM with higher-level System 2 agents represents a promising direction for future research.

\section*{Acknowledgement}

This research was conducted with the support of the Shanghai Qi Zhi Institute \& Spirit AI Innovation Program. Funding and support for this work were also provided by the Tsinghua University - Keystone Electrical (Zhejiang) Co., Ltd. Joint Research Center for Embodied Multimodal Artificial Intelligence (JCEMAI). Additionally, we would like to extend our thanks to the Xiongan AI Institute. This project is also supported by the Beijing Natural Science Foundation (JR25019).

We sincerely thank Prof. Huazhe Xu and his team for their support with the I2RT YAM robotic arm, especially Shenghan Lin for generously helping us set up the system. We also gratefully acknowledge support from the Frontier Exploration team led by Yingdong Hu, as well as from Richard Wang at Spirit AI. We also extend our thanks to Shengjie Wang for the insightful discussions.

\FloatBarrier
\bibliographystyle{plainnat}
\bibliography{references}

\clearpage
\appendix
\columnfloatsfalse
\addtocontents{toc}{\protect\setcounter{tocdepth}{2}}
\setcounter{tocdepth}{2}
\begingroup
\renewcommand{\contentsname}{Appendix Contents}
\hypersetup{linktoc=all}
\tableofcontents
\endgroup
\clearpage

\section{Details and Supplementary Experiments for Section 3}
\label{app:sweeps}

\subsection{Detailed evaluation protocol}

\noindent\textbf{Pretraining.}
We draw our pretraining data from six robot datasets, namely
DROID~\citep{khazatsky2024droid}, BC-Z~\citep{jang2022bcz},
RT-1~\citep{brohan2023rt1}, MolmoAct2-BimanualYAM~\citep{fang2026molmoact2},
BridgeData V2~\citep{walke2023bridgedata}, and
ABC-130K~\citep{allshire2026abc}.
Task-stratified subsampling within each dataset yields a corpus of
approximately 245~hours of robot demonstrations.
To control for pretraining data, we train all variants on this same
subset for 25,000 optimizer steps with a global batch size of 256.

\noindent\textbf{Fine-tuning and evaluation.}
We fine-tune all variants exclusively on the OpenVLA-filtered
LIBERO dataset~\citep{openvla2024} for 25,000 optimizer steps
with a global batch size of 128.
We follow the official evaluation protocols for
LIBERO~\citep{liu2023libero} and LIBERO-Plus~\citep{liberoplus2025}.
We measure in-domain manipulation performance on the original LIBERO suites.
To assess out-of-distribution (OOD) generalization, we evaluate the same
LIBERO-fine-tuned policies zero-shot on the perturbed LIBERO-Plus tasks.

\noindent\textbf{Hyperparameters.}
We train all trainable components of the 110M-parameter reference WAM
from scratch, including the visual encoder.
It follows the same encoder--predictor--action-expert architecture as
the 1.6B model in Section~\ref{sec:experiments}, with reduced module
widths and depths (Table~\ref{tab:roadmap-reference-model}).
We use the same stage-specific augmentations across variants
(Table~\ref{tab:roadmap-augmentations}).
Multi-crop alignment and masked latent prediction additionally use the
objective-specific crops and masks studied in Appendix~\ref{app:visual_objectives}.
We detail the remaining key hyperparameters in the following subsections.

\begin{table}[!htbp]
  \centering
  \small
  \caption{Architecture of the online reference WAM in pretraining.
  The action expert uses four encoder-connected blocks and eight
  predictor-connected blocks.}
  \label{tab:roadmap-reference-model}
  \begin{tabular}{@{}lrrrrr@{}}
    \toprule
    Component & Blocks & Width & FFN width & Heads & Params.\ (M) \\
    \midrule
    Visual encoder (ViT-B/16) & 12 & 768 & 3072 & 12 & 86.8 \\
    Predictor                 &  8 & 320 & 1280 &  5 & 16.6 \\
    Action expert               & 12 & 128 &  512 &  5 &  8.3 \\
    \midrule
    Total & & & & & 111.7 \\
    \bottomrule
  \end{tabular}
\end{table}

\begin{table}[!htbp]
  \centering
  \small
  \caption{Shared image augmentations for pretraining and fine-tuning.}
  \label{tab:roadmap-augmentations}
  \begin{tabular}{@{}lcc@{}}
    \toprule
    Setting & Pretraining & Fine-tuning \\
    \midrule
    Color jitter probability & $1.0$ & $0.8$ \\
    Brightness factor        & $[0.7,1.3]$ & $[0.7,1.3]$ \\
    Contrast factor          & $[0.6,1.4]$ & $[0.6,1.4]$ \\
    Saturation factor        & $[0.5,1.5]$ & $[0.5,1.5]$ \\
    Hue shift                & $[-0.08,0.08]$ & $[-0.05,0.05]$ \\
    Random crop (linear scale, probability) & $0.95$, $1.0$ & Disabled \\
    \bottomrule
  \end{tabular}
\end{table}

\FloatBarrier

\subsection{Integrating Action Supervision.}
\label{app:arch_ablation}

\subsubsection{Implementation Details}

\textbf{Details of the action-conditioning topologies.}
Across all three topologies, we keep the encoder and predictor
architectures fixed and use 12 action-expert blocks of the same design.
\emph{Encoder Only.} The 12 action-expert blocks connect one-to-one
to the 12 encoder layers in depth order.
\emph{Predictor Only.} The action expert connects to all eight predictor
layers in depth order, with layers 2, 4, 6, and 8 each feeding two
consecutive action-expert blocks.
This keeps the action expert at 12 blocks, ensuring the same depth and
parameter count across all three topologies.
\emph{Encoder + Predictor.} The first four action-expert blocks connect
to encoder layers at 25\%, 50\%, 75\%, and 100\% of the encoder depth,
and the remaining eight connect one-to-one to the eight predictor layers.
In an additional ablation, using the last four encoder layers yielded
a lower average success rate on LIBERO-Plus than using the four
evenly spaced layers (Table~\ref{tab:encoder-layer-selection}).

\begin{table}[t]
  \centering
  \small
  \caption{\textbf{Encoder layer selection in the joint topology.}
  The action expert connects to all eight predictor layers in both settings.
  Results are average success rates (\%).}
  \label{tab:encoder-layer-selection}
  \begin{tabular}{@{}lcc@{}}
    \toprule
    Encoder layers & LIBERO & LIBERO-Plus \\
    \midrule
    Evenly spaced (3, 6, 9, 12) & 93.25 & \textbf{26.65} \\
    Last four (9--12)           & \textbf{93.85} & 25.93 \\
    \bottomrule
  \end{tabular}
\end{table}

\subsubsection{Supplementary Experiments}

\begin{figure}[!htbp]
  \centering
  \includegraphics[width=\textwidth]{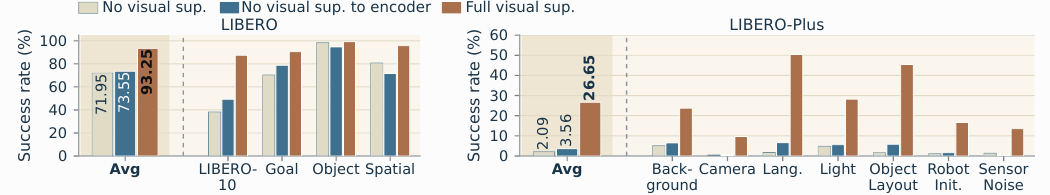}
  \caption{\textbf{Visual supervision for encoder learning.}
  Visual supervision remains essential for encoder learning,
  even with full action supervision.
  Results are reported on LIBERO (left) and LIBERO-Plus (right).
  See Fig.~\ref{fig:supervision_framework} for the model architecture.}
  \label{fig:visual-supervision-ablation}
\end{figure}

\textbf{With action supervision introduced,
visual prediction is still essential for encoder learning.}
In Section~\ref{sec:roadmap-supervision}, we examined how action
supervision shapes different model components.
Symmetrically, here we further investigate how supervision from
visual future prediction should be allocated across modules.
Following the same gradient-routing approach, we progressively extend
visual prediction supervision from none, to the predictor alone,
and finally to both the predictor and encoder.
As shown in Fig.~\ref{fig:visual-supervision-ablation},
when the encoder receives only action supervision
(\emph{No visual sup. to encoder}),
despite both losses training the predictor,
average success drops from 93.25\% to 73.55\% on LIBERO
and from 26.65\% to 3.56\% on LIBERO-Plus.
Removing visual prediction entirely (\emph{No visual sup.})
yields similarly poor results
(71.95\% and 2.09\%, respectively).
These results highlight the critical role of visual prediction
in shaping useful representations,
which action supervision alone cannot adequately provide.

\FloatBarrier

\subsection{Designing the Visual Supervision Objective}
\label{app:visual_objectives}
\label{app:visual_objective_details}

\paragraph{Controlled evaluation.}
To ensure a controlled comparison, we vary only the visual prediction
objective across variants.
As in our other experiments, we keep the model architecture, training
dataset, and other experimental settings identical across variants.
We perform action prediction in a forward pass separate from
visual prediction.
Across all variants, this forward pass predicts the future action
chunk conditioned on the current observed frame and the task instruction.

\paragraph{Masked latent prediction.}
We mask a subset $M$ of spatiotemporal token positions in the future
clip and predict their features from visible-token representations
and mask tokens.
Let $\hat z_i$ denote the predicted feature at position $i$
and $\bar z_i$ its corresponding EMA target.
The objective for one sampled mask is
\begin{equation}
  \mathcal{L}_{\mathrm{mask}}
  = \frac{1}{|M|} \sum_{i \in M}
    \|\hat z_i - \bar z_i\|_1.
  \label{eq:app-masked-latent}
\end{equation}

Following the V-JEPA family~\citep{vjepa2024,vjepa2_2025},
we generate masks from spatial blocks at two scales.
We use last-layer features and sample the two mask groups separately,
forming each mask as the union of its blocks.
In the selected configuration, the first group uses eight blocks
at an area scale of $0.15$ and the second uses two at $0.70$,
with both scales defined relative to the valid spatial area.
Each block spans the full temporal extent of the future clip.
Within each group, we average absolute errors over masked tokens
and feature dimensions.
Padding and unavailable visual regions are excluded from the loss.

We sweep the overall masking ratio by varying the number of blocks
in each group. An overall masking ratio of approximately $70\%$
yields the highest average success rates on both LIBERO and
LIBERO-Plus (Table~\ref{tab:masking-ratio}).

\begin{table}[htbp]
  \centering
  \small
  \caption{\textbf{Masking ratio ablation.}
  The masking ratio is the estimated mean fraction of valid tokens
  masked across the two groups. Results are average success rates (\%).}
  \label{tab:masking-ratio}
  \begin{tabular}{@{}lcccc@{}}
    \toprule
    Masking ratio & Blocks at scale $0.15$ & Blocks at scale $0.70$
                  & LIBERO & LIBERO-Plus \\
    \midrule
    $50\%$ &  3 & 1 & 93.00 & 22.80 \\
    $70\%$ &  8 & 2 & \textbf{93.60} & \textbf{24.81} \\
    $88\%$ & 18 & 8 & 90.15 & 19.05 \\
    \bottomrule
  \end{tabular}
\end{table}

\paragraph{Multi-crop alignment.}
The student processes global and local views indexed by $\mathcal{V}$,
while the EMA teacher processes only global views indexed by
$\mathcal{V}_g$, all sampled from the same future clip.
Following DINO~\citep{dino2021}, we minimize the cross-entropy
between their output distributions:
\begin{equation}
  \mathcal{L}_{\mathrm{align}}
  = -\sum_{g \in \mathcal{V}_g}
      \sum_{\substack{v \in \mathcal{V} \\ v \neq g}}
      (\bar p^{\,g})^\top \log p^{\,v},
  \label{eq:app-multicrop-alignment}
\end{equation}
where
\[
  p^{\,v} = \operatorname{softmax}\!\left(
    \frac{q^{\,v}}{\tau_s}\right),
  \qquad
  \bar p^{\,g} = \operatorname{softmax}\!\left(
    \frac{\bar q^{\,g}-\mathbf{c}}{\tau_t}\right).
\]
Here, $q^{\,v}$ and $\bar q^{\,g}$ denote the student and
EMA-teacher logits computed from the \texttt{[CLS]} tokens.
We use $\tau_s=0.1$ for the student and sharpen the teacher
targets with $\tau_t=0.04$.

The teacher encoder and DINO head are updated by EMA,
and teacher probabilities are detached from the gradient computation.
The center $\mathbf{c}$ is initialized to zero and updated
with EMA momentum $0.9$ using the mean teacher logits over
the global views of all valid clips in each batch.
Matching global-view pairs are excluded from the alignment loss.
The implementation averages cross-entropy over the remaining pairs:
with $G=|\mathcal{V}_g|$ and $V=|\mathcal{V}|$,
it equals Eq.~\eqref{eq:app-multicrop-alignment} divided by $G(V-1)$.
The default two global and eight local views give $18$ such pairs.
Cross-entropy is summed over the DINO-head output dimensions
and averaged over valid samples.

Following LeVJEPA~\citep{kuhn2026levjepa}, we apply random
horizontal flipping, color jitter, and grayscale conversion only
to local views, with probabilities $0.5$, $0.8$, and $0.2$,
respectively. For each view, we share the sampled augmentation
parameters across all frames. DINO~\citep{dino2021} applies these
three transformations to both global and local views. In our
experiments, the LeVJEPA augmentation recipe yielded better
downstream manipulation performance than the DINO recipe.

Our multi-crop hyperparameter sweep identifies the local crop scale
as the most influential hyperparameter for downstream manipulation
performance. Table~\ref{tab:local-crop-scale} compares local crop
scales using a global crop scale of $[0.80, 1.00]$. Among the
settings shown, $[0.30, 0.70]$ yields the highest mean success rate
across LIBERO and LIBERO-Plus ($45.60\%$).

\begin{table}[htbp]
  \centering
  \small
  \caption{\textbf{Local crop scale ablation.}
  We report success rates (\%) and their mean across the two benchmarks.}
  \label{tab:local-crop-scale}
  \begin{tabular}{@{}lccc@{}}
    \toprule
    Local crop scale & LIBERO & LIBERO-Plus & Mean \\
    \midrule
    $[0.02, 0.25]$ & 78.70 &  3.81 & 41.26 \\
    $[0.02, 0.40]$ & 78.20 &  4.42 & 41.31 \\
    $[0.10, 0.50]$ & 81.15 &  9.07 & 45.11 \\
    $[0.30, 0.70]$ & 80.20 & 11.00 & \textbf{45.60} \\
    $[0.50, 0.90]$ & 79.50 & 10.69 & 45.10 \\
    \bottomrule
  \end{tabular}
\end{table}

Specifically, we find that local crops need to retain sufficient
spatial context. Overly small crops can miss the robot end-effector
or task-relevant objects, degrading manipulation performance.

\paragraph{Next-frame and future-chunk prediction.}
Let $Z_s$ denote the online-encoded latent of observation $O_s$,
and $\bar Z_s$ its stop-gradient EMA target.
The action segment $A_s$ contains the control actions between
observations $O_s$ and $O_{s+1}$.
We use $K=4$ supervised future frames for both variants.

Next-frame prediction uses teacher forcing:
the prediction at time $t+k$ conditions on the observed latent
prefix $Z_{t:t+k-1}$ and action segments $A_{t:t+k-1}$.
Its objective is
\begin{equation}
  \scalebox{0.85}{$\displaystyle
    \mathcal{L}_{\mathrm{next}}
    = \frac{1}{K} \sum_{k=1}^{K}
    \big\| f_\phi\big(Z_{t+k} \mid Z_{t:t+k-1}, A_{t:t+k-1}\big)
      - \bar Z_{t+k} \big\|^2.
  $}
  \label{eq:app-next-frame}
\end{equation}

Future-chunk prediction conditions every prediction on the
same current latent $Z_t$, together with the action segments
up to the corresponding target time:
\begin{equation}
  \scalebox{0.85}{$\displaystyle
    \mathcal{L}_{\mathrm{chunk}}
    = \frac{1}{K} \sum_{k=1}^{K}
    \big\| f_\phi\big(Z_{t+k} \mid Z_t, A_{t:t+k-1}\big)
      - \bar Z_{t+k} \big\|^2.
  $}
  \label{eq:app-future-chunk}
\end{equation}
Here, $f_\phi(Z_{t+k}\mid\cdot)$ denotes the predictor output
for the specified target frame.

\paragraph{Prediction objectives for future chunks.}
Building on the future-chunk formulation in
Section~\ref{sec:roadmap-prediction}, we compare direct MSE
regression with two flow-matching parameterizations:
velocity ($v$) prediction and clean-latent ($x_0$) prediction.
Prior studies~\citep{zheng2025diffusiontransformersrepresentationautoencoders,singh2026raev2}
have investigated prediction objectives for high-dimensional
latent generation, primarily focusing on generative quality.
Here, we examine their effectiveness for manipulation-oriented
representation learning.
We retain $\mathcal{L}_{\mathrm{chunk}}$ in
Eq.~\eqref{eq:app-future-chunk}
as the direct-regression baseline and use the same chunk
horizon $K=4$, current visual conditioning, and action
conditioning for both flow-matching variants.
Following RAE and RAEv2~\citep{zheng2025diffusiontransformersrepresentationautoencoders,singh2026raev2},
we use a shifted logit-normal distribution for training noise levels.
For each future chunk, we sample a shared noise level
$\tau=\operatorname{sigmoid}(\xi+\log s)$, where
$\xi\sim\mathcal{N}(0,1)$ and $s$ controls the noise shift.
The noisy target latents are
\begin{equation}
  \widetilde Z_{t+k}^{\,\tau}
  = (1-\tau)\bar Z_{t+k}+\tau\epsilon^{(k)},
  \qquad
  \epsilon^{(k)}
  \overset{\mathrm{i.i.d.}}{\sim}\mathcal{N}(0,I).
  \label{eq:app-future-noising}
\end{equation}
Here, $\bar Z_{t+k}$ is the EMA target latent, and
$\epsilon^{(k)}$ denotes independent noise for the $k$-th
future frame.
Larger $s$ shifts the sampling distribution toward higher
noise levels.
We evaluate $s\in\{5,12,24,32\}$ and a pure-noise variant
that fixes $\tau=1$.
For the prediction at frame $t+k$, both parameterizations
take the noisy prefix
$\widetilde Z_{t+1:t+k}^{\,\tau}$ and $\tau$ as inputs,
conditioned on the current visual latent $Z_t$ and
action segments $A_{t:t+k-1}$.

\paragraph{(i)~Velocity prediction.}
The predictor $f_\phi^{(v)}$ outputs
$\hat V_{t+k}^{\,\tau}$ and is trained against
$V_{t+k}^{\,\tau}
=(\widetilde Z_{t+k}^{\,\tau}-\bar Z_{t+k})/\tau_\delta$,
where $\tau_\delta=\max(\tau,\delta)$ and $\delta=0.05$
provides a numerical floor shared by both parameterizations:
\begin{equation}
  \mathcal{L}_{v}
  = \mathbb{E}_{\tau,\epsilon}
    \left[
      \frac{1}{K}\sum_{k=1}^{K}
      \left\|
        \hat V_{t+k}^{\,\tau}-V_{t+k}^{\,\tau}
      \right\|^2
    \right].
  \label{eq:app-future-velocity}
\end{equation}

\paragraph{(ii)~Clean-latent prediction.}
The predictor $f_\phi^{(x_0)}$ outputs
$\hat Z_{t+k}^{\,\tau}$ to reconstruct the clean target
$\bar Z_{t+k}$ from the same noisy inputs.
Following RAEv2~\citep{singh2026raev2},
we convert this output to velocity and compute the loss
against the same target $V_{t+k}^{\,\tau}$:
\begin{equation}
  \mathcal{L}_{x_0}
  = \mathbb{E}_{\tau,\epsilon}
    \left[
      \frac{1}{K}\sum_{k=1}^{K}
      \left\|
        \frac{
          \widetilde Z_{t+k}^{\,\tau}
          -\hat Z_{t+k}^{\,\tau}
        }{\tau_\delta}
        -V_{t+k}^{\,\tau}
      \right\|^2
    \right].
  \label{eq:app-future-clean}
\end{equation}

\paragraph{Physical probing.}
\phantomsection\label{app:probing-visual}
Table~\ref{tab:prediction-target-probing} reports physical-state
probing results from encoder representations after fine-tuning.
Future-chunk prediction achieves the highest linear probing $R^2$
for all four physical properties on both LIBERO and LIBERO-Plus,
suggesting that it better preserves manipulation-relevant physical
information than the alternative objectives.

\begin{table*}[!htbp]
\centering
\caption{\textbf{Physical probing of visual prediction objectives.}
Linear $R^2\!\uparrow$ from encoder latents.}
\label{tab:prediction-target-probing}
\begingroup
\fontsize{8}{10}\selectfont
\setlength{\tabcolsep}{4pt}
\renewcommand{\arraystretch}{1.05}
\setlength{\aboverulesep}{1.4pt}
\setlength{\belowrulesep}{1.4pt}
\newcommand{\targetProbe}[2]{#1\,\ensuremath{\pm}\,#2}
\newcommand{\targetProbeBest}[2]{\targetProbe{\textbf{#1}}{\textbf{#2}}}
\newsavebox{\targetProbingTable}
\iflatexml\else
\begin{lrbox}{\targetProbingTable}
\fi
\begin{tabular}{@{}l*{4}{c}@{}}
\toprule
Objective & \shortstack{EEF\\position} & \shortstack{Arm joint\\position}
          & \shortstack{Gripper joint\\position} & \shortstack{Manipulated object\\position} \\
\midrule
\multicolumn{5}{@{}l}{\textit{(a) LIBERO}} \\
\addlinespace[1pt]
Action only & \targetProbe{0.8838}{0.0015} & \targetProbe{0.4925}{0.0083} & \targetProbe{0.4380}{0.0083} & \targetProbe{0.9528}{0.0009} \\
Next-frame AR & \targetProbe{0.9888}{0.0002} & \targetProbe{0.8850}{0.0019} & \targetProbe{0.8871}{0.0040} & \targetProbe{0.9944}{0.0003} \\
Future chunk & \targetProbeBest{0.9905}{0.0002} & \targetProbeBest{0.8929}{0.0029} & \targetProbeBest{0.9239}{0.0051} & \targetProbeBest{0.9949}{0.0003} \\
Masked latent prediction & \targetProbe{0.9800}{0.0002} & \targetProbe{0.8385}{0.0029} & \targetProbe{0.8779}{0.0015} & \targetProbe{0.9898}{0.0005} \\
Multi-crop alignment & \targetProbe{0.9746}{0.0006} & \targetProbe{0.8193}{0.0040} & \targetProbe{0.7246}{0.0034} & \targetProbe{0.9870}{0.0007} \\
\midrule
\multicolumn{5}{@{}l}{\textit{(b) LIBERO-Plus}} \\
\addlinespace[1pt]
Action only & \targetProbe{0.6803}{0.0060} & \targetProbe{0.2820}{0.0051} & \targetProbe{0.2642}{0.0060} & \targetProbe{0.9320}{0.0007} \\
Next-frame AR & \targetProbe{0.9439}{0.0007} & \targetProbe{0.7428}{0.0063} & \targetProbe{0.8051}{0.0013} & \targetProbe{0.9846}{0.0005} \\
Future chunk & \targetProbeBest{0.9482}{0.0010} & \targetProbeBest{0.7531}{0.0069} & \targetProbeBest{0.8762}{0.0033} & \targetProbeBest{0.9861}{0.0005} \\
Masked latent prediction & \targetProbe{0.9353}{0.0005} & \targetProbe{0.7046}{0.0065} & \targetProbe{0.8395}{0.0034} & \targetProbe{0.9784}{0.0003} \\
Multi-crop alignment & \targetProbe{0.9208}{0.0008} & \targetProbe{0.6630}{0.0047} & \targetProbe{0.5909}{0.0032} & \targetProbe{0.9780}{0.0007} \\
\bottomrule
\end{tabular}
\iflatexml\else
\end{lrbox}
\usebox{\targetProbingTable}
\fi
\par\vspace{2pt}
\iflatexml\else
\begin{minipage}{\wd\targetProbingTable}
\fi
\fontsize{7}{8.5}\selectfont\raggedright
Mean $\pm$ sample SD over five episode splits; probes fitted separately within each domain.
Bold: best per property and domain.
\iflatexml\else
\end{minipage}
\fi
\endgroup
\end{table*}

\FloatBarrier

\providecommand{\reportlayout}[2]{#2}
\reportlayout{\begin{wraptable}{r}{0.50\textwidth}}{\begin{table}[t]}
  \centering
  \caption{\textbf{Future latent prediction.}
  Average success rates (\%) on LIBERO and LIBERO-Plus.}
  \label{tab:prediction-parameterization}
  \begingroup
  \setlength{\tabcolsep}{2.2pt}
  \renewcommand{\arraystretch}{1.0}
  \setlength{\aboverulesep}{1.4pt}
  \setlength{\belowrulesep}{1.4pt}
  \fontsize{8}{9}\selectfont
  \begin{tabular*}{\linewidth}{@{}l@{\extracolsep{\fill}}lcc@{}}
    \toprule
    Formulation & Noise shift & LIBERO $\uparrow$ & LIBERO-Plus $\uparrow$ \\
    \midrule
    Direct regression & \textemdash & \textbf{95.85} & 27.20 \\
    \midrule
    \multirow{5}{*}{FM ($v$)} & 5 & 92.90 & 22.26 \\
     & 12 & 93.80 & 26.91 \\
     & 24 & 94.05 & \textbf{30.43} \\
     & 32 & 94.00 & 29.03 \\
     & Pure noise & 93.65 & 28.80 \\
    \midrule
    \multirow{5}{*}{FM ($x_0$)} & 5 & 90.65 & 19.82 \\
     & 12 & 91.75 & 21.38 \\
     & 24 & 90.75 & 21.05 \\
     & 32 & 92.45 & 23.13 \\
     & Pure noise & 91.25 & 20.87 \\
    \bottomrule
  \end{tabular*}
  \par\vspace{3pt}
  \begin{minipage}{\linewidth}
    \fontsize{7}{8.5}\selectfont\raggedright
    Direct regression predicts future latent chunks.
    FM predicts flow velocity $v$ or clean latents $x_0$.
    Noise shift controls training timestep sampling;
    Pure noise denotes Gaussian future inputs.
    Bold: best per benchmark.
  \end{minipage}
  \endgroup
\reportlayout{\end{wraptable}}{\end{table}}

\paragraph{Relatively high-noise velocity prediction benefits future-chunk prediction.}
As shown in Table~\ref{tab:prediction-parameterization}, increasing the
noise shift from $s=5$ to $s=24$ substantially improves the LIBERO-Plus
success rate for velocity prediction. At $s=24$ (with approximately
$83.7\%$ of samples satisfying $\tau>0.9$), velocity prediction achieves
the best performance on LIBERO-Plus, outperforming direct regression
and all $x_0$-prediction variants while remaining competitive on LIBERO.
The LIBERO-Plus success rate drops at $s=32$ and with pure noise.

\reportlayout{\clearpage}{}

\subsection{Regularization Against Representation Collapse}
\label{app:collapse_details}

\noindent\textbf{EMA.}
We maintain the target encoder as an exponential moving average of the
online encoder (Section~\ref{sec:roadmap-collapse}). After optimizer
step $n$, we update the target encoder parameters as
\begin{equation}
  \bar{\theta}^{(n)}
  = m \bar{\theta}^{(n-1)} + (1-m)\theta^{(n)},
\end{equation}
where $\theta^{(n)}$ denotes the updated online encoder parameters
and $m$ is the EMA momentum.

We retain the backbone's final LayerNorm in both encoders and apply an
additional non-affine LayerNorm to each target patch token across its
feature channels, with $\varepsilon=10^{-6}$. The normalized latents
$\bar Z_{t+k}$ serve as stop-gradient targets for visual prediction.
We use a fixed momentum of $m=0.999$ during pretraining and $m=0.9999$
during fine-tuning.

\noindent\textbf{SIGReg.}
Following LeWM~\citep{lewm2026}, we use SIGReg~\citep{balestriero2025lejepa}
to encourage encoded features to follow $\mathcal{N}(\mathbf{0},\mathbf{I})$.
For a batch of features $Z\in\mathbb{R}^{B\times d}$, we sample $M$
random unit directions $u^{(m)}\in\mathbb{S}^{d-1}$ and average the
Epps--Pulley statistic over the one-dimensional projections,
\begin{equation}
\begin{aligned}
  \operatorname{SIGReg}(Z)
    &= \frac{1}{M}\sum_{m=1}^{M} T(Zu^{(m)}),\\
  T(h)
    &= B \int_{\mathbb{R}}
       \bigl|\widehat{\varphi}_{h}(\omega)-\varphi_0(\omega)\bigr|^2
       \varphi_0(\omega)\,\mathrm{d}\omega .
\end{aligned}
\end{equation}
Here $\widehat{\varphi}_{h}(\omega)=B^{-1}\sum_{b=1}^{B}
e^{\mathrm{i}\omega h_b}$ is the empirical characteristic function,
and $\varphi_0(\omega)=e^{-\omega^2/2}$ is the characteristic function
of $\mathcal{N}(0,1)$.

LeWM applies SIGReg to projected \texttt{[CLS]} tokens.
For a controlled comparison across our WAM variants, we retain the
patch-token representation and apply the same statistic independently
at each spatial position of the current observation.
Let $Z^{(p)}$ collect the projected online encoder features at
patch position $p$ across the global mini-batch. We average the
resulting losses over the $P$ patch positions,
\begin{equation}
  \mathcal{L}_{\mathrm{SIGReg}}
  = \frac{1}{P}\sum_{p=1}^{P}\operatorname{SIGReg}(Z^{(p)}).
\end{equation}
In our experiments, this formulation performed better than applying
SIGReg to spatially mean-pooled features or randomly subsampled patch
tokens. We use $M=4096$ directions shared across workers and 17
uniformly spaced nodes on $[0,3]$ for trapezoidal quadrature.

Following LeWM~\citep{lewm2026},
we apply SIGReg to the output of a two-layer MLP after the visual encoder.
The projector preserves the latent dimensionality and applies batch
normalization and GELU after the first linear layer. Adding the projector
increases the average success rate from 59.95\% to 80.45\% on LIBERO and
from 1.55\% to 4.54\% on LIBERO-Plus. These results are consistent with
LeJEPA's observation that projector design can influence downstream
performance under SIGReg.

Our sweep shows that tuning the SIGReg loss weight $\lambda$
is important for downstream performance. We start from $\lambda=0.09$,
the value in the official LeWM training configuration, and evaluate
nearby weights. Table~\ref{tab:app-regularization-settings} reports the selected
SIGReg weights for action attach and action detach.

\noindent\textbf{VISReg.}
VISReg~\citep{wu2026visreg} regularizes the latent distribution toward an
isotropic Gaussian through scale, shape, and centering terms.
Let $Z\in\mathbb{R}^{N\times D}$ contain the features at one patch
position across $N$ samples. Let $\mu$ be the channel-wise mean and
$\sigma$ the channel-wise population standard deviation. We standardize
the features as
$\widetilde{Z}_{ij}=(Z_{ij}-\mu_j)/\operatorname{sg}(\sigma_j)$,
where $\operatorname{sg}$ denotes stop-gradient. The three terms are
\begin{equation}
  \begin{aligned}
    \mathcal{L}_{\mathrm{scale}}
      &= \frac{1}{D}\sum_{j=1}^{D}(\sigma_j-1)^2,\\
    \mathcal{L}_{\mathrm{shape}}
      &= \frac{1}{NK}\sum_{k=1}^{K}
        \big\|\operatorname{sort}(\widetilde{Z}w_k)-q\big\|_2^2,\\
    \mathcal{L}_{\mathrm{center}}
      &= \frac{1}{D}\|\mu\|_2^2.
  \end{aligned}
\end{equation}
Here $w_k$ is a random unit direction, and $q$ contains the quantiles
$q_i=\Phi^{-1}(i/(N+1))$, where $\Phi$ is the standard normal CDF.

As in SIGReg, we compute the regularizer across the global batch at
each spatial patch position of the online history latents and average
the resulting losses across positions to obtain
$\mathcal{L}_{\mathrm{VISReg}}$.
Following \citet{wu2026visreg} and the SIGReg setup above, we use a
batch-normalized projector implemented as a dimension-preserving
two-layer MLP with batch normalization and GELU after the first
linear layer.
We assign unit weights to the three terms and sample $K=4096$ directions
independently on each GPU.

Our sweep shows that downstream success rates are sensitive to the
VISReg loss weight $\lambda$. The training objective is
\begin{equation}
  \mathcal{L}
  = \mathcal{L}_{\mathrm{act}}
    + (1-\lambda)\mathcal{L}_{\mathrm{pred}}
    + \lambda\mathcal{L}_{\mathrm{VISReg}},
\end{equation}
where $\mathcal{L}_{\mathrm{act}}$ and $\mathcal{L}_{\mathrm{pred}}$
are the action and visual prediction losses, respectively.
Since VISReg and SIGReg have different loss scales, we use
$\lambda=0.6$ as the starting point for the sweep, following the
recommendation for smaller datasets in \citet{wu2026visreg}.
The selected VISReg weights for action attach and action detach
are also reported in Table~\ref{tab:app-regularization-settings}.

\begin{table}[t]
  \centering
  \small
  \caption{\textbf{Selected regularization weights.}
  For each regularizer and action-gradient setting, we select
  the weight that maximizes the arithmetic mean of the average
  success rates on LIBERO and LIBERO-Plus.
  These settings are used in Fig.~\ref{fig:collapse-regularization}.}
  \label{tab:app-regularization-settings}
  \begin{tabular}{@{}lcc@{}}
    \toprule
    Regularizer & Action attach: $\lambda$ & Action detach: $\lambda$ \\
    \midrule
    SIGReg & 0.09 & 0.06 \\
    SIGReg + EMA & 0.03 & 0.12 \\
    VISReg & 0.3 & 0.9 \\
    VISReg + EMA & 0.3 & 0.1 \\
    \bottomrule
  \end{tabular}
\end{table}

\FloatBarrier

\subsection{Regularization Against Slow Feature Shortcut}
\label{app:temporal_details}

\subsubsection{Supplementary Diagnostic Experiments for Slow Feature Shortcut}

\noindent\textbf{Interpretation of $R_{\mathrm{temporal}}$.}
The law of total variance provides a statistical interpretation of
Eq.~\eqref{eq:temporal_variance_ratio}:
\[
  \operatorname{Var}_{N,T}(z)
  = \mathbb{E}_{N}\!\left[\operatorname{Var}_{T}(z)\right]
  + \operatorname{Var}_{N}\!\left(\mathbb{E}_{T}[z]\right).
\]
Here, $N$ indexes patch trajectories and $T$ indexes frames within each
trajectory. The first term measures temporal variation within trajectories,
while the second measures variation across their temporal means.
Thus, $R_{\mathrm{temporal}}$ quantifies the fraction of total latent
variance attributable to within-trajectory temporal variation.
A decrease in this fraction indicates that latent variation becomes
increasingly concentrated in temporally averaged features.
Preserving overall latent variance therefore does not by itself ensure
that temporal variation is preserved.

\begin{figure}[!htbp]
  \centering
  \iflatexml\includegraphics[width=\textwidth]{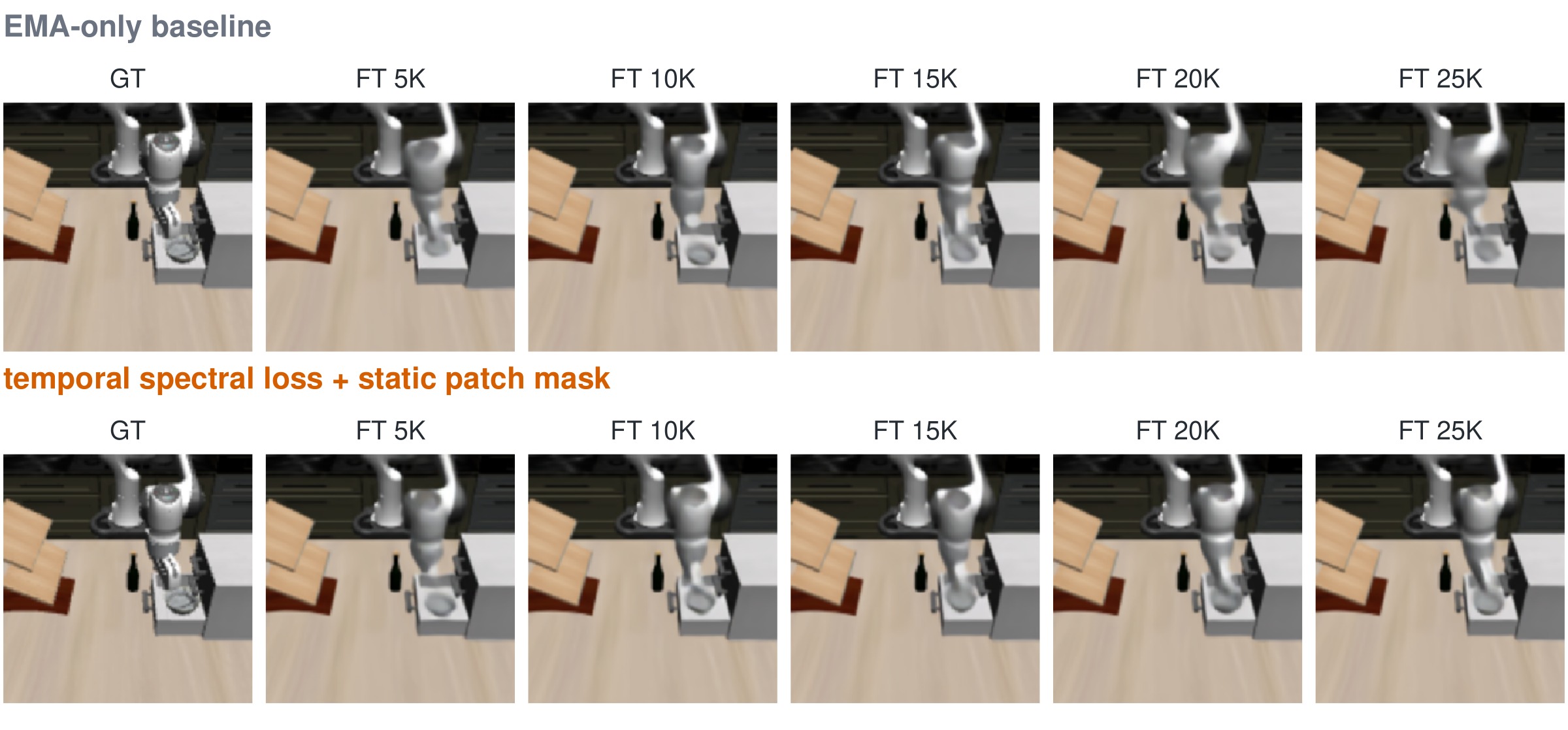}\else\includegraphics[width=\textwidth]{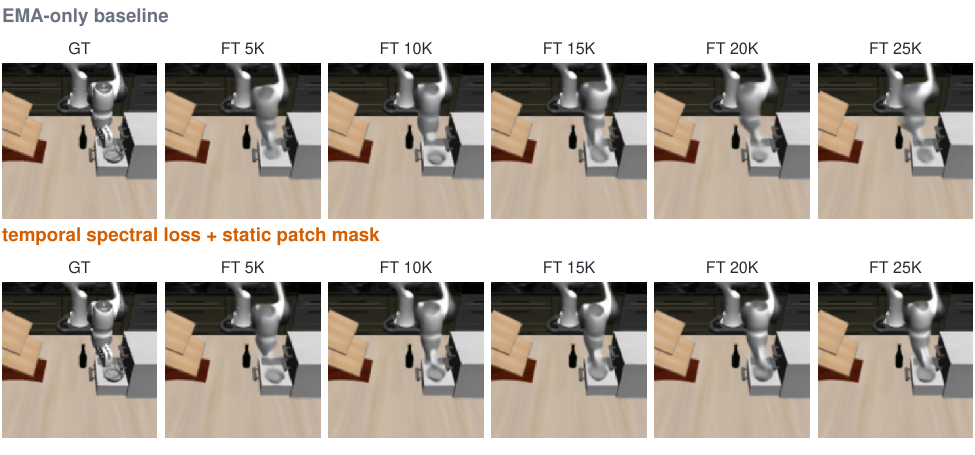}
\fi
  \caption{\textbf{Decoded future latents during fine-tuning.}
  Results on an object layout OOD scene from LIBERO-Plus.
  Each row begins with the ground-truth image (GT), followed by
  decoded predictions at successive fine-tuning checkpoints.}
  \label{fig:decoded-latent-progression}
\end{figure}

\noindent\textbf{Visualizing predicted future latents.}
For each checkpoint during fine-tuning, we train an image decoder
to decode the predicted future latents into images.
Fig.~\ref{fig:decoded-latent-progression} presents the decoding results
for a scene from the object layout OOD dimension of LIBERO-Plus.
As fine-tuning progresses, the EMA-only baseline gradually loses
information about dynamic regions, particularly the robot end effector,
leaving only a blurred appearance.
In contrast, with temporal spectral loss and static patch masking,
these dynamic regions become progressively clearer, suggesting that
the model gradually learns to capture this information.

\FloatBarrier

\subsubsection{Implementation Details}

\noindent\textbf{Temporal spectral loss.}
We match the temporal DCT coefficients of the reconstructed future
latents $\hat{\mathbf{Z}}$ to those of the clean future latents
$\bar{\mathbf{Z}}$ produced by the EMA encoder.
Following Eq.~\eqref{eq:temporal_spectral_loss}, we apply the
orthonormal DCT-II basis $\mathbf{U}$ to the prediction residual
along the $K$ future steps and compute
\[
\begin{aligned}
  \hat{\mathbf{Z}}
    &= \widetilde{\mathbf{Z}}^{\,\tau} - \tau\hat{\mathbf{V}}, \\
  \widetilde{\mathbf{E}}_k
    &= \sum_{j=1}^{K} U_{kj}
       \left(\hat{\mathbf{Z}}_j - \bar{\mathbf{Z}}_j\right), \\
  \mathcal{L}_{\mathrm{spec}}
    &= \mathbb{E}\!\left[
       \sum_{k=0}^{K-1}\omega_k\|\widetilde{\mathbf{E}}_k\|_2^2
       \right].
\end{aligned}
\]
The DC mode ($k=0$) captures the temporal mean, and the AC modes
($k\geq 1$) capture temporal variation.
We sweep the AC-to-DC weight ratio
(Table~\ref{tab:app-temporal-spectral-weights}).

\begin{table}[!htbp]
  \centering
  \small
  \caption{\textbf{Temporal spectral weight sweep.}
  Results are success rates (\%). Avg is the arithmetic mean of the
  two benchmark scores.}
  \label{tab:app-temporal-spectral-weights}
  \begin{tabular}{@{}lccc@{}}
    \toprule
    AC-to-DC weight ratio & LIBERO & LIBERO-Plus & Avg \\
    \midrule
    $1:2$ & 94.65 & 27.80 & 61.23 \\
    $1:1$ & 95.25 & 29.86 & 62.56 \\
    $2:1$ & \textbf{95.55} & \textbf{35.50} & \textbf{65.53} \\
    $3:1$ & 95.15 & 32.97 & 64.06 \\
    $4:1$ & 94.55 & 28.00 & 61.28 \\
    \bottomrule
  \end{tabular}
\end{table}

Relative to equal weighting, moderately upweighting the AC modes
($2:1$) improves success on both benchmarks, with a larger gain on
LIBERO-Plus. Success drops at $4:1$. Upweighting the DC mode ($1:2$)
reduces success on both benchmarks. These results suggest that
encouraging the model to learn temporal dynamics can benefit manipulation.

\noindent\textbf{Static patch mask.}
For each patch $p$, we measure the fraction $r_p$ of energy in the
non-DC modes of the temporal DCT of its clean EMA latents
$\bar{z}_{t,p}$. Let $\mathcal{T}_p$ index the valid steps in the
history and future latent sequence for patch $p$. Following the
main text, we compute $r_p$ in the time domain using the temporal
mean $\mu_p$,
\[
\begin{aligned}
\mu_p &= \frac{1}{|\mathcal{T}_p|}
         \sum_{t\in\mathcal{T}_p}\bar{z}_{t,p},\\
r_p &= \frac{\sum_{t\in\mathcal{T}_p}
              \|\bar{z}_{t,p}-\mu_p\|_2^2}
             {\sum_{t\in\mathcal{T}_p}
              \|\bar{z}_{t,p}\|_2^2}.
\end{aligned}
\]
We identify patches with $r_p \le \eta$ as nearly static, with $\eta = 0.02$, and exclude them from the future prediction loss in order of increasing $r_p$, capping the excluded fraction at 30\% of the valid spatial patches per eligible camera view. This cap avoids over-masking in nearly-static video clips, leaving $\mathcal{L}_{\text{spec}}$ to emphasize temporal dynamics on the remaining tokens. 

\subsection{Putting the Recipe Together}
\label{app:final_recipe}

We distill the findings from our controlled studies into a final
recipe, \textbf{E2EWAM}. Among the variants trained from scratch,
E2EWAM achieves the highest performance, reaching average success
rates of $96.70\%$ on LIBERO and $36.90\%$ on LIBERO-Plus.
Building on this final recipe, we further find that initializing
the visual encoder with pretrained V-JEPA~2.1
weights~\citep{murlabadia2026vjepa21} yields additional gains,
improving success rates to $96.85\%$ on LIBERO and $46.55\%$
on LIBERO-Plus.

Motivated by these gains, we also adopt pretrained V-JEPA~2.1
weights to initialize the visual encoder in our scale-up
experiments in Section~\ref{sec:experiments}.
However, due to computational constraints, we do not evaluate
training the visual encoder from scratch at this larger scale.
Whether the benefits of pretrained initialization diminish as
the amount of training data increases, and whether training
the visual encoder from scratch at scale can yield better
performance, remain important directions for our future exploration.

\section{Implementation Details}
\label{app:implementation}

\subsection{Pretraining Dataset}
\label{app:pretraining_data}
\noindent\textbf{Data sources.}
We use six open-source robot datasets. We obtain MolmoAct2-BimanualYAM,
DROID~\citep{khazatsky2024droid}, RT-1~\citep{brohan2023rt1},
BC-Z~\citep{jang2022bcz}, and Bridge~\citep{walke2023bridgedata}
from the MolmoAct2 release~\citep{fang2026molmoact2}, and also include
ABC-130K~\citep{allshire2026abc}. Our egocentric human data come from
EgoDex~\citep{hoque2025egodex} and EgoVerse~\citep{punamiya2026egoverse}.

\noindent\textbf{Data cleaning.}
After additional quality filtering, the pretraining corpus covers
6,382.50 hours of robot and egocentric human data. 
For robot data, we exclude clips with abrupt jumps in recorded states
or action commands, or with oscillations in the state trajectories.
We also remove clips or episodes with confirmed video defects, including
black screens and unreadable or truncated recordings.
For egocentric human data, we exclude clips with unreliable hand tracking,
gaps in capture timestamps, or abrupt changes in hand or camera motion.
We discard entire episodes with excessive camera translation or without
at least one moving hand with sufficient tracking coverage.

\noindent\textbf{Temporal resampling of egocentric human data.}
Human motions in egocentric recordings tend to be faster than those
in robot demonstrations. To narrow this speed gap, we resample EgoDex
and EgoVerse at half speed. For a clip anchored at source frame $t$,
an offset of $k$ steps on the model's 30~Hz time grid queries source
frame position $t+k/2$. We sample video frames with a frame stride of
$s_{\mathrm f}=4$ action steps on the model timeline. The current frame
and $K=4$ future frames therefore have offsets $js_{\mathrm f}$ for
$j=0,\ldots,K$. At half speed, these correspond to source-frame offsets
$0,2,4,6,8$ from the anchor. The 32 action waypoints are sampled every
half frame from $t+0.5$ to $t+16$.
For action samples between source frames, we linearly interpolate
wrist translations and fingertip positions expressed in the wrist
frame, and use spherical linear interpolation for wrist rotations.

\subsection{Model Architecture}
\label{app:model_architecture}

\noindent\textbf{Scaled-up model architecture.}
The 1.6B E2EWAM used in Section~\ref{sec:experiments} scales up the
encoder, predictor, and action expert of the reference model in
Section~\ref{sec:roadmap} (Table~\ref{tab:e2ewam-scale-up}).
The predictor is a 24-block diffusion Transformer (DiT) that processes
the current observation tokens together with the noisy future latents.
Following the implementation in Wan2.2~\citep{wan2025}, each block
applies self-attention over visual tokens, cross-attention to the task
instruction, and a feed-forward network. Visual
self-attention is causal across frames, and the diffusion timestep
modulates the self-attention and feed-forward sublayers.
A frozen UMT5-XXL encoder supplies the instruction token embeddings,
which the predictor and action expert project separately to their
respective hidden widths.
We initialize the visual encoder with pretrained weights from
V-JEPA~2.1~\citep{murlabadia2026vjepa21} and train the predictor
and action expert from scratch.

The action expert denoises an action chunk with 28 DiT blocks.
We connect these blocks to the visual encoder and predictor through
Mixture-of-Transformers (MoT) attention. In each block, action queries
attend jointly to the action tokens and the visual tokens from the
paired layer. The first four blocks use learned projections of the
hidden states from encoder layers 10, 20, 30, and 40, respectively,
as their visual keys and values. The remaining 24 blocks are paired
one-to-one with the predictor blocks in depth order and reuse their
self-attention keys and values for the current observation.
Each action block then applies cross-attention to the projected
instruction embeddings, followed by a feed-forward network.

\begin{table}[!htbp]
  \centering
  \small
  \caption{Architecture of the online E2EWAM used in the scale-up evaluation.}
  \label{tab:e2ewam-scale-up}
  \begin{tabular}{@{}lrrrrr@{}}
    \toprule
    Component & Blocks & Width & FFN width & Heads & Params.\ (M) \\
    \midrule
    Visual encoder (ViT-G/16) & 40 & 1408 & 6144 & 22 & 1,013.27 \\
    Predictor                & 24 & 1024 & 4096 & 16 &   426.69 \\
    Action expert            & 28 &  512 & 2048 & 16 &   192.89 \\
    \midrule
    Total & & & & & 1,632.85 \\
    \bottomrule
  \end{tabular}
\end{table}

\subsection{Cross-Embodiment Action and Temporal Conditioning}
\label{app:action_representation}

\noindent\textbf{Action space.}
We represent each action step as a 48-dimensional vector with the
following channel layout for robots and human hands, respectively,
where $L$ and $R$ denote the left and right sides and the numbers
below each block give its dimension:
\begin{align*}
\mathbf a_t^{\mathrm{robot}} = \big[
  &\underbrace{\mathrm{EEF}_L}_{9},
   \underbrace{\mathrm{EEF}_R}_{9},
   \underbrace{\mathrm{joint}_L}_{7},
   \underbrace{\mathrm{joint}_R}_{7},
   \underbrace{g_L}_{1}, \underbrace{g_R}_{1},
   \mathbf 0_{14}\big],\\
\mathbf a_t^{\mathrm{human}} = \big[
  &\underbrace{\mathrm{wrist}_L}_{9},
   \underbrace{\mathrm{wrist}_R}_{9},
   \underbrace{\mathrm{fingertips}_L}_{15},
   \underbrace{\mathrm{fingertips}_R}_{15}\big].
\end{align*}
Each robot arm has a scalar gripper command $g$ and seven channels
reserved for joint supervision. Where joint targets are available,
we encode the relative joint pose as the target joint configuration
minus the configuration observed at the start of the chunk.
For each hand, the fingertip block records the change in the 3D
wrist-local coordinates of each of its five fingertips from the start
of the chunk.

For robot datasets with end-effector (EEF) pose waypoints, we follow
the relative trajectory representation of UMI~\citep{chi2024universal}.
Let $(\mathbf p_0,\mathbf R_0)$ denote the observed EEF pose at the
start of the action chunk and $(\mathbf p_t,\mathbf R_t)$ the target
pose at step $t$, both expressed in the same reference frame.
We express every waypoint in the EEF frame at step $0$. Its translation
is $\mathbf R_0^\top(\mathbf p_t-\mathbf p_0)$ and its rotation is
$\mathbf R_0^\top\mathbf R_t$. We concatenate the 3D translation
with the first two rows of the relative rotation matrix, flattened
in row order, to form the 9D EEF block.
RT-1, BC-Z, and Bridge provide per-step EEF deltas. For these
datasets, we retain each step's translation increment and encode
its rotation increment with the same 6D representation.

For each human hand, let $(\mathbf p_t,\mathbf R_t)$ be the wrist
pose at step $t$ in world coordinates, with $\mathbf R_t$ mapping
wrist coordinates to world coordinates. We index the current frame
by $0$ and denote its camera-to-world rotation by $\mathbf C_0$.
Let $\mathbf u_{t,i}$ be the position of fingertip $i$ in the wrist
frame at step $t$. The human action targets are
\begin{align*}
  \Delta\mathbf p_t
    &= \mathbf C_0^\top(\mathbf p_t-\mathbf p_0),\\
  \Delta\mathbf R_t
    &= \mathbf C_0^\top\mathbf R_t\mathbf R_0^\top\mathbf C_0,\\
  \Delta\mathbf u_{t,i}
    &= \mathbf u_{t,i}-\mathbf u_{0,i},
    \qquad i=1,\ldots,5.
\end{align*}
We flatten the first two rows of $\Delta\mathbf R_t$ to obtain
the 6D wrist rotation representation.

We apply per-dimension $z$-score normalization to translations,
fingertip displacements, and relative joint poses. Our pretraining datasets
span multiple embodiments with different action distributions, so we
estimate per-dimension means and standard deviations separately for
each dataset by default. MolmoAct2-BimanualYAM and ABC-130K share the YAM
embodiment, so we pool their valid samples when estimating the mean
and standard deviation of each dimension. We apply the same pooling
to EgoDex and EgoVerse, which share the bimanual human embodiment.
We use the 6D rotations directly and map gripper values $g$ from
$[0,1]$ to $[-1,1]$ by $2g-1$. Unused or unavailable target channels
are zero-padded and excluded from the action loss.

The action expert maps each noisy 48-dimensional action to a
512-dimensional token with an embodiment-specific linear layer.
After the shared Transformer and time-modulated layer normalization,
a second embodiment-specific linear layer projects each token back
to the 48-dimensional action space. Datasets with the same embodiment
share the corresponding action projections.
Table~\ref{tab:e2ewam-action-spaces} lists the supervised action
spaces and sampling rates for the pretraining datasets.

\begin{table}[!htbp]
  \centering
  \small
  \caption{Action supervision and sampling rates in pretraining.
  EgoDex and EgoVerse are retimed to half speed on a 30\,Hz model grid.}
  \label{tab:e2ewam-action-spaces}
  \begin{tabular}{@{}llr@{}}
    \toprule
    Dataset & Supervised action space & Rate (Hz) \\
    \midrule
    MolmoAct2-BimanualYAM & Relative EEF pose, relative joint pose, gripper & 30 \\
    DROID        & Relative EEF pose, relative joint pose, gripper & 15 \\
    RT-1         & EEF delta, gripper                        &  3 \\
    BC-Z         & EEF delta, gripper                        & 10 \\
    Bridge       & EEF delta, gripper                        &  5 \\
    ABC-130K     & Relative EEF pose, relative joint pose, gripper & 30 \\
    EgoDex       & Relative wrist pose, fingertip offsets     & 30 \\
    EgoVerse     & Relative wrist pose, fingertip offsets     & 30 \\
    \bottomrule
  \end{tabular}
\end{table}

\noindent\textbf{Action conditioning.}
For each future latent frame, we condition the predictor on the
normalized action segment between the preceding sampled frame
and the target frame.
For the $j$-th future frame ($j=1,\ldots,K$), this segment contains
action steps $(j-1)s_{\mathrm f}+1$ through $js_{\mathrm f}$ of the
32-step action chunk. We concatenate the $s_{\mathrm f}$ actions, each
of dimension 48, and embed the resulting vector with a two-layer MLP.
Its first linear layer maps $48s_{\mathrm f}=192$ inputs to $1024$
dimensions with embodiment-specific weights.
A shared, zero-initialized linear layer maps the SiLU-activated
embedding to six $1024$-dimensional AdaLN-Zero vectors. We add them
to the diffusion-time modulation of the corresponding frame's tokens
to supply the shifts, scales, and gates for the self-attention and
feed-forward sublayers.

\noindent\textbf{Temporal encoding across sampling rates.}
We retain each robot dataset's native action sampling rate and use
elapsed time to place video and action tokens on a common time axis.
For a stream sampled at $r$ temporal steps per second, consecutive
steps are $1/r$ seconds apart. Their temporal positions in rotary
position embeddings (RoPE) therefore differ by
\[
  \Delta p = \frac{r_{\mathrm{base}}}{r},
\]
where $r_{\mathrm{base}}=1.875$ is the number of RoPE position units
per second.

For a clip with $f$ action steps per second, we use $r=f$ for action
tokens. Since we sample one video frame every $s_{\mathrm f}$ action
steps, we use $r=f/s_{\mathrm f}$ for video tokens. Starting from
position zero at the current frame, we advance the temporal position
by the corresponding increment for each action step or sampled frame.
We use these positions for video tokens in the predictor and for both
video and action tokens in the action expert.

\subsection{Hyperparameters}
\label{app:hyperparameters}
\label{app:frozen_details}

\noindent\textbf{Noise settings.}
For each training example, we independently sample a noise level
for the predictor and another for the action expert.
For the predictor, we draw $\xi\sim\mathcal{N}(0,1)$ and use
$\tau_z=\operatorname{sigmoid}(\xi+\log 24)$ for all four future frames.
We perturb the target latents with independent standard Gaussian noise
as in Eq.~\eqref{eq:app-future-noising} and use the predicted velocity
to recover the clean latents for the temporal spectral loss
in Eq.~\eqref{eq:temporal_spectral_loss}.
For the action expert, we draw $u\sim\operatorname{Beta}(1.5,1)$
and use $\tau_a=0.001+0.999u$ for all 32 steps of the action chunk.
Given the normalized action chunk $A$ and independent standard
Gaussian noise $\epsilon_a$, the expert takes
$(1-\tau_a)A+\tau_a\epsilon_a$ as input and predicts the velocity
$\epsilon_a-A$ on the supervised action channels.

For each frozen latent space in the controlled comparison,
we independently sweep the hyperparameters
and select the best-performing configuration.
In particular, Wan2.2 VAE benefits from lower visual noise
levels than those used for V-JEPA~2.1 and DINOv3.
We find that the best-performing noise setting identified by our
sweep is consistent with the visual noise schedules used in
previous works~\citep{fastwam2026,lingbotva2026,dreamzero2026}:
we sample $u_z\sim\mathcal{U}(0,1)$ and set
$\tau_z=5u_z/(1+4u_z)$.
Table~\ref{tab:latent-noise-settings} summarizes the selected settings.

\begin{table}[!htbp]
  \centering
  \small
  \caption{Noise settings and loss-weight ratios for the
  controlled comparison.
  Shifted logit-normal sampling uses
  $\tau_z=\operatorname{sigmoid}(\xi+\log s)$,
  where $\xi\sim\mathcal{N}(0,1)$.
  All variants share the action-noise distribution specified above.}
  \label{tab:latent-noise-settings}
  \begin{tabular}{@{}llcccc@{}}
    \toprule
    \multirow{2}{*}{Latent space}
    & \multirow{2}{*}{Visual noise sampling}
    & \multirow{2}{*}{Shift $s$}
    & \multicolumn{3}{c}{Action : visual} \\
    \cmidrule(l){4-6}
    & & & Pretrain & LIBERO & RoboTwin \\
    \midrule
    Frozen Wan2.2 VAE & Shifted uniform & 5
    & $1:1$ & $1:1$ & $1:1.25$ \\
    Frozen V-JEPA 2.1 & Shifted logit-normal & 24
    & $1:1.25$ & $1:1.25$ & $1:1.25$ \\
    Frozen DINOv3 & Shifted logit-normal & 24
    & $1:1.25$ & $1:1.25$ & $1:1.25$ \\
    \midrule
    E2EWAM & Shifted logit-normal & 24
    & $1:1.25$ & $1:1.25$ & $1:1.25$ \\
    \bottomrule
  \end{tabular}
\end{table}

\noindent\textbf{Pretraining hyperparameters.}
Table~\ref{tab:e2ewam-pretraining-hyperparameters} summarizes the
pretraining settings.
We predict the next 32 waypoints and sample one video frame every four steps.
In the from-scratch experiments in Section~\ref{sec:roadmap},
we train the visual encoder jointly with the future predictor and
action expert from the first update.
For the scale-up experiments in Section~\ref{sec:experiments},
we initialize the visual encoder with pretrained V-JEPA~2.1 weights,
while the future predictor and action expert start from random weights.
During the 5,000-step learning rate warmup, we keep the encoder
frozen while the predictor and action expert learn. This protects
the pretrained visual features from the early gradients of these
randomly initialized modules.
We then unfreeze the encoder and train all three modules end-to-end
at the same learning rate.

\begin{table}[!htbp]
  \centering
  \small
  \caption{Pretraining hyperparameters for the 1.6B E2EWAM.}
  \label{tab:e2ewam-pretraining-hyperparameters}
  \begin{tabular}{@{}ll@{}}
    \toprule
    Hyperparameter & Setting \\
    \midrule
    Optimizer updates & 125,000 \\
    Global batch size & 1,024 clips \\
    Action chunk size & 32 steps \\
    Frame stride & 4 steps \\
    History length & 1 frame (current observation) \\
    Number of future frames predicted & 4 frames \\
    Optimizer & AdamW, $(\beta_1,\beta_2)=(0.9,0.95)$, $\epsilon=10^{-8}$ \\
    Weight decay & 0.01 \\
    Gradient clipping & Maximum gradient norm of 1.0 \\
    Learning rate schedule & 5,000-step linear warmup to $1.5\times10^{-4}$, then cosine decay to $10^{-6}$ \\
    Visual encoder & Frozen for 5,000 updates, then trained at the base learning rate \\
    Target encoder EMA momentum & 0.9998 throughout training \\
    Loss weights & Temporal spectral loss 1.25, action loss 1.0 \\
    \bottomrule
  \end{tabular}
\end{table}

\noindent\textbf{Image augmentation and camera views.}
We use between one and three camera views per training clip.
We apply the random crop and color jitter in
Table~\ref{tab:e2ewam-pretraining-augmentations} to the selected views,
using the same sampled parameters across frames and cameras.
At each time step, we then arrange the augmented views into a single
$384\times384$ RGB image for the visual encoder.
Table~\ref{tab:e2ewam-pretraining-views} specifies which cameras we
select for each dataset and how we place their views in the image.

\begin{table}[!htbp]
  \centering
  \small
  \caption{Image augmentations used to pretrain the 1.6B E2EWAM.
  We apply random cropping followed by color jitter to every training clip.
  We sample the crop location, color jitter parameters, and order of the
  four color adjustments once per clip and share them across all frames
  and camera views.}
  \label{tab:e2ewam-pretraining-augmentations}
  \begin{tabular}{@{}l>{\raggedright\arraybackslash}p{0.64\linewidth}@{}}
    \toprule
    Augmentation & Pretraining setting \\
    \midrule
    Random crop       & Retain 95\% of each view's width and height at a uniformly sampled location \\
    Brightness factor & Uniform on $[0.7,1.3]$ \\
    Contrast factor   & Uniform on $[0.6,1.4]$ \\
    Saturation factor & Uniform on $[0.5,1.5]$ \\
    Hue shift         & Uniform on $[-0.08,0.08]$ \\
    \bottomrule
  \end{tabular}
\end{table}

\begin{table}[!htbp]
  \centering
  \small
  \caption{Camera views and input layouts used during pretraining.
  We preserve the aspect ratio of each view and center the resulting
  layout on the canvas. For DROID and Bridge, we cycle through the
  available exterior or auxiliary cameras across training epochs and
  keep the selected camera fixed throughout each clip.
  Sizes are given in pixels (width $\times$ height).}
  \label{tab:e2ewam-pretraining-views}
  \newcommand{\viewlayoutcell}[1]{%
    \parbox[c][3.5\baselineskip][c]{\linewidth}{\raggedright #1}}
  \begin{tabular}{@{}>{\raggedright\arraybackslash}m{0.25\linewidth}
                     >{\raggedright\arraybackslash}m{0.25\linewidth}
                     >{\raggedright\arraybackslash}m{0.43\linewidth}@{}}
    \toprule
    Dataset & Views used per clip & Layout on the $384\times384$ input \\
    \midrule
    \viewlayoutcell{MolmoAct2-\newline BimanualYAM} &
      \viewlayoutcell{Top, left wrist, and right wrist} &
      \viewlayoutcell{Top view ($337\times190$) above two wrist views
      ($168\times94$ each).} \\
    \viewlayoutcell{ABC-130K} &
      \viewlayoutcell{Top, left wrist, and right wrist} &
      \viewlayoutcell{Top view above two side-by-side wrist views.
      View sizes depend on the aspect ratio of the unpadded frame.} \\
    \viewlayoutcell{DROID} &
      \viewlayoutcell{Wrist and one of two exterior views} &
      \viewlayoutcell{Wrist above exterior, each $336\times189$.} \\
    \viewlayoutcell{Bridge} &
      \viewlayoutcell{Primary and one auxiliary view when available} &
      \viewlayoutcell{Primary above auxiliary, each $192\times192$.
      With only the primary view, center it at $272\times272$.} \\
    \viewlayoutcell{RT-1, BC-Z} &
      \viewlayoutcell{One RGB camera view} &
      \viewlayoutcell{Fit within $384\times304$ and center.} \\
    \viewlayoutcell{EgoDex, EgoVerse} &
      \viewlayoutcell{One egocentric RGB view} &
      \viewlayoutcell{Fit within $384\times304$ and center.} \\
    \bottomrule
  \end{tabular}
  \par\smallskip
  \begin{minipage}{0.98\linewidth}
    \footnotesize\raggedright
    For ABC-130K, the size pairs (top view, each wrist view), from most to
    least frequent across training episodes, are
    \mbox{$(292\times219,\,146\times110)$}, \mbox{$(337\times190,\,168\times94)$},
    \mbox{$(320\times200,\,160\times100)$}, \mbox{$(290\times221,\,145\times110)$},
    and \mbox{$(283\times226,\,142\times114)$}.
  \end{minipage}
\end{table}

\noindent\textbf{LIBERO fine-tuning.}
We fine-tune a single policy on the OpenVLA-filtered LIBERO
dataset~\citep{openvla2024}, jointly covering the Spatial, Object,
Goal, and Long suites.
We evaluate the resulting policy on LIBERO~\citep{liu2023libero}
for in-domain performance and on LIBERO-Plus~\citep{liberoplus2025}
for out-of-distribution generalization.
Following the official zero-shot evaluation protocol, we use the
same LIBERO-fine-tuned policy on LIBERO-Plus without further training.
Table~\ref{tab:e2ewam-libero-hyperparameters} summarizes the
fine-tuning and inference settings.

\noindent\textit{\underline{Evaluation instructions.}}
We retain the instruction strings produced by the released
LIBERO-Plus evaluation code.
For non-language perturbations, this implementation derives
instructions from task filenames, inadvertently appending
task-irrelevant perturbation metadata.
For example, the instruction ``open the top drawer and put the
bowl inside'' receives the suffix
``\texttt{view 0 0 100 0 0 initstate 4}''.
This issue has also been documented in prior
work~\citep{yan2026flexpi,yin2026simplememvla}.
Although correcting the instruction extraction substantially
improves performance in our diagnostic evaluations, we leave
these strings unchanged in all reported benchmark results.
This preserves consistency with the released protocol and
comparability with methods evaluated under the same instruction
handling.

We further find that the extraneous text can distort the
final-layer representations of the frozen UMT5-XXL encoder.
Under this interference, instructions for different tasks can
have higher cosine similarity than instructions referring to
the same task.
This ambiguity weakens task discrimination and is associated
with reduced success on Goal tasks.
The ambiguity is already present in the frozen text embeddings
before they enter the world--action model, independently of
our learned visual representations.
To mitigate this effect while preserving the released evaluation
instructions, we use the output of the 16th Transformer block
during both LIBERO fine-tuning and evaluation.
The text encoder remains frozen throughout.
As language representations are not the focus of this study, we leave
further exploration of more effective and stable language embeddings
to future work.

\begin{table}[!htbp]
  \centering
  \small
  \caption{LIBERO fine-tuning and inference settings.}
  \label{tab:e2ewam-libero-hyperparameters}
  \begin{tabular}{@{}>{\raggedright\arraybackslash}p{0.41\linewidth}
                     >{\raggedright\arraybackslash}p{0.55\linewidth}@{}}
    \toprule
    Hyperparameter & Setting \\
    \midrule
    Optimizer & AdamW \\
    Global batch size & 128 clips \\
    Optimizer updates & 30,000 \\
    Learning rate schedule & Warmup: 1,500 updates \\
      & Constant: 22,500 updates \\
      & Cosine decay: 6,000 updates \\
    Peak learning rate & $2\times10^{-4}$ \\
    Final learning rate & $10^{-6}$ \\
    Adam coefficients & $(\beta_1,\beta_2)=(0.9,0.95)$, $\epsilon=10^{-8}$ \\
    Weight decay & 0.01 \\
    Gradient clip norm & 1.0 \\
    Visual encoder & Trainable throughout; base learning rate \\
    Target encoder EMA momentum & 0.9999 \\
    Visual loss weight & 1.25 \\
    Action loss weight & 1.0 \\
    \midrule
    Action chunk size & 16 steps \\
    Frame stride & 4 steps \\
    History frames & 1 current frame \\
    Future frames & 4 future frames \\
    Text features & Frozen UMT5-XXL; 16th Transformer block \\
    Input image & $384\times384$; two $192\times192$ views \\
    Actions executed per prediction & 16 \\
    \bottomrule
  \end{tabular}
\end{table}

\FloatBarrier

\noindent\textbf{RoboTwin fine-tuning.}
We fine-tune a single policy across all 50 RoboTwin~2.0 tasks, drawing
all training clips from the 50 released Clean demonstrations per task
(2,500 episodes in total).
As in prior work~\citep{lilawam}, we remove stationary segments from the
demonstrations before sampling clips.
Following the official evaluation protocol~\citep{robotwin2_2025},
we evaluate each task over 100 rollouts in each of the Clean and
Randomized settings.

Since the predictor and action expert have both been pretrained, we
can update the visual encoder jointly with them from the start of
fine-tuning. All three modules share the same 1,500-step linear
learning-rate warmup.

\noindent\textit{\underline{Action space and normalization.}}
For each arm, we predict joint offsets and end-effector poses relative
to the start of the action chunk, together with absolute gripper openness.
We apply per-dimension $z$-score normalization to joint offsets and
end-effector translations, use the 6D rotations directly, and map
gripper openness from $[0,1]$ to $[-1,1]$.
At execution, we add the denormalized joint offsets to the joint
positions observed at the start of the chunk. This yields a 14D command
comprising six joint-position targets and one gripper command per arm.

\begin{table}[!htbp]
  \centering
  \small
  \caption{RoboTwin fine-tuning hyperparameters.}
  \label{tab:e2ewam-robotwin-hyperparameters}
  \begin{tabular}{@{}>{\raggedright\arraybackslash}p{0.41\linewidth}
                     >{\raggedright\arraybackslash}p{0.55\linewidth}@{}}
    \toprule
    Hyperparameter & Setting \\
    \midrule
    Optimizer                          & AdamW \\
    Global batch size                  & 256 clips \\
    Optimizer updates                  & 30,000 \\
    Learning rate schedule             & Warmup: 1,500 updates \\
                                       & Constant: 22,500 updates \\
                                       & Cosine decay: 6,000 updates \\
    Peak learning rate                 & $10^{-4}$ \\
    Final learning rate                & $10^{-6}$ \\
    Weight decay                       & 0.01 \\
    Gradient clip norm                 & 1.0 \\
    Target encoder EMA momentum        & 0.9999 \\
    Visual loss weight                 & 1.25 \\
    Action loss weight                 & 1.0 \\
    \midrule
    Action chunk size                  & 16 steps \\
    Frame stride                       & 4 steps \\
    History length                     & 1 frame (current observation) \\
    Future prediction frames           & 4 \\
    \bottomrule
  \end{tabular}
\end{table}

\clearpage
\section{Supplementary Results for Section 4}

\subsection{Simulation Benchmarks}
\label{app:simulation_results}

\begingroup
\setlength{\intextsep}{4pt} %
\begin{table*}[!ht]
  \begin{minipage}{\textwidth}
    \setlength{\columnwidth}{\linewidth}
    \centering
    \caption{Success rates (\%) on LIBERO and LIBERO-Plus.}
    \label{tab:system_sim}
    \fontsize{8}{9.2}\selectfont
    \setlength{\tabcolsep}{0pt}
    \iflatexml\setlength{\tabcolsep}{3pt}\fi
    \renewcommand{\arraystretch}{1.05}
    \iflatexml
    \newcommand{\pretrainlarge}[1]{\colorbox{black!8}{#1}}
    \else
    \newcommand{\pretrainlarge}[1]{%
      \begingroup
      \setlength{\fboxsep}{0.8pt}%
      \colorbox{black!8}{\makebox[52pt][c]{\strut #1}}%
      \endgroup
    }
    \fi
    \newcommand{\extrapretrain}[1]{%
      \makebox[62pt][c]{#1}%
    }
    \begin{tabular*}{\columnwidth}{@{\extracolsep{\fill}}lccccc@{}}
      \toprule
      \makebox[196pt][l]{Method} &
      \parbox[c]{34pt}{\centering Params\\(B)} &
      \parbox[c]{64pt}{\centering Embodied\\pretrain} & \parbox[c]{62pt}{\centering Extra\\pretrain} &
      \makebox[48pt]{LIBERO} &
      \parbox[c]{48pt}{\centering LIBERO\\Plus} \\
      \midrule
      \multicolumn{6}{@{}l}{\textit{Vision--language--action models (VLA)}}
  \\
      \addlinespace[3pt]
      $\pi_0$~\citep{black2025pi0}
        & 3.3 & \pretrainlarge{10,000+\,h} & -- & 94.15 & 53.60 \\
      $\pi_0$-FAST~\citep{pertsch2025fast}
        & 3.0 & \pretrainlarge{10,000+\,h} & -- & 85.50 & 61.60 \\
      RIPT-VLA~\citep{tan2025riptvla}
        & 7.6 & 970K\,eps & -- & 97.50 & 68.40 \\
      OpenVLA-OFT~\citep{kim2025oft}
        & 7.6 & 970K\,eps & -- & 97.10 & 69.60 \\
      X-VLA~\citep{zheng2025xvla}
        & 0.9 & 290K\,eps & -- & 98.10 & 71.40 \\
      AVA-VLA~\citep{xiao2026avavla}
        & 7.7 & 970K\,eps & -- & 98.20 & 74.70 \\
      VLA-JEPA~\citep{sun2026vlajepa}
        & 2.3 & 296K\,eps & -- & 97.20 & 79.50 \\
      ABot-M0~\citep{yang2026abotm0}
        & 4.6 & 9,500+\,h & -- & 98.60 & 80.50 \\
      $\pi_{0.5}$~\citep{physicalintelligence2025pi05}
        & 3.4 & \pretrainlarge{10,000+\,h} & \extrapretrain{NR\,VL} & 96.85 & 84.40 \\
      Qwen-RobotManip Context~\citep{yuan2026qwenrobotmanip}
        & 4.8 & \pretrainlarge{38,100\,h} & \extrapretrain{28M\,VL} & 99.20 & 91.40 \\
      \midrule
      \multicolumn{6}{@{}l}{\textit{World--action models (WAM)}} \\
      \addlinespace[3pt]
      WorldVLA~\citep{cen2025worldvla}
        & 7.0 & 0\,h & -- & 79.10 & 25.00 \\
      Fast-WAM~\citep{fastwam2026}
        & 6.0 & 0\,h & -- & 97.60 & 51.50 \\
      OpenWAM-$\alpha$~\citep{openwam2026}
        & 6.0 & 6,369\,h & -- & 99.30 & 69.20 \\
      Faster-WAM~\citep{ma2026fasterwam}
        & 5.0 & 0\,h & -- & 98.30 & 78.30 \\
      GE-Act~\citep{genie2025}
        & 2.2 & 2,967\,h & -- & 94.40 & 80.30 \\
      Being-H0.7~\citep{luo2026beingh07}
        & 3.0 & \pretrainlarge{215,000\,h} & \extrapretrain{NR\,VL} & 99.20 & 82.10 \\
      Cosmos Policy~\citep{cosmospolicy2026}
        & 2.0 & 0\,h & -- & 98.50 & 82.20 \\
      InternW0-$\Delta$~\citep{miao2026internw0delta}
        & 8.4 & \pretrainlarge{20,000+\,h} & -- & -- & 92.80 \\
      \midrule
      \textbf{E2EWAM (Ours)} & \textbf{1.6} & \textbf{6,383\,h} & -- & \textbf{98.35} & \textbf{84.46} \\
      \bottomrule
    \end{tabular*}
    \par\vspace{2pt}
    {\fontsize{7}{8.4}\selectfont\raggedright
      Gray: $\geq$10K h or $\geq$1M eps, using separate thresholds.
      K/M denote thousand/million.\par
      Extra pretrain: VL = vision--language samples; NR = amount not reported.\par
      We omit UniWAM because its reported LIBERO-Plus evaluation\footnotemark{}
      did not follow the benchmark's official \mbox{zero-shot protocol~\citep{liberoplus2025}}.\par}
  \end{minipage}
  \par\vspace{10pt}
  \begin{minipage}{\textwidth}
    \setlength{\columnwidth}{\linewidth}
    \centering
    \caption{Success rates (\%) on RoboTwin~2.0 under Clean-only training.}
    \label{tab:robotwin_sim}
    \fontsize{8}{9.2}\selectfont
    \setlength{\tabcolsep}{0pt}
    \iflatexml\setlength{\tabcolsep}{3pt}\fi
    \renewcommand{\arraystretch}{1.05}
    \iflatexml
    \newcommand{\pretrainlarge}[1]{\colorbox{black!8}{#1}}
    \else
    \newcommand{\pretrainlarge}[1]{%
      \begingroup
      \setlength{\fboxsep}{0.8pt}%
      \colorbox{black!8}{\makebox[52pt][c]{\strut #1}}%
      \endgroup
    }
    \fi
    \newcommand{\extrapretrain}[1]{%
      \makebox[62pt][c]{#1}%
    }
    \begin{tabular*}{\columnwidth}{@{\extracolsep{\fill}}lccccc@{}}
      \toprule
      \makebox[196pt][l]{Method} &
      \parbox[c]{34pt}{\centering Params\\(B)} &
      \parbox[c]{64pt}{\centering Embodied\\pretrain} & \parbox[c]{62pt}{\centering Extra\\pretrain} &
      \makebox[48pt]{C$\to$C} & \makebox[48pt]{C$\to$R} \\
      \midrule
      \multicolumn{6}{@{}l}{
        \textit{Vision--language--action models (VLA)}} \\
      \addlinespace[3pt]
      Xiaomi-Robotics-0~\citep{cai2026xiaomirobotics0}
        & 4.7 & 200M\,ts & \extrapretrain{80M+\,VL} & 62.86 & 18.22 \\
      ABot-M0~\citep{yang2026abotm0}
        & 4.6 & 9,500+\,h & -- & 57.40 & 30.36 \\
      X-VLA~\citep{zheng2025xvla}
        & 0.9 & 290K\,eps & -- & 67.96 & 20.95 \\
      Spatial Forcing~\citep{li2025spatialforcing}
        & 3.4 & \pretrainlarge{10,000+\,h} & \extrapretrain{NR\,VL} & 77.18 & 26.74 \\
      $\pi_{0.5}$~\citep{physicalintelligence2025pi05}
        & 3.4 & \pretrainlarge{10,000+\,h} & \extrapretrain{NR\,VL} & 70.66 & 46.00 \\
      GigaBrain-0.7~\citep{gigabrain2026}
        & 3.5 & \pretrainlarge{37,257\,h} & \extrapretrain{272M\,VL} & 66.80 & 67.92 \\
      Qwen-RobotManip Context~\citep{yuan2026qwenrobotmanip}
        & 4.8 & \pretrainlarge{38,100\,h} & \extrapretrain{28M\,VL} & 84.70 & 69.40 \\
      \midrule
      \multicolumn{6}{@{}l}{
        \textit{World--action models (WAM)}} \\
      \addlinespace[3pt]
      AHA-WAM~\citep{cai2026ahawam}
        & 7.2 & 0\,h & -- & 64.32 & 3.20 \\
      X-WAM~\citep{guo2026xwam}
        & 6.7 & 5,874\,h & -- & 69.98 & 25.84 \\
      LingBot-VA~\citep{lingbotva2026}
        & 5.3 & \pretrainlarge{16,000\,h} & -- & 80.70 & 34.60 \\
      4D-WAM~\citep{yang20264dwam}
        & 5.3 & \pretrainlarge{16,000\,h} & -- & 81.50 & 41.80 \\
      WorldScape Policy 2.0~\citep{su2026worldscape2}
        & 5.6 & 4,983\,h & -- & 84.92 & 38.50 \\
      OpenWAM-$\alpha$~\citep{openwam2026}
        & 6.0 & 6,369\,h & -- & 89.40 & 48.70 \\
      UniWAM~\citep{song2026uniwam}
        & 8.0 & \pretrainlarge{10,013\,h} & \extrapretrain{8.3M\,VL} & 75.14 & 68.32 \\
      ME-DEX1.0~\citep{zhang2026medex}
        & 6.3 & \pretrainlarge{1.01M\,eps}
        & \extrapretrain{23K\,eps\,tac} & 89.58 & 68.12 \\
      InternW0-$\Delta$~\citep{miao2026internw0delta}
        & 8.4 & \pretrainlarge{20,000+\,h} & -- & 90.00 & 71.90 \\
      \midrule
      \textbf{E2EWAM (Ours)}
        & \textbf{1.6} & \textbf{6,383\,h} & -- & \textbf{83.06} & \textbf{56.78} \\
      \bottomrule
    \end{tabular*}
    \par\vspace{2pt}
    {\fontsize{7}{8.4}\selectfont\raggedright
      Gray: $\geq$10K h or $\geq$1M eps, using separate thresholds.
      K/M denote thousand/million.\par
      Extra pretrain: VL = vision--language samples; NR = amount not reported;
      eps = episodes; tac = tactile.\par}
  \end{minipage}
\end{table*}
\footnotetext{\fontsize{7.5}{9}\selectfont
  The authors fine-tuned the evaluated checkpoint on LIBERO-Plus
  (Wenxuan Song, \href{https://github.com/UniWAM/UniWAM/issues/4\#issuecomment-6108212377}{UniWAM GitHub issue~\#4}, October~11, 2026).}

\clearpage
\endgroup
\subsection{Physical Probing}
\label{app:probing}
\label{app:probing-encoder}

Following the physical probing setup in
Section~\ref{sec:experiment-settings}, we evaluate all encoders
using their pretrained weights without fine-tuning on the target
benchmark datasets. This ensures a fair comparison in which
none of the encoders has been trained on these datasets.
All encoders remain frozen throughout probing.

We replay the official demonstration trajectories in simulation
to collect privileged physical states and fit a separate linear
predictor for each physical property from the encoder representations.
We independently repeat probe fitting and evaluation over five
random episode-level train/test splits.
For RoboTwin 2.0, we fit separate predictors for the left and right
arms for each arm-specific property and average their corresponding
$R^2$ scores within each split.
Table~\ref{tab:encoder-probing} reports the mean and sample standard
deviation of $R^2$ across the five splits.

\begin{table*}[!htbp]
\centering
\caption{\textbf{Physical probing results.}
Linear $R^2\!\uparrow$.}
\label{tab:encoder-probing}
\begingroup
\fontsize{7}{9}\selectfont
\setlength{\tabcolsep}{1.35pt}
\setlength{\aboverulesep}{1.4pt}
\setlength{\belowrulesep}{1.4pt}
\newcommand{\encProbe}[2]{#1\,\ensuremath{\pm}\,#2}
\newcommand{\encProbeBest}[2]{\encProbe{\textbf{#1}}{\textbf{#2}}}
\newsavebox{\encoderProbingTable}
\iflatexml\else
\begin{lrbox}{\encoderProbingTable}
\fi
\begin{tabular}{@{}l*{8}{c}@{}}
\toprule
 & \multicolumn{4}{c}{Static} & \multicolumn{4}{c@{}}{Dynamic} \\
\cmidrule(lr){2-5}\cmidrule(lr){6-9}
Model & EEF & Arm joint & Gripper joint & Object
      & EEF & Arm joint & Gripper joint & Object \\
\midrule
\multicolumn{9}{@{}l}{\textit{(a) LIBERO}} \\
\addlinespace[1pt]
Wan2.2 VAE & \encProbe{0.826}{0.002} & \encProbe{0.330}{0.007} & \encProbe{0.294}{0.005} & \encProbe{0.972}{0.001} & \encProbe{0.358}{0.014} & \encProbe{0.260}{0.010} & \encProbe{0.175}{0.009} & \encProbe{0.610}{0.010} \\
V-JEPA~2.1 & \encProbe{0.986}{0.000} & \encProbe{0.889}{0.002} & \encProbe{0.875}{0.007} & \encProbe{0.992}{0.000} & \encProbe{0.811}{0.009} & \encProbe{0.637}{0.010} & \encProbe{0.763}{0.019} & \encProbe{0.765}{0.006} \\
DINOv3 & \encProbe{0.988}{0.000} & \encProbe{0.896}{0.003} & \encProbe{0.891}{0.005} & \encProbe{0.994}{0.001} & \encProbe{0.840}{0.005} & \encProbe{0.663}{0.009} & \encProbe{0.781}{0.024} & \encProbe{0.816}{0.007} \\
VC-1 & \encProbe{0.984}{0.000} & \encProbe{0.875}{0.002} & \encProbe{0.878}{0.004} & \encProbe{0.994}{0.000} & \encProbe{0.818}{0.005} & \encProbe{0.640}{0.003} & \encProbe{0.778}{0.021} & \encProbe{0.819}{0.013} \\
R3M & \encProbe{0.992}{0.000} & \encProbe{0.907}{0.006} & \encProbe{0.854}{0.009} & \encProbeBest{0.996}{0.000} & \encProbe{0.853}{0.006} & \encProbe{0.703}{0.007} & \encProbe{0.749}{0.025} & \encProbeBest{0.857}{0.008} \\
CAIP & \encProbe{0.983}{0.000} & \encProbe{0.864}{0.004} & \encProbe{0.812}{0.005} & \encProbe{0.993}{0.001} & \encProbe{0.814}{0.006} & \encProbe{0.622}{0.009} & \encProbe{0.712}{0.020} & \encProbe{0.809}{0.010} \\
\addlinespace[1pt]
\textbf{E2EWAM} & \encProbeBest{0.995}{0.000} & \encProbeBest{0.936}{0.002} & \encProbeBest{0.934}{0.003} & \encProbe{0.996}{0.000} & \encProbeBest{0.892}{0.004} & \encProbeBest{0.740}{0.006} & \encProbeBest{0.859}{0.014} & \encProbe{0.848}{0.008} \\
\midrule
\multicolumn{9}{@{}l}{\textit{(b) LIBERO-Plus}} \\
\addlinespace[1pt]
Wan2.2 VAE & \encProbe{0.575}{0.004} & \encProbe{0.182}{0.001} & \encProbe{0.142}{0.008} & \encProbe{0.945}{0.001} & \encProbe{0.212}{0.003} & \encProbe{0.159}{0.004} & \encProbe{0.095}{0.008} & \encProbe{0.464}{0.012} \\
V-JEPA~2.1 & \encProbe{0.967}{0.000} & \encProbe{0.803}{0.006} & \encProbe{0.855}{0.001} & \encProbe{0.986}{0.000} & \encProbe{0.752}{0.003} & \encProbe{0.578}{0.006} & \encProbe{0.726}{0.008} & \encProbe{0.719}{0.012} \\
DINOv3 & \encProbe{0.970}{0.000} & \encProbe{0.811}{0.008} & \encProbe{0.871}{0.002} & \encProbe{0.988}{0.001} & \encProbe{0.771}{0.003} & \encProbe{0.589}{0.006} & \encProbe{0.741}{0.013} & \encProbe{0.766}{0.008} \\
VC-1 & \encProbe{0.955}{0.001} & \encProbe{0.767}{0.005} & \encProbe{0.826}{0.003} & \encProbe{0.987}{0.000} & \encProbe{0.740}{0.006} & \encProbe{0.560}{0.007} & \encProbe{0.717}{0.006} & \encProbe{0.756}{0.011} \\
R3M & \encProbe{0.969}{0.000} & \encProbe{0.813}{0.005} & \encProbe{0.771}{0.007} & \encProbe{0.991}{0.000} & \encProbe{0.768}{0.005} & \encProbe{0.601}{0.008} & \encProbe{0.655}{0.011} & \encProbe{0.779}{0.011} \\
CAIP & \encProbe{0.948}{0.000} & \encProbe{0.735}{0.005} & \encProbe{0.685}{0.006} & \encProbe{0.986}{0.000} & \encProbe{0.682}{0.004} & \encProbe{0.523}{0.007} & \encProbe{0.589}{0.009} & \encProbe{0.742}{0.004} \\
\addlinespace[1pt]
\textbf{E2EWAM} & \encProbeBest{0.982}{0.001} & \encProbeBest{0.862}{0.006} & \encProbeBest{0.910}{0.002} & \encProbeBest{0.992}{0.001} & \encProbeBest{0.831}{0.002} & \encProbeBest{0.663}{0.001} & \encProbeBest{0.815}{0.006} & \encProbeBest{0.795}{0.009} \\
\midrule
\multicolumn{9}{@{}l}{\textit{(c) RoboTwin (Randomized)}} \\
\addlinespace[1pt]
Wan2.2 VAE & \encProbe{0.100}{0.007} & \encProbe{0.076}{0.007} & \encProbe{0.047}{0.012} & \encProbe{0.157}{0.066} & \encProbe{0.195}{0.004} & \encProbe{0.228}{0.010} & \encProbe{0.008}{0.004} & \encProbe{0.105}{0.038} \\
V-JEPA~2.1 & \encProbe{0.808}{0.005} & \encProbe{0.690}{0.026} & \encProbe{0.581}{0.012} & \encProbe{0.408}{0.026} & \encProbe{0.612}{0.017} & \encProbe{0.629}{0.010} & \encProbe{0.266}{0.014} & \encProbe{0.400}{0.028} \\
DINOv3 & \encProbe{0.761}{0.004} & \encProbe{0.643}{0.022} & \encProbe{0.577}{0.014} & \encProbe{0.428}{0.025} & \encProbe{0.523}{0.005} & \encProbe{0.536}{0.008} & \encProbe{0.310}{0.029} & \encProbe{0.361}{0.037} \\
VC-1 & \encProbe{0.740}{0.006} & \encProbe{0.624}{0.017} & \encProbe{0.463}{0.005} & \encProbe{0.386}{0.028} & \encProbe{0.549}{0.012} & \encProbe{0.573}{0.008} & \encProbe{0.198}{0.009} & \encProbe{0.403}{0.038} \\
R3M & \encProbe{0.528}{0.009} & \encProbe{0.397}{0.008} & \encProbe{0.268}{0.013} & \encProbe{0.203}{0.126} & \encProbe{0.277}{0.013} & \encProbe{0.281}{0.012} & \encProbe{0.059}{0.006} & \encProbe{0.077}{0.054} \\
CAIP & \encProbe{0.595}{0.011} & \encProbe{0.486}{0.021} & \encProbe{0.369}{0.009} & \encProbe{0.331}{0.035} & \encProbe{0.370}{0.012} & \encProbe{0.407}{0.009} & \encProbe{0.097}{0.018} & \encProbe{0.256}{0.040} \\
\addlinespace[1pt]
\textbf{E2EWAM} & \encProbeBest{0.850}{0.005} & \encProbeBest{0.728}{0.027} & \encProbeBest{0.601}{0.006} & \encProbeBest{0.431}{0.027} & \encProbeBest{0.680}{0.014} & \encProbeBest{0.715}{0.006} & \encProbeBest{0.313}{0.010} & \encProbeBest{0.412}{0.030} \\
\bottomrule
\end{tabular}
\iflatexml\else
\end{lrbox}
\usebox{\encoderProbingTable}
\fi
\par\vspace{2pt}
\iflatexml\else
\begin{minipage}{\wd\encoderProbingTable}
\fi
\fontsize{7}{8.5}\selectfont\raggedright
Mean $\pm$ sample SD over five episode splits; probes fitted separately within each domain.
Bold: best before rounding, per metric and domain.
\iflatexml\else
\end{minipage}
\fi
\endgroup
\end{table*}

\clearpage

\begingroup
\edef\savedSupplementaryFigureCounter{\number\value{figure}}
\setcounter{figure}{0}
\renewcommand{\thefigure}{\thesection\arabic{figure}}
\renewcommand{\theHfigure}{appendix.\thesection.\arabic{figure}}
\subsection{Real-World Progress Scores}
\label{app:real_progress}

\noindent\textbf{Progress scores.}
Figure~\ref{fig:real-robot-progress} reports partial completion for the
same five tasks and 12 task--condition pairs as the success rates in
Fig.~\ref{fig:real-robot-results}. A rollout receives full credit only
when it meets the task's success criterion.

\begin{figure}[!ht]
  \centering
  \captionsetup{skip=4pt}
  \includegraphics[width=\linewidth]{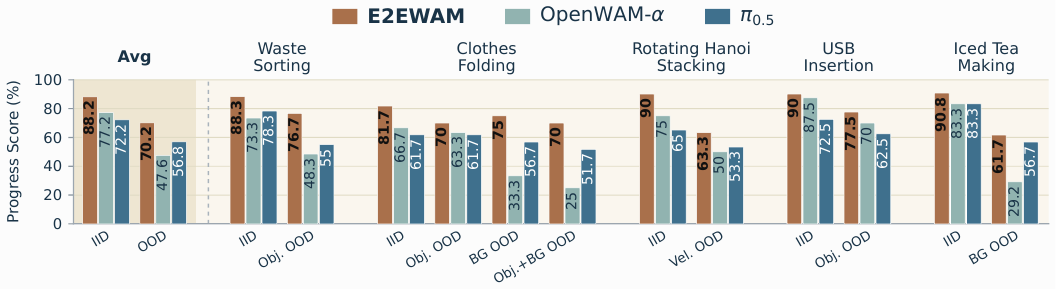}
  \caption{\textbf{Real-robot progress scores.}
  Progress score (PS) across five tasks and 12 task--condition pairs.
  Shaded groups report task-balanced averages: Avg IID averages the
  five IID settings; Avg OOD first averages OOD conditions within each
  task, then averages over tasks. Methods, conditions, and abbreviations
  follow Fig.~\ref{fig:real-robot-results}.
  Bar labels report scores in percent, rounded to one decimal where needed.}
  \label{fig:real-robot-progress}
\end{figure}

\FloatBarrier

\setcounter{figure}{\savedSupplementaryFigureCounter}
\endgroup

\clearpage
\begingroup
\edef\savedRealFigureCounter{\number\value{figure}}
\setcounter{figure}{0}
\renewcommand{\thefigure}{\thesection\arabic{figure}}
\renewcommand{\theHfigure}{appendix.\thesection.\arabic{figure}}
\section{Real-World Experimental Details}
\label{app:real_protocol}
\subsection{Robot Platform}
\label{app:robot_platform}

\noindent
\begin{minipage}[t]{0.44\linewidth}
\vspace{0pt}
\paragraph{Platform.}
We use an I2RT YAM bimanual robot for all five tasks
(Figure~\ref{fig:yam_platform}). Each arm has six revolute joints and
a linear gripper, giving a 14-D joint-position command interface.
One external RealSense D455 and two wrist-mounted D405 cameras capture
RGB at $640\times480$ and 30 FPS. All three views are used for policy
input and recording; depth is not used.

\paragraph{Observations.}
All models receive the same views and task-instruction text, with
model-specific preprocessing (Table~\ref{tab:real_execution_settings}).
Models retain their native state encodings and language templates.
Camera streams are aligned in software using host timestamps and causal
frame selection.
Each policy query uses fresh images from all three cameras and the
current robot state.
\end{minipage}\hfill
\begin{minipage}[t]{0.52\linewidth}
\vspace{0pt}
\centering
\iflatexml\includegraphics[width=\linewidth]{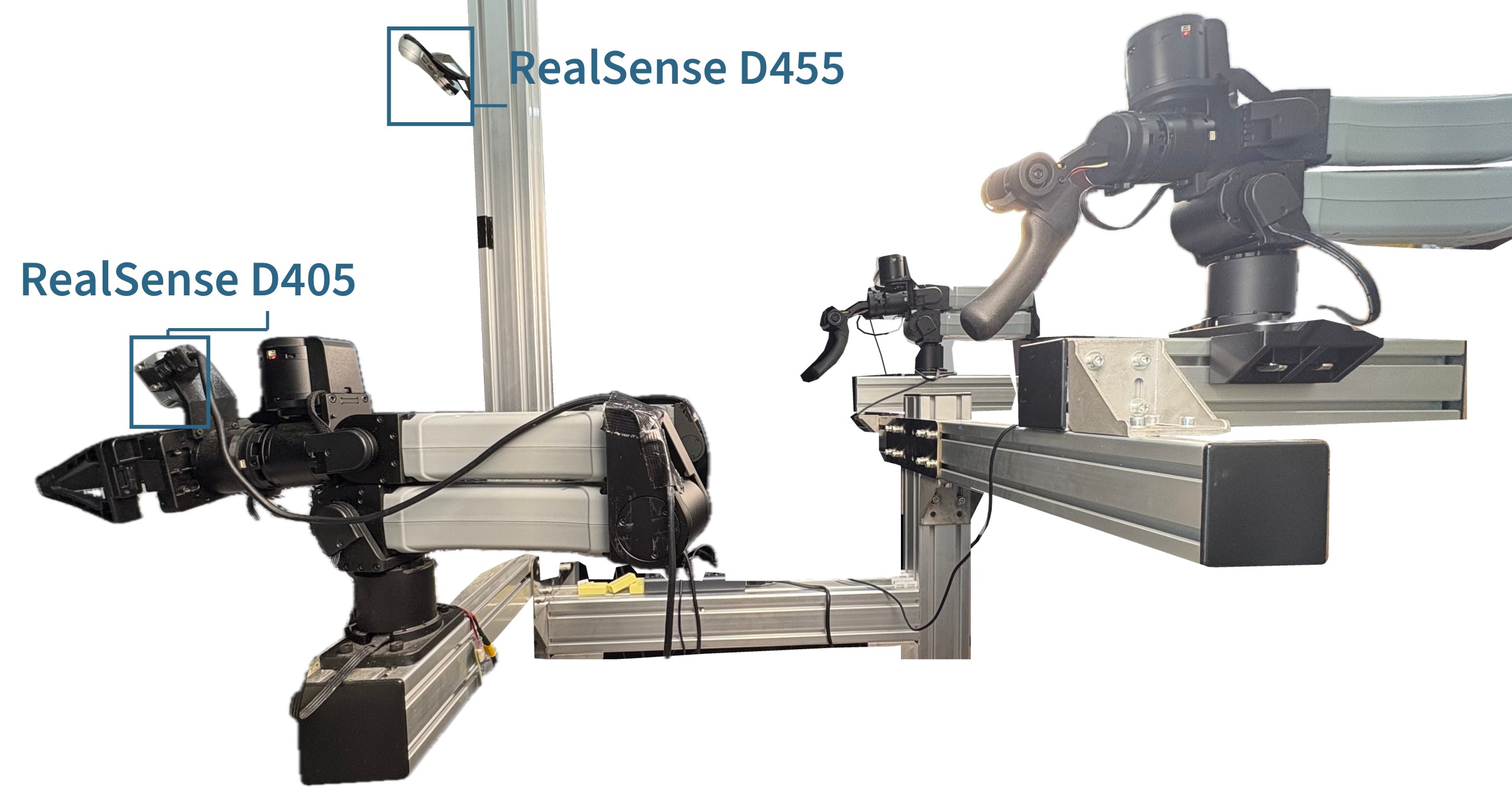}\else\includegraphics[width=\linewidth]{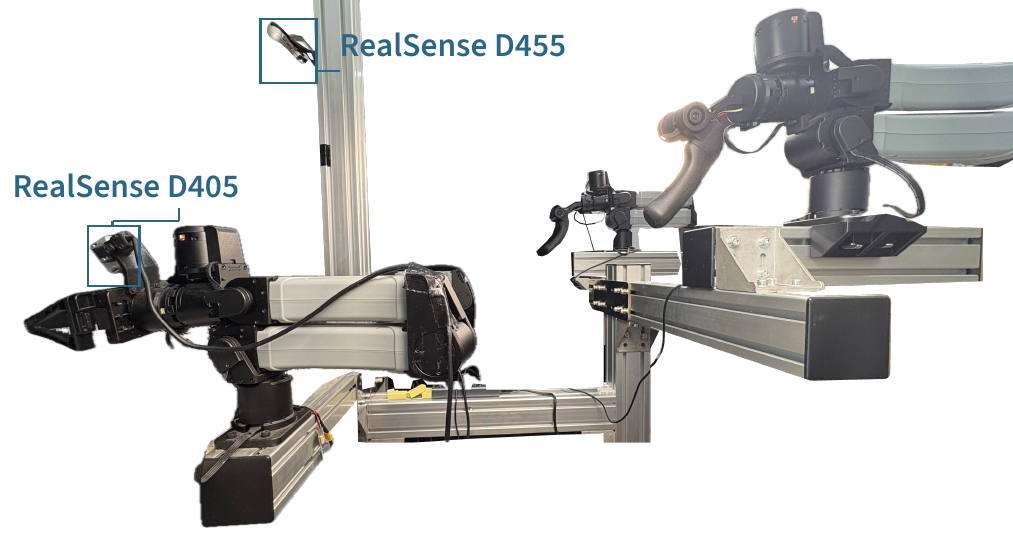}
\fi
\captionsetup{hypcap=false}
\captionof{figure}{YAM apparatus.}
\label{fig:yam_platform}
\end{minipage}

\FloatBarrier
\subsection{Policy Fine-Tuning}
\label{app:real_training}

We fine-tune a separate policy for each task, initializing E2EWAM from
our 125k-step pretrained checkpoint and the baselines from the released
OpenWAM-$\alpha$ and $\pi_{0.5}$ checkpoints. All methods use a global
batch of 64 training windows.
Model-specific fine-tuning settings are summarized in Table~\ref{tab:real_execution_settings}.

\paragraph{Fine-tuning protocol.}
For each task, all methods use the same demonstration data, with model-specific preprocessing
and optimization settings. We tune the key parameters, including the learning rate separately for each method and task
to select the best-performing configuration.

\paragraph{Image augmentation.}
E2EWAM uses no image augmentation for any of the five tasks.
For the two baselines, we follow their official image augmentation settings.
OpenWAM-$\alpha$ uses ColorJitter$(0.2,0.2,0.2,0)$ for all five tasks.
$\pi_{0.5}$ uses ColorJitter$(0.3,0.4,0.5)$ on all views, with a random
crop to 95\% of the image width and height, resizing, and rotation within
$\pm5^\circ$ on the external view only. Augmentation is disabled at evaluation.

\paragraph{Execution.}
Policies run on a local RTX 4090 (24 GB) and communicate with the executor
over TCP. Rotating Hanoi Stacking tests whether the robot can time ring
placement on a moving post. With synchronous execution, differences in
inference latency would lead to different pauses between action chunks
and could affect placement success. To reduce this effect, we keep the
inference rate, action execution rate, and all settings for asynchronous
action prefetching identical across the three policies. For the other
four tasks, the robot executes each 32-action chunk at a nominal
30\,Hz before querying the policy again.
(Table~\ref{tab:real_execution_settings}).

\begin{table}[ht]
  \centering
  \caption{Model-specific settings for real-robot execution and fine-tuning. Image dimensions are height $\times$ width.}
  \label{tab:real_execution_settings}
  \small
  \setlength{\tabcolsep}{4pt}
  \renewcommand{\arraystretch}{1.12}
  \begin{tabular}{@{}>{\raggedright\arraybackslash}p{0.16\linewidth}>{\raggedright\arraybackslash}p{0.255\linewidth}>{\raggedright\arraybackslash}p{0.255\linewidth}>{\raggedright\arraybackslash}p{0.255\linewidth}@{}}
    \toprule
    Setting & E2EWAM & OpenWAM-$\alpha$ & $\pi_{0.5}$ \\
    \midrule
    RGB input & $384\times384$ composite & $384\times320$ composite
      & Three $224\times224$ images \\
    Resizing & Preserve aspect ratio; pad & Direct resize
      & Preserve aspect ratio; pad \\
    State input & No network state conditioning
      & 20-D end-effector and gripper state
      & 14-D joint and gripper state \\
    Action & 12 relative joint targets + 2 absolute gripper commands
      & 20-D absolute end-effector and gripper commands
      & 12 relative joint targets + 2 absolute gripper commands \\
    Sampling steps & 20 & 10 & 10 \\
    Global batch size & 64 & 64 & 64 \\
    Peak learning rate & $5\times10^{-5}$ & $10^{-4}$ & $2.5\times10^{-5}$ \\
    Weight decay & $0.01$ & $0.01$ & $10^{-10}$ \\
    Warmup & 5\% of scheduled steps & 5\% of scheduled steps & 1,000 steps \\
    LR schedule & Cosine to $10^{-6}$ & Cosine to $10^{-6}$ & Cosine to $2.5\times10^{-6}$ over 30k steps \\
    Image augmentation & None & ColorJitter & ColorJitter; crop and rotation on the external view \\
    \bottomrule
  \end{tabular}
\end{table}

\FloatBarrier

\clearpage
\subsection{Task Details and Failure Analysis}
\label{app:real_demonstrations}
\label{app:real_task_analysis}
\label{app:real_supplementary}

We collected 545 successful demonstrations (5.70 hours) by teleoperating
the YAM. All three policies use the same demonstrations and window-sampling
plan for each task. Table~\ref{tab:real_demonstrations} lists the counts and
durations, and Fig.~\ref{fig:real_ood_scenes} shows the in-domain (ID) and
out-of-distribution (OOD) conditions. We report success rates in
Fig.~\ref{fig:real-robot-results} and progress scores in
Fig.~\ref{fig:real-robot-progress} using the criteria in Appendix~\ref{app:real_scoring}.

\par\medskip
\noindent\begin{minipage}{\linewidth}
\centering
\iflatexml\includegraphics[width=.93\linewidth]{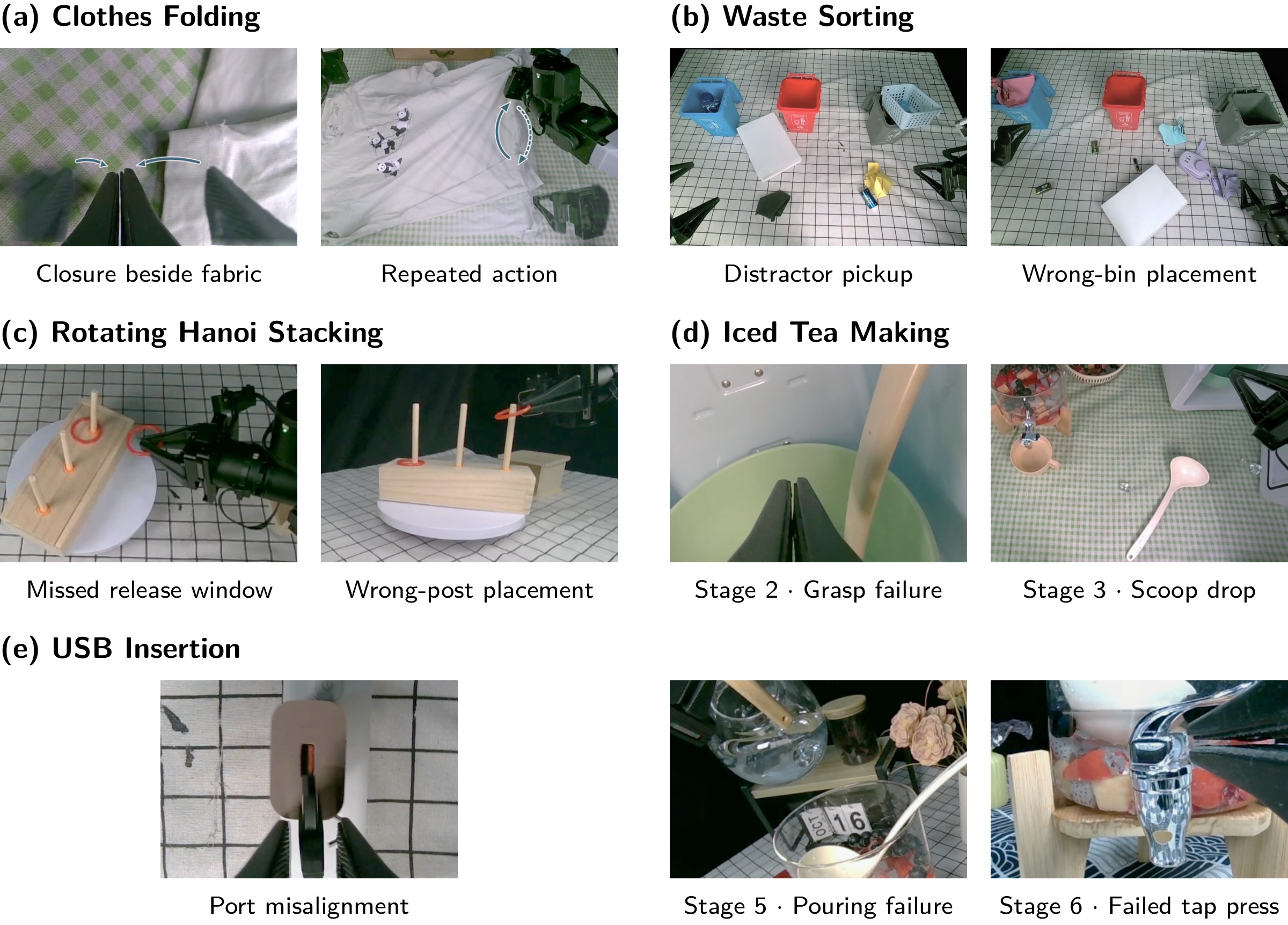}\else\includegraphics[width=.93\linewidth]{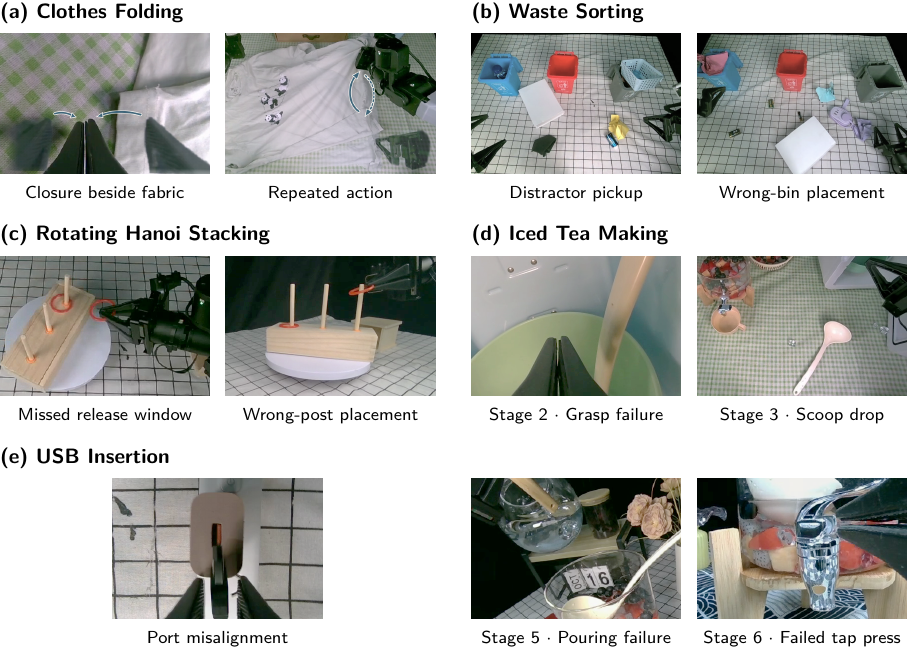}\fi\par
\captionsetup{hypcap=false,skip=8pt}
\captionof{figure}{\textbf{Representative real-robot failure modes.}
}
\label{fig:real_failure_modes}
\end{minipage}\par\smallskip

\newcommand{\taskdetailheading}[1]{\noindent\textit{\underline{#1}.}\enspace\ignorespaces}

\begingroup
\setlength{\smallskipamount}{1.5pt plus 1pt minus 1pt}
\par\addvspace{0.3\baselineskip}
\noindent\begin{minipage}{\linewidth}
\hrule height 0.35pt\relax\vspace{2pt}
\noindent\textbf{Clothes Folding}\par\smallskip
\taskdetailheading{Task setup} The robot unfolds and spreads the shirt flat,
folds the sleeves and sides inward, brings up the hem, and makes the final
folds crosswise. Demonstrations and ID trials use the same four printed
shirts on a white grid-patterned tablecloth
(Fig.~\ref{fig:real_ood_scenes}(a)). The robot must distinguish the shirt
from the background to locate a usable edge or corner for each grasp and
infer from the shirt's shape which steps are complete and what action should
come next.\par\smallskip
\taskdetailheading{OOD settings} We evaluate object OOD with four held-out
shirts in the original scene, background OOD with the training shirts in
four held-out tabletop scenes, and combined OOD with held-out shirts in
held-out scenes (Fig.~\ref{fig:real_ood_scenes}(a)). The object shift varies
shirt color and print. The background shift varies the tablecloth's color
and pattern as well as the objects surrounding the garment and their
placement. These conditions test whether the policy can still locate usable
grasp points and recognize the current folding stage under visual
shifts.\par\smallskip
\taskdetailheading{Failure analysis} Across failed rollouts, a common grasp
error was closing the gripper beside the fabric or catching the wrong edge,
indicating difficulty localizing the shirt boundary. Another recurring
pattern was repeating the same action after the shirt had reached the
intended intermediate state, suggesting difficulty recognizing that a step
was complete (Fig.~\ref{fig:real_failure_modes}(a)). E2EWAM more consistently secured usable edges and advanced
once a step was complete. Together with its higher progress scores across
all three OOD settings (Fig.~\ref{fig:real-robot-progress}), these behaviors
suggest more reliable grasp localization and better tracking of the
garment's folding state under visual shifts.
\end{minipage}\par

\par\addvspace{0.3\baselineskip}
\noindent\begin{minipage}{\linewidth}
\hrule height 0.35pt\relax\vspace{2pt}
\noindent\textbf{Waste Sorting}\par\smallskip
\taskdetailheading{Task setup} The robot picks out bottles, batteries, and
paper balls from a cluttered tabletop and places each item in its designated
bin along the far side of the table.
Demonstrations and ID trials draw from three bottle types, three battery
types, and three paper-ball colors (Fig.~\ref{fig:real_ood_scenes}(b)).
The robot must distinguish waste from distractors to select what to pick
up and recognize each selected item's category to choose where to place
it.\par\smallskip
\taskdetailheading{OOD settings} We evaluate object OOD using a pool of
three held-out variants per category. Bottle and battery appearances
change and the paper balls come in different colors
(Fig.~\ref{fig:real_ood_scenes}(b)). This tests whether the robot can
still pick out waste in clutter and place unfamiliar instances into the
bins assigned to their categories.\par\smallskip
\taskdetailheading{Failure analysis} We observed two notable errors during
sorting: attempting to grasp a distractor and placing a waste item in the
wrong bin after a correct pickup
(Fig.~\ref{fig:real_failure_modes}(b)). Both reflect mismatches in semantic
associations, either between objects and waste categories or between
categories and their designated bins. Across 20 Object OOD rollouts per
policy, semantic errors were observed in 2 E2EWAM rollouts,
16 OpenWAM-$\alpha$ rollouts, and 5 $\pi_{0.5}$ rollouts. The lower
incidence for E2EWAM and $\pi_{0.5}$ suggests that both retained more
reliable semantic understanding when object appearance changed.
Consequently, E2EWAM pushed task progress to later stages and completed
more categories (Fig.~\ref{fig:real-robot-progress}). Aside from semantic
mismatches, its remaining failures occurred when the robot correctly
sorted one or two categories but did not continue sorting, whereas
OpenWAM-$\alpha$ more often stalled early due to such mismatches.
\end{minipage}\par

\par\addvspace{0.3\baselineskip}
\noindent\begin{minipage}{\linewidth}
\hrule height 0.35pt\relax\vspace{2pt}
\noindent\textbf{Rotating Hanoi Stacking}\par\smallskip
\taskdetailheading{Task setup} While the turntable rotates, the robot picks
up a ring from a stationary wooden block and places it over the post that
already holds a ring. Demonstrations and ID trials use red or blue rings
and two rotation speeds, low and medium
(Fig.~\ref{fig:real_ood_scenes_continued}(c)). The task tests whether the
robot can infer the turntable's rotation speed from successive visual
observations and anticipate the target post's position during
placement.\par\smallskip
\taskdetailheading{OOD settings} We evaluate velocity OOD at a held-out
speed faster than either ID setting. The robot must use its observations
of the target's motion to adapt the position and timing of ring placement
to the unseen speed.\par\smallskip
\taskdetailheading{Failure analysis} In the most representative failure mode,
the robot misjudged the turntable's rotation speed and released the ring at
the wrong time, causing the ring to miss the target post
(Fig.~\ref{fig:real_failure_modes}(c)). E2EWAM completed the task in 45\% of trials
at the held-out speed, compared with 25\% for OpenWAM-$\alpha$ and 30\%
for $\pi_{0.5}$
(Fig.~\ref{fig:real-robot-results}), suggesting better generalization of
moving-target placement to the unseen speed.
\end{minipage}\par

\par\addvspace{0.3\baselineskip}
\noindent\begin{minipage}{\linewidth}
\hrule height 0.35pt\relax\vspace{2pt}
\noindent\textbf{Iced Tea Making}\par\smallskip
\taskdetailheading{Task setup} The robot opens the refrigerator, retrieves
the scoop of ice, and empties it into the dispenser. It then picks up the
kettle, pours water into the same dispenser, and presses the tap to
dispense the drink into the serving cup. Demonstrations and ID trials use
a white grid-patterned tablecloth with a seen set of distractors
(Fig.~\ref{fig:real_ood_scenes_continued}(d)). The robot must locate the
relevant contact points, such as the scoop handle and the tap, and track
which steps are complete so it can select the next tool or
action.\par\smallskip
\taskdetailheading{OOD settings} We evaluate four held-out scenes that
vary the color and pattern of the tablecloth together with the surrounding
distractors (Fig.~\ref{fig:real_ood_scenes_continued}(d)). The task
equipment stays the same. This tests whether the robot can still find and
operate the familiar tools and controls in the correct order under
changes in the surrounding scene.\par\smallskip
\taskdetailheading{Failure analysis} A recurring early-stage failure was
opening the refrigerator but failing to lift out the scoop. A
representative late-stage error was failing to press the dispenser tap
correctly after adding both ice and water, leaving the serving cup empty (Fig.~\ref{fig:real_failure_modes}(d)).
E2EWAM retrieved the scoop in all 20 OOD trials, compared with 10 for
OpenWAM-$\alpha$, showing more reliable scoop retrieval under the scene
shift. Its mean progress score was 61.7\%, compared with 56.7\% for
$\pi_{0.5}$ and 29.2\% for OpenWAM-$\alpha$
(Fig.~\ref{fig:real-robot-progress}), indicating that it advanced farther
through the ordered sequence on average.
\end{minipage}\par

\par\addvspace{0.3\baselineskip}
\noindent\begin{minipage}{\linewidth}
\hrule height 0.35pt\relax\vspace{2pt}
\noindent\textbf{USB Insertion}\par\smallskip
\taskdetailheading{Task setup} The robot unplugs a USB connector from a
power strip and inserts it into a charging adapter on the same strip.
Demonstrations and ID trials use two power strips and three charging
adapters (Fig.~\ref{fig:real_ood_scenes_continued}(e)). The robot must
locate the small receiving port and adjust the plug's position and
orientation to seat it fully.
\end{minipage}\par\smallskip
\noindent\begin{minipage}{\linewidth}
\taskdetailheading{OOD settings} We evaluate object OOD with three
held-out charging adapters on the same two power strips
(Fig.~\ref{fig:real_ood_scenes_continued}(e)). The change in housing shape
and appearance tests whether the robot can locate the receiving port and
align the plug for insertion into an unfamiliar
adapter.\par\smallskip
\taskdetailheading{Failure analysis} A representative failure was
misalignment at the receiving port, which prevented the plug from
seating and could cause it to fall beside the power strip (Fig.~\ref{fig:real_failure_modes}(e)). E2EWAM
completed insertion in 55\% of object-OOD trials, compared with 40\%
for OpenWAM-$\alpha$ and 35\% for $\pi_{0.5}$
(Fig.~\ref{fig:real-robot-results}). This suggests that E2EWAM uses fine
visual cues about the plug's position and orientation relative to the
port more reliably when aligning it on unfamiliar adapters.
\end{minipage}\par

\endgroup

\subsection{Task Definitions and Scoring Criteria}
\label{app:real_scoring}

Table~\ref{tab:real_scoring_rubrics} summarizes the task stages and their
progress scores. The same rubric applies to all methods
and ID/OOD conditions.
For $N$ trials, let $q_i\in[0,1]$ be the progress score of trial $i$.
Success requires $q_i=1$. We report success rate (SR) and progress
score (PS), both as percentages over all trials:
\begin{equation}
  \mathrm{SR}=\frac{100}{N}\sum_{i=1}^{N}\mathbf{1}[q_i=1],
  \qquad
  \mathrm{PS}=\frac{100}{N}\sum_{i=1}^{N}q_i.
  \label{eq:real_metrics}
\end{equation}

For the real-robot comparison in Fig.~\ref{fig:real-robot-results}, we run
$N=20$ trials per method for each of the 12 task--condition pairs.
For the data scaling study in Fig.~\ref{fig:data_scaling}(b), we use the same
task-specific OOD settings as in the real-robot comparison.
At each pretraining data scale, we evaluate the generalist policy on all five
tasks with $N=15$ independent trials per task.

\begin{table}[ht]
  \centering
  \caption{Task stages and progress-score criteria.}
  \label{tab:real_scoring_rubrics}
  \small
  \setlength{\tabcolsep}{4pt}
  \renewcommand{\arraystretch}{1.0}
  \begin{tabular}{@{}>{\raggedright\arraybackslash}p{0.25\linewidth}c>{\raggedright\arraybackslash}p{0.65\linewidth}@{}}
    \toprule
    Task & Score & Criterion \\
    \midrule
    \textbf{Clothes Folding} & $0$ & The robot neither lays the garment flat nor completes folding. \\
      & $1/3$ & The garment is laid flat. \\
      & $2/3$ & Folding is complete but unsatisfactory (e.g., uneven folds). \\
      & $1$ & Folding is completed with satisfactory quality. \\
    \midrule
    \textbf{Waste Sorting} & $0$ & No category is fully sorted. \\
      & $1/3$ & Exactly one category is fully sorted. \\
      & $2/3$ & Exactly two categories are fully sorted. \\
      & $1$ & All three categories are fully sorted. \\
      & & A category is fully sorted when all its items are in its designated bin. Categories are counted at the end of the rollout or, if earlier, at the first clear attempt to grasp a distractor; a rollout with such an attempt scores at most $2/3$. \\
    \midrule
    \textbf{Rotating Hanoi Stacking} & $0$ & The rollout fails before the robot picks up the ring. \\
      & $1/3$ & The ring is picked up, but is not released around a post. \\
      & $2/3$ & The ring is placed around an incorrect post and released. \\
      & $1$ & The ring is placed on the post that already holds a ring and released. \\
    \midrule
    \textbf{Iced Tea Making} & $0$ & The rollout fails before the robot opens the refrigerator. \\
      & $1/6$ & \textbf{Open the refrigerator:} The refrigerator door is opened to allow access to its contents. \\
      & $2/6$ & \textbf{Pick up the ice scoop:} The ice scoop is grasped and lifted. \\
      & $3/6$ & \textbf{Add the ice cubes:} The ice cubes are emptied into the dispenser. \\
      & $4/6$ & \textbf{Pick up the kettle:} The kettle is grasped and lifted. \\
      & $5/6$ & \textbf{Pour the water:} Water is poured from the kettle into the dispenser. \\
      & $1$ & \textbf{Dispense the drink:} After all preceding stages are completed, the dispenser tap is pressed to dispense the drink into the serving cup. \\
    \midrule
    \textbf{USB Insertion} & $0$ & The rollout fails before the robot unplugs the USB plug. \\
      & $1/2$ & The USB plug is successfully unplugged. \\
      & $1$ & The USB plug is successfully inserted. \\
    \bottomrule
  \end{tabular}
\end{table}

\FloatBarrier

\begingroup
\makeatletter
\setlength{\@fptop}{0pt}
\makeatother
\clearpage
\subsection{ID and OOD Evaluation Scenes}
\label{app:real_ood_scenes}

Figure~\ref{fig:real_ood_scenes} groups the ID and OOD conditions by task,
using evaluation frames and object cutouts. 

\begin{figure}[!htp]
  \centering
  \iflatexml\includegraphics[width=\linewidth,keepaspectratio]{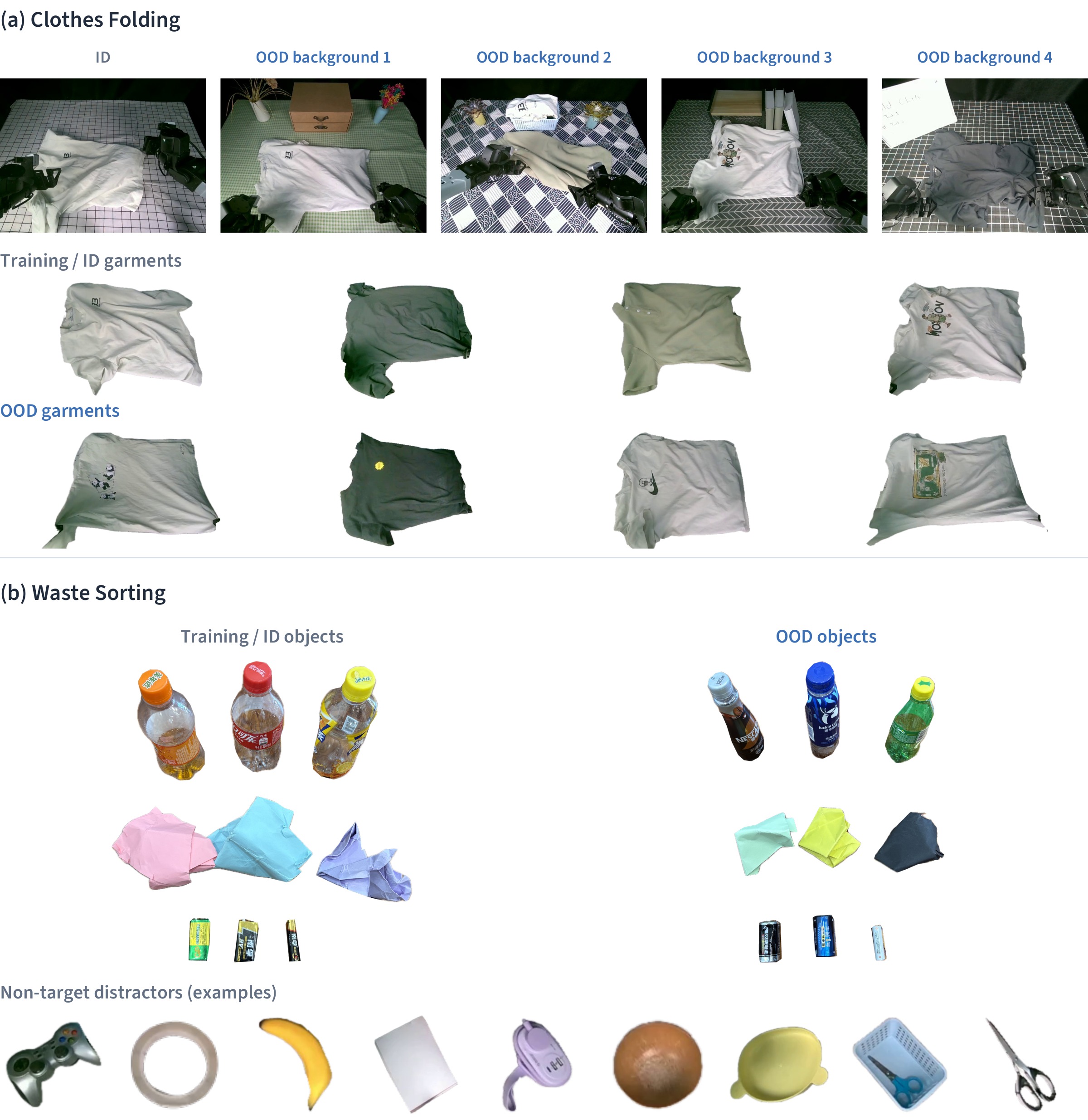}\else\includegraphics[width=\linewidth,keepaspectratio]{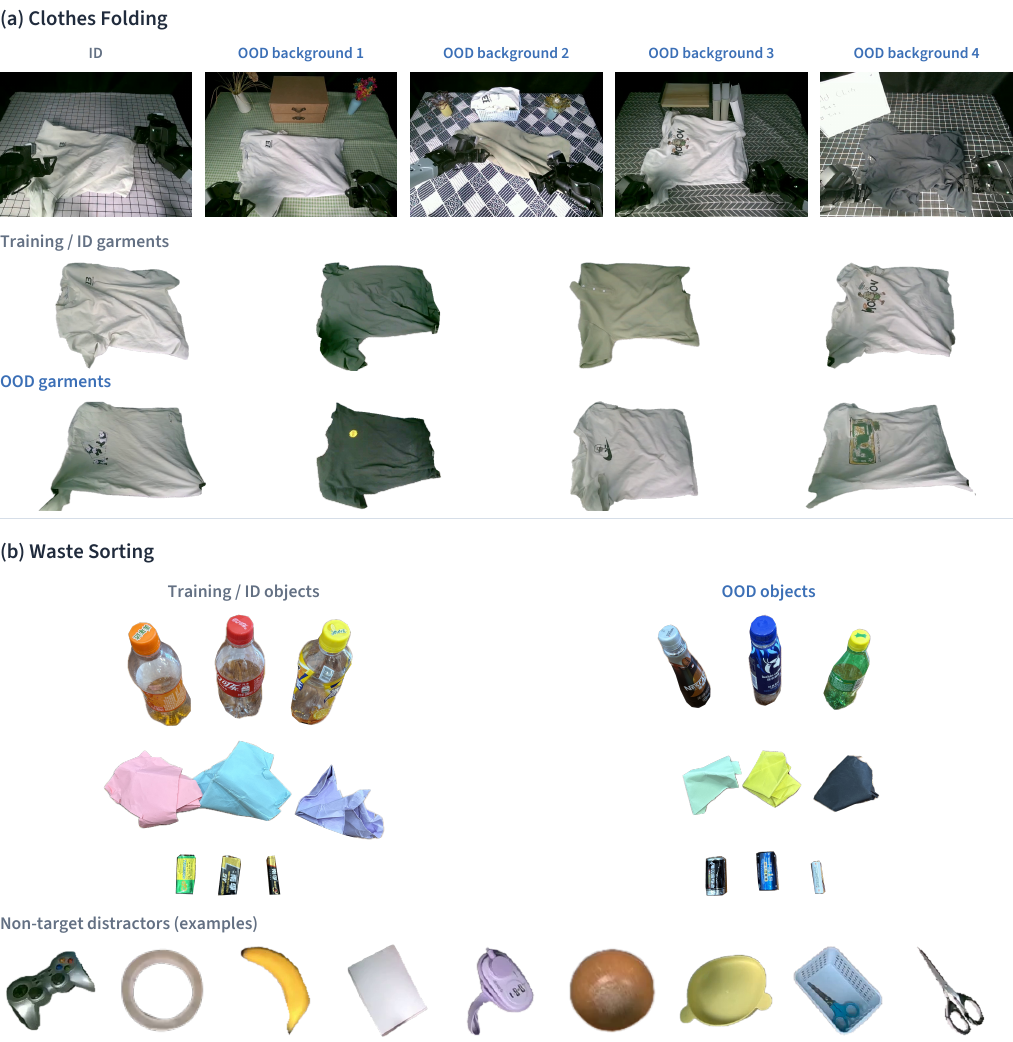}
\fi
  \caption{\textbf{ID and OOD conditions by task.}
  (a) Clothes Folding: the ID scene and four OOD backgrounds, followed by
  four training garments and four unseen garments. OOD evaluation changes
  the garment, the scene, or both.
  (b) Waste Sorting: training and OOD sets each contain three bottle types,
  three battery types, and three paper-ball colors. Example non-target
  distractors appear below.}
  \label{fig:real_ood_scenes}
\end{figure}

\clearpage
\iflatexml
  \addtocounter{figure}{-1}
  \let\ContinuedFloat\relax
\fi
\begin{figure}[!ht]
  \ContinuedFloat
  \centering
  \iflatexml\includegraphics[width=0.77\linewidth]{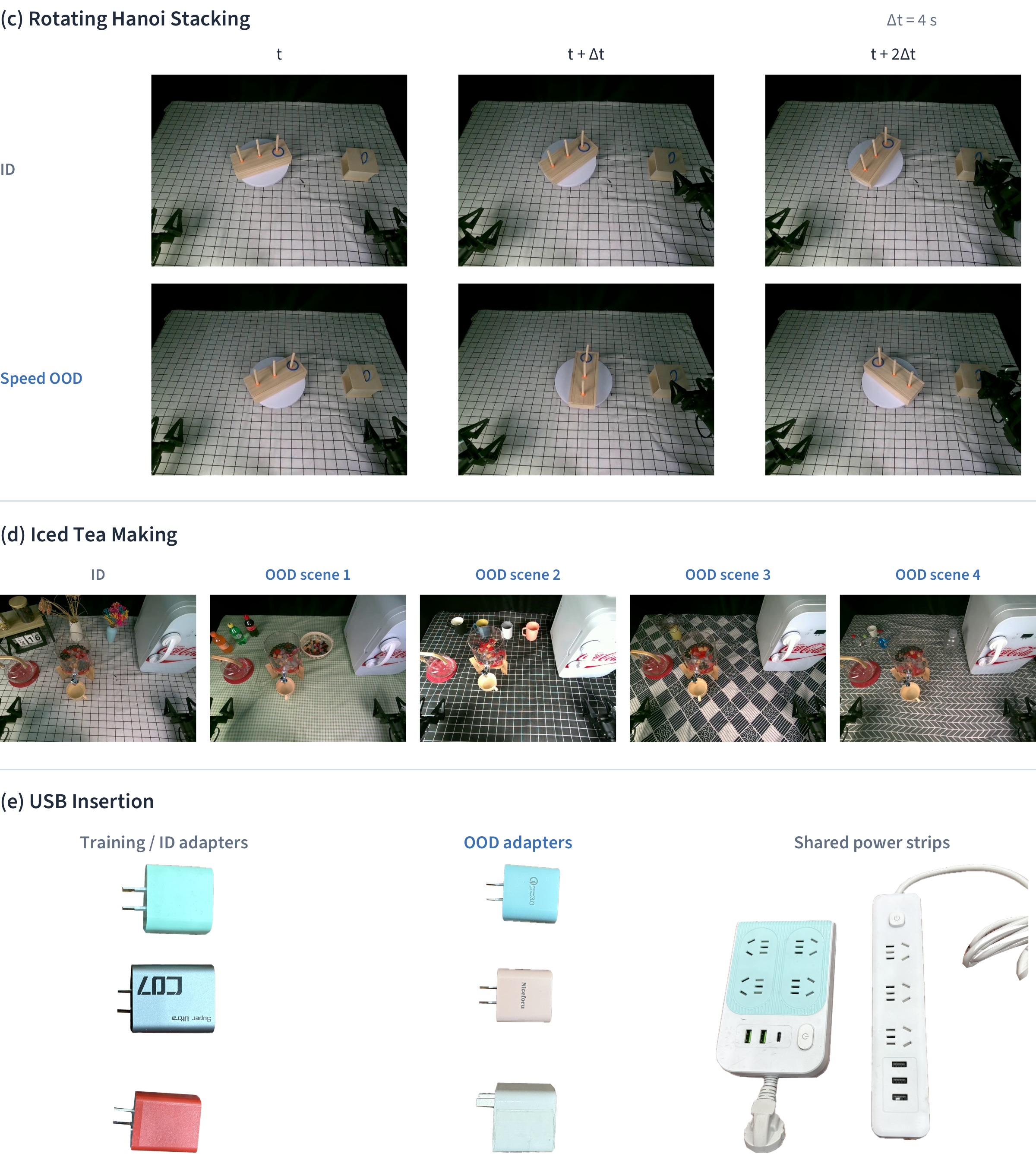}\else\includegraphics[width=0.77\linewidth]{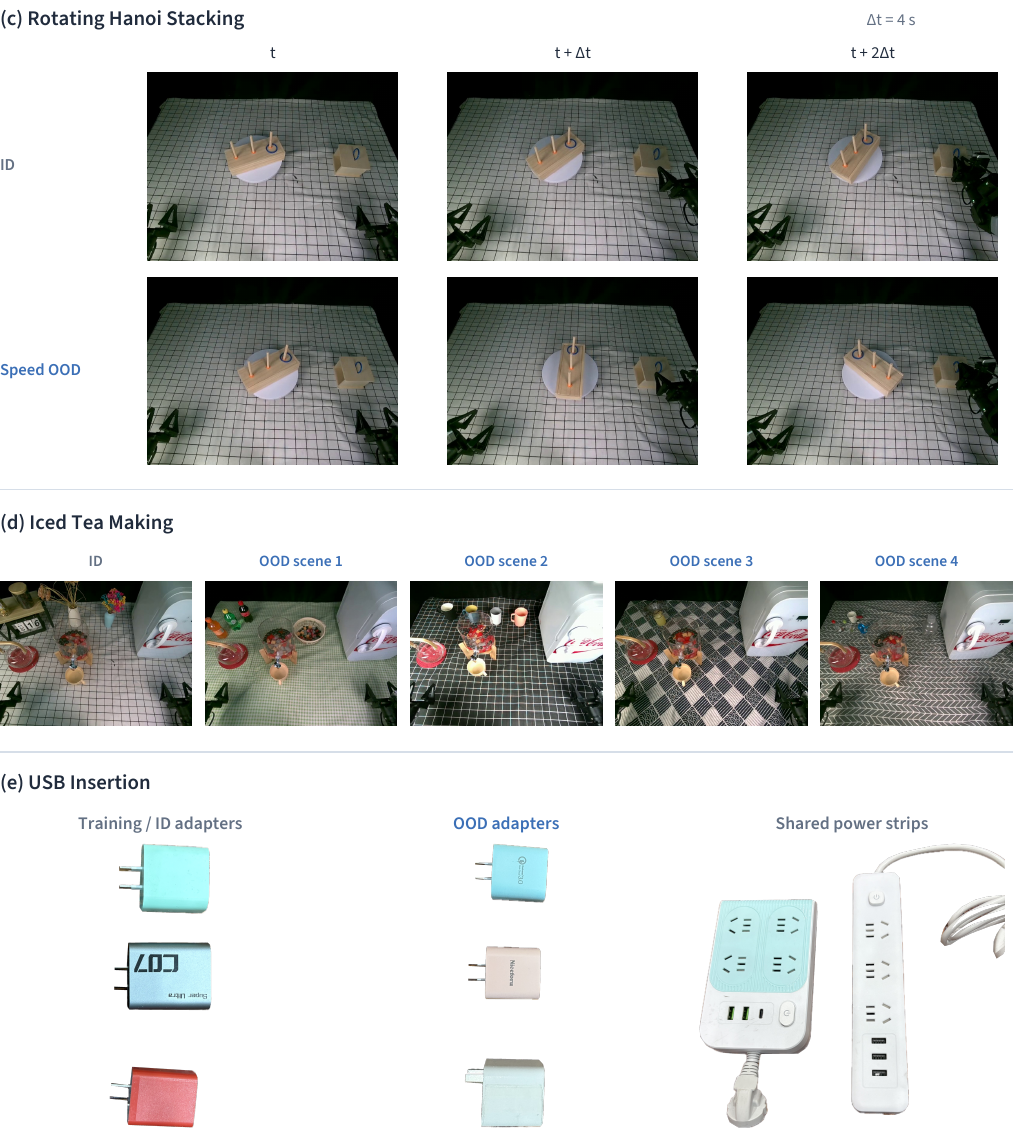}
\fi
  \caption{\textbf{ID and OOD conditions by task (continued).}
  (c) Rotating Hanoi Stacking: higher rotation speed under OOD, with unchanged ring
  colors; adjacent frames are separated by 4\,s.
  (d) Iced Tea Making: four joint background-and-clutter changes, with unchanged
  task objects.
  (e) USB Insertion: three training charging adapters and three held-out
  charging adapters used with the same two power strips.}
  \label{fig:real_ood_scenes_continued}
\end{figure}

\FloatBarrier
\endgroup

\iflatexml
\begin{table}[!ht]
\centering
\caption{Real-robot training data.}
\else
\noindent\begin{minipage}{\linewidth}
\begin{center}
  \captionsetup{hypcap=false}
  \captionof{table}{Real-robot training data.}
\fi
  \label{tab:real_demonstrations}
  \small
  \setlength{\tabcolsep}{9pt}
  \renewcommand{\arraystretch}{1.10}
  \begin{tabular}{@{}lrrr@{}}
    \toprule
    Task & Episodes & Mean duration (s) & Total duration (min) \\
    \midrule
    Clothes Folding & 95 & 48.00 & 76.00 \\
    Waste Sorting & 84 & 41.57 & 58.20 \\
    Rotating Hanoi Stacking & 100 & 28.00 & 46.67 \\
    Iced Tea Making & 94 & 77.49 & 121.40 \\
    USB Insertion & 172 & 13.93 & 39.94 \\
    \midrule
    \textbf{Total} & \textbf{545} & \textbf{37.67} & \textbf{342.21} \\
    \bottomrule
  \end{tabular}
\iflatexml
\end{table}
\else
\end{center}
\end{minipage}
\fi
\par\clearpage
\setcounter{figure}{\savedRealFigureCounter}
\endgroup

\end{document}